\documentclass[10pt,twocolumn,letterpaper]{article}

\usepackage[accsupp]{axessibility}  
\usepackage{iccv}
\usepackage{times}
\usepackage{epsfig}
\usepackage{graphicx}
\usepackage{amsmath}
\usepackage{amssymb}

\usepackage{multirow}
\usepackage{graphicx}
\usepackage{caption}

\usepackage{enumitem}
\usepackage{comment}
\usepackage{subcaption}
\usepackage{xspace}
\makeatletter
\DeclareRobustCommand\onedot{\futurelet\@let@token\@onedot}
\def\@onedot{\ifx\@let@token.\else.\null\fi\xspace}
\def\eg{\emph{e.g}\onedot}

\def\etal{\emph{et al}\onedot}
\makeatother

\usepackage{amssymb}
\usepackage{amsmath}
\newcommand{\bx}{\boldsymbol{x}}
\newcommand{\be}{\boldsymbol{e}}
\newcommand{\by}{\boldsymbol{y}}

\usepackage{dirtytalk}

\usepackage[table, x11names]{xcolor}
\usepackage{colortbl}

\usepackage{xspace}

\usepackage[breaklinks=true,bookmarks=false]{hyperref}

\iccvfinalcopy 

\def\iccvPaperID{****} 

\ificcvfinal\fi

\begin{document}

\title{Multi-Modal RGB-D Scene Recognition Across Domains}

\author{Andrea Ferreri$^{1}$ \hspace{5mm} Silvia Bucci$^{1,2}$ \hspace{5mm} Tatiana Tommasi$^{1,2}$\\  
 $^1$Politecnico di Torino\hspace{5mm} $^2$Italian Institute of  Technology, Italy \\ 
\tt\small  \hspace{-7mm} andrea.ferreri@studenti.polito.it  \hspace{5mm}\{silvia.bucci,  tatiana.tommasi\}@polito.it }

\maketitle
\ificcvfinal\thispagestyle{empty}\fi

\begin{abstract}
Scene recognition is one of the basic problems in computer vision research with extensive applications in robotics. When available, depth images provide helpful geometric cues that complement the RGB texture information and help to identify discriminative scene image features. 

Depth sensing technology developed fast in the last years and a great variety of 3D cameras have been introduced, each with different acquisition properties. However, those properties are often neglected when targeting big data collections, so multi-modal images are gathered disregarding their original nature. In this work, we put under the spotlight the existence of a possibly severe domain shift issue within multi-modality scene recognition datasets. As a consequence, a scene classification model trained on one camera may not generalize on data from a different camera,  only providing a low recognition performance. Starting from the well-known SUN RGB-D dataset, we designed an experimental testbed to study this problem and we use it to benchmark the performance of existing methods.

Finally, we introduce a novel adaptive scene recognition approach that leverages self-supervised translation between modalities. Indeed, learning to go from RGB to depth and vice-versa is an unsupervised procedure that can be trained jointly on data of multiple cameras and may help to bridge the gap among the extracted feature distributions. Our experimental results confirm the effectiveness of the proposed approach.

\end{abstract}

\section{Introduction}
\begin{figure}[!t]
\resizebox{\columnwidth}{!}{
\begin{tabular}{c@{~}c@{~}c@{~}c@{~}c}
\multirow{2}{*}{\small{}} &  {Kinect v1} & {Kinect v2} & {Realsense} & {Xtion} \\
\cline{2-5}
 & \multicolumn{4}{c}{classroom} \\
{\rotatebox{90}{\quad ~\small{RGB}}} &
\includegraphics[width=0.1\textwidth]{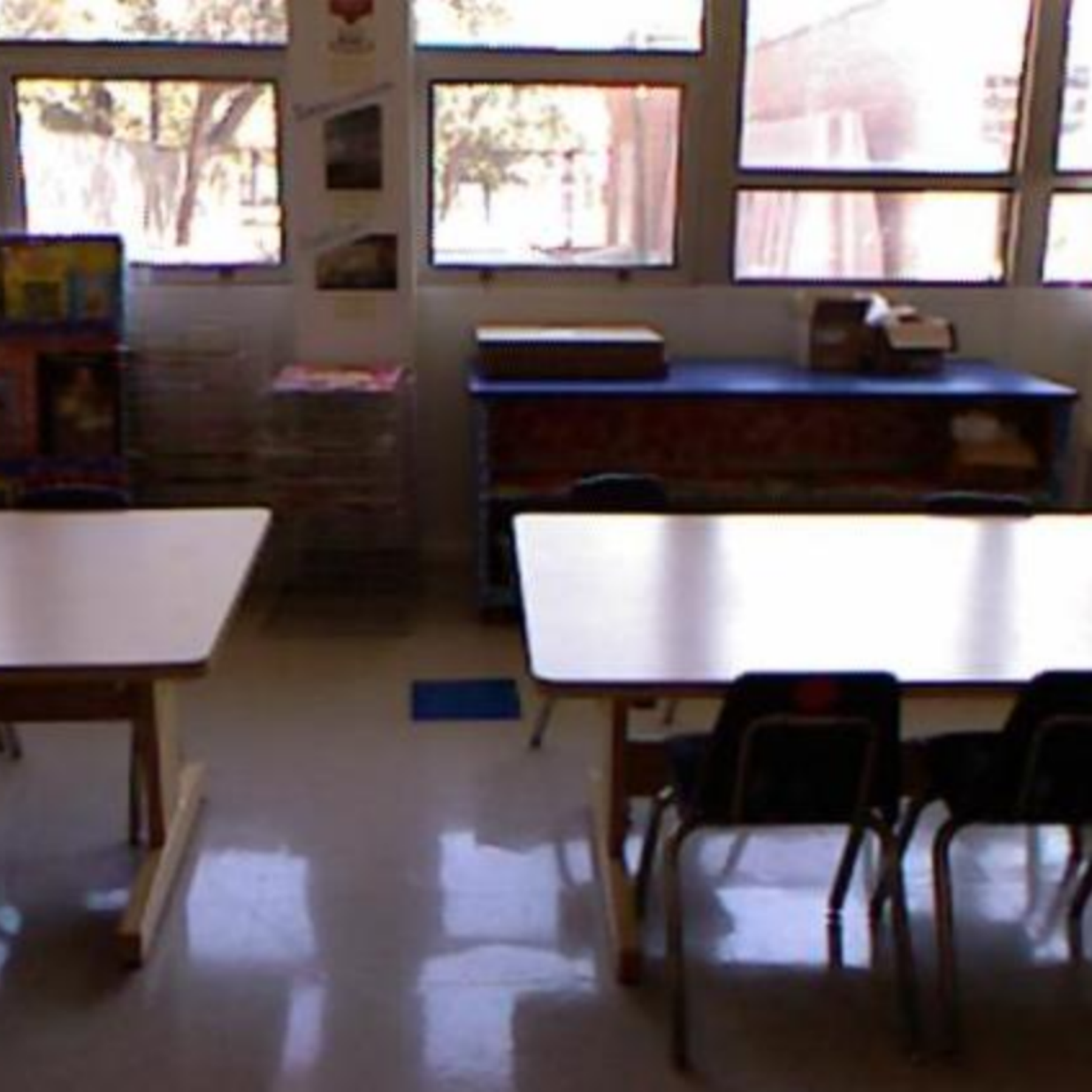} & \includegraphics[width=0.1\textwidth]{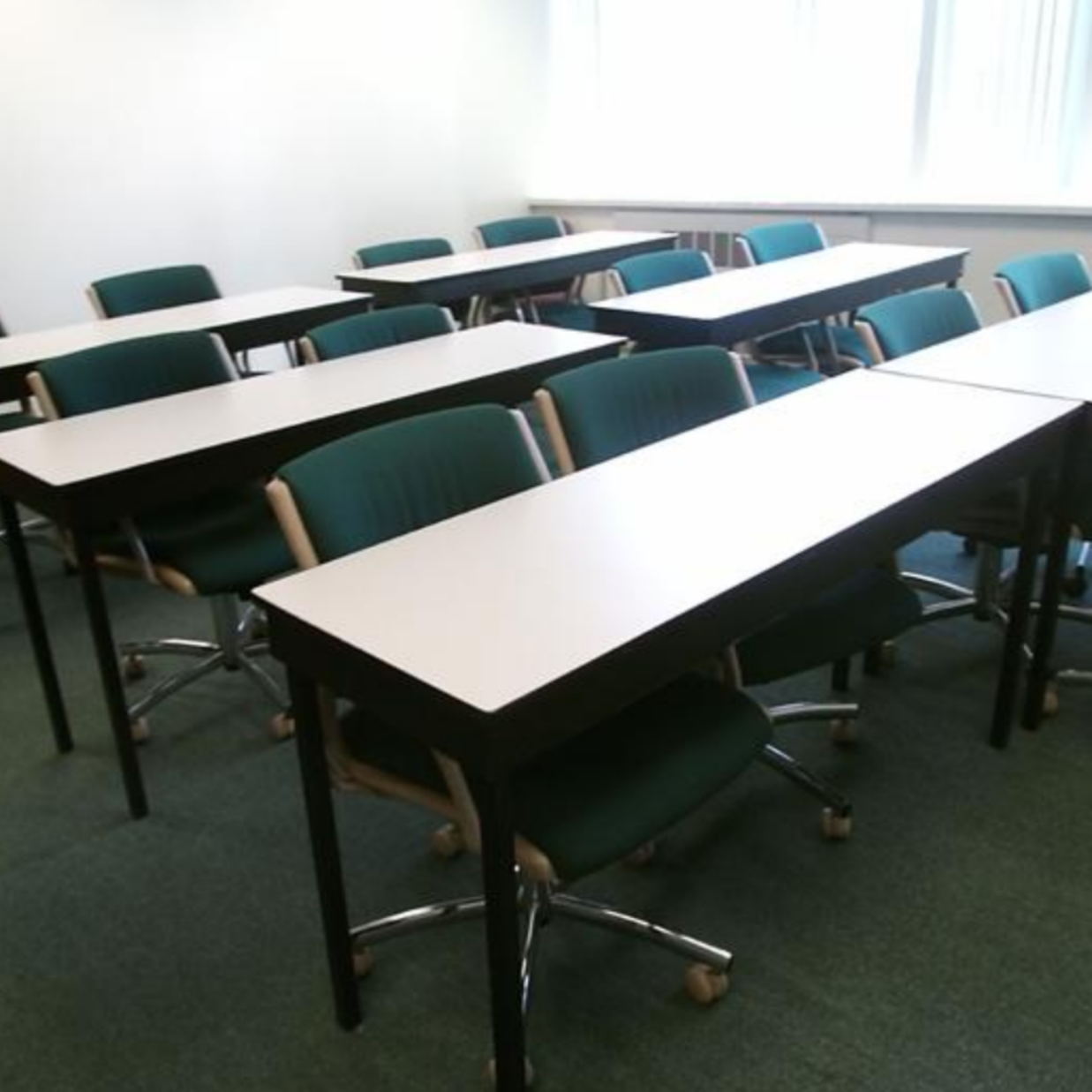} & \includegraphics[width=0.1\textwidth]{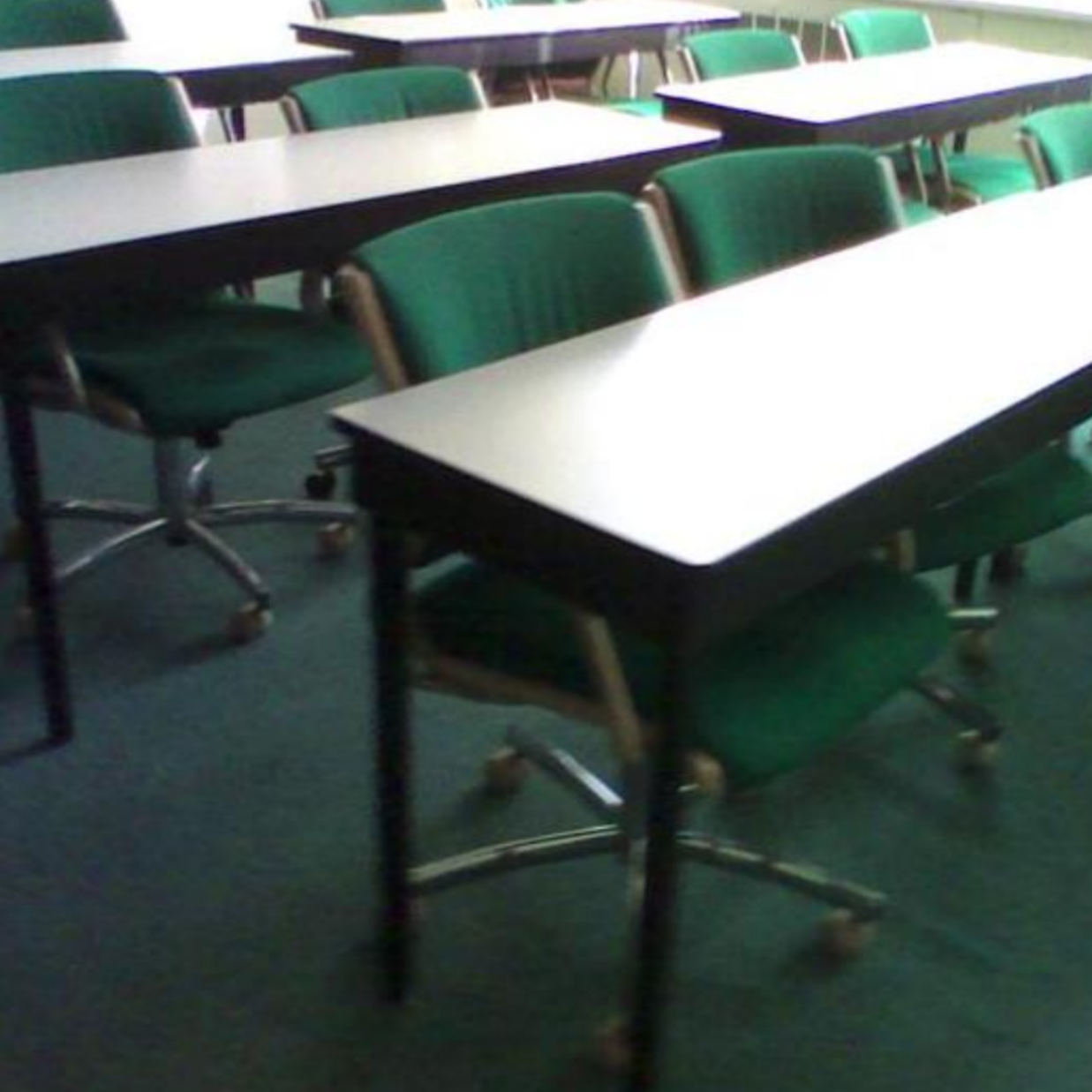} & \includegraphics[width=0.1\textwidth]{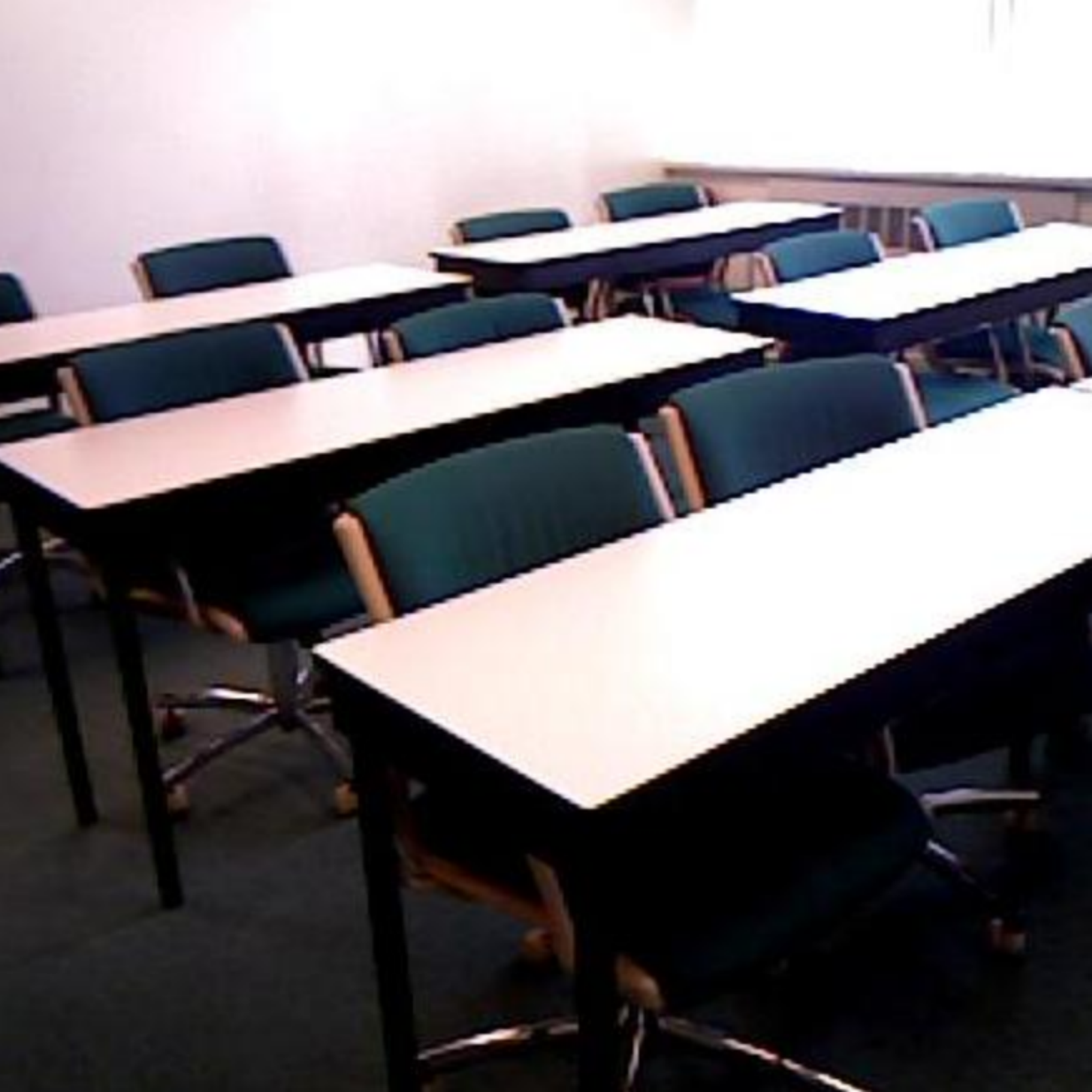}\\
 \rotatebox{90}{\small{Depth (HHA)}} & 
\includegraphics[width=0.1\textwidth]{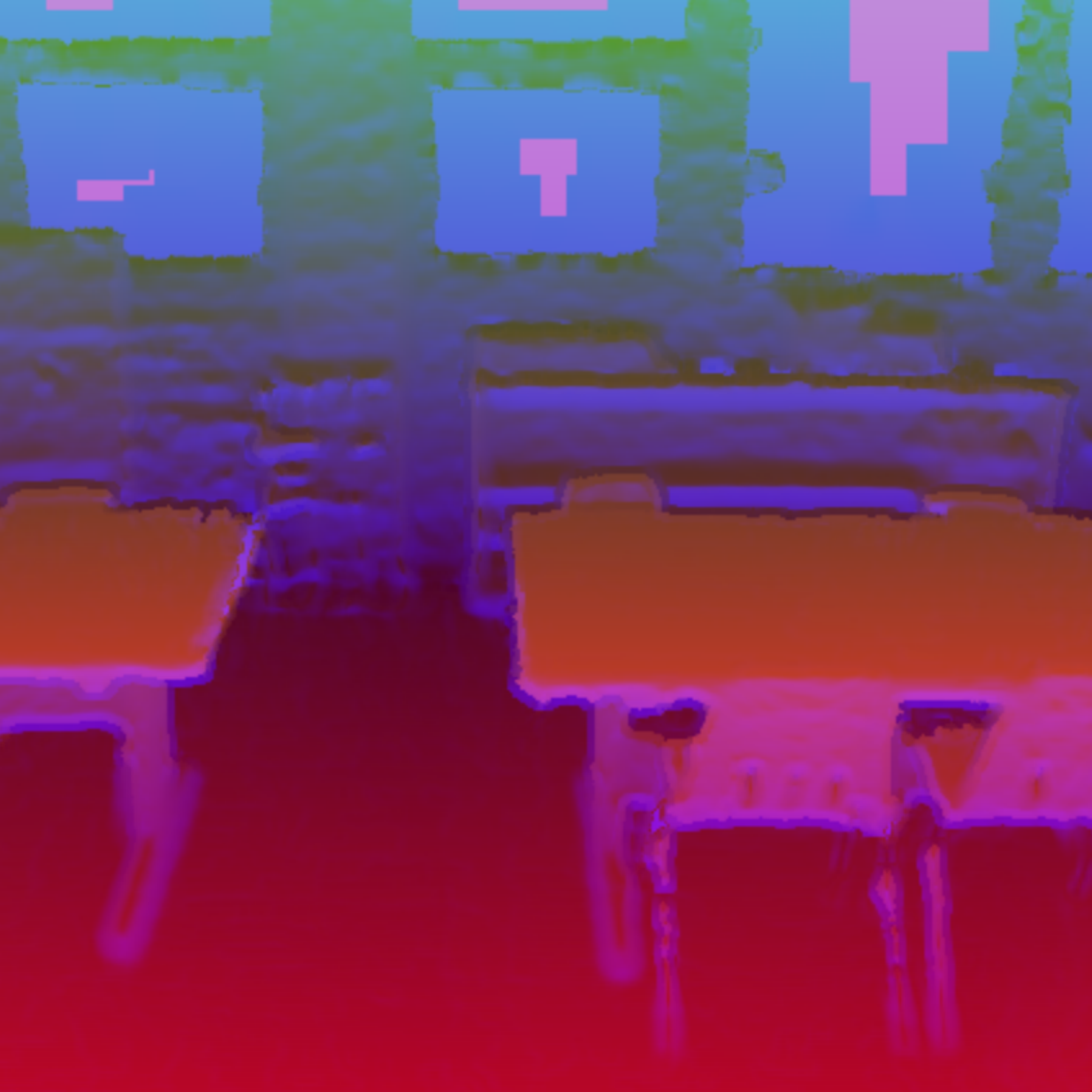} & \includegraphics[width=0.1\textwidth]{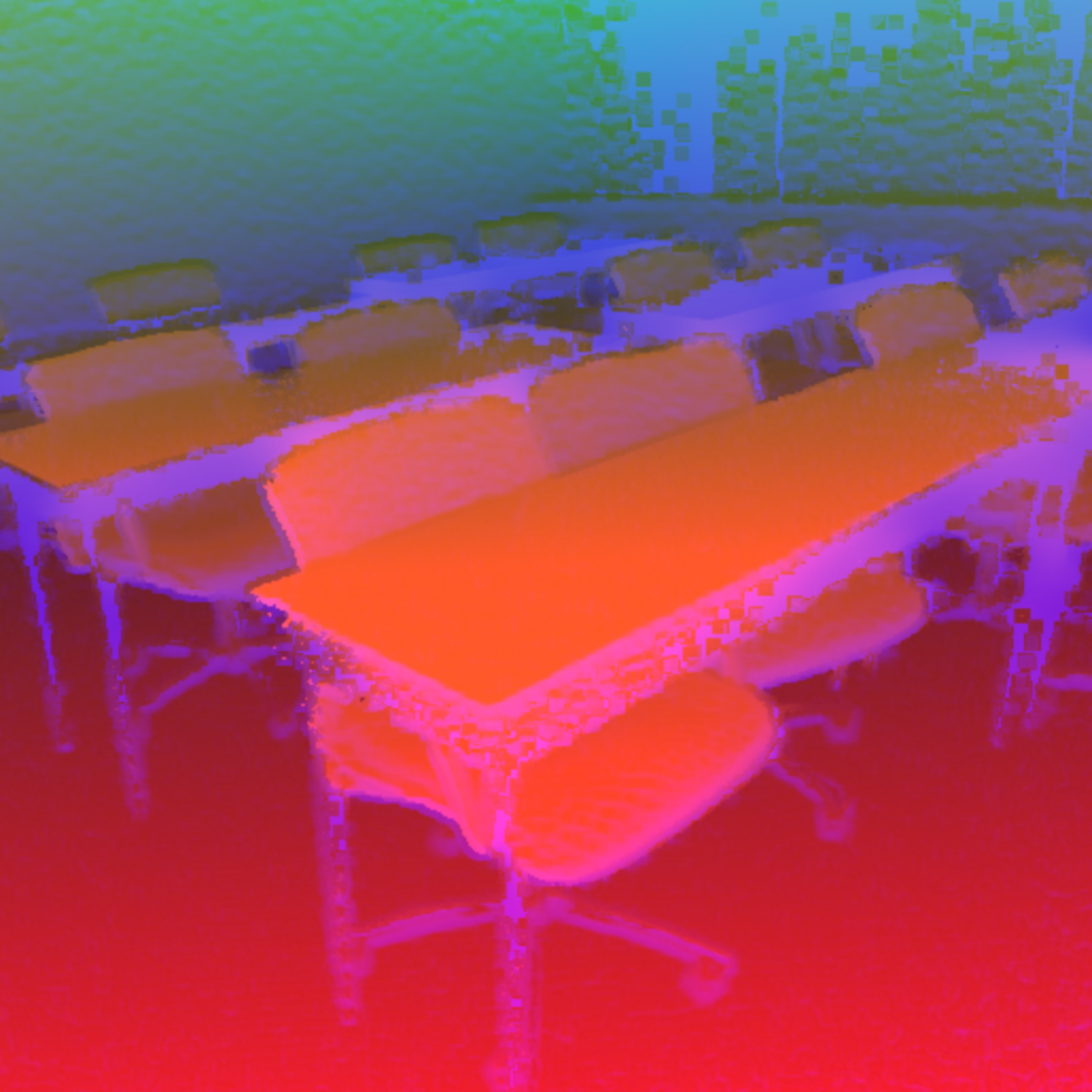} & \includegraphics[width=0.1\textwidth]{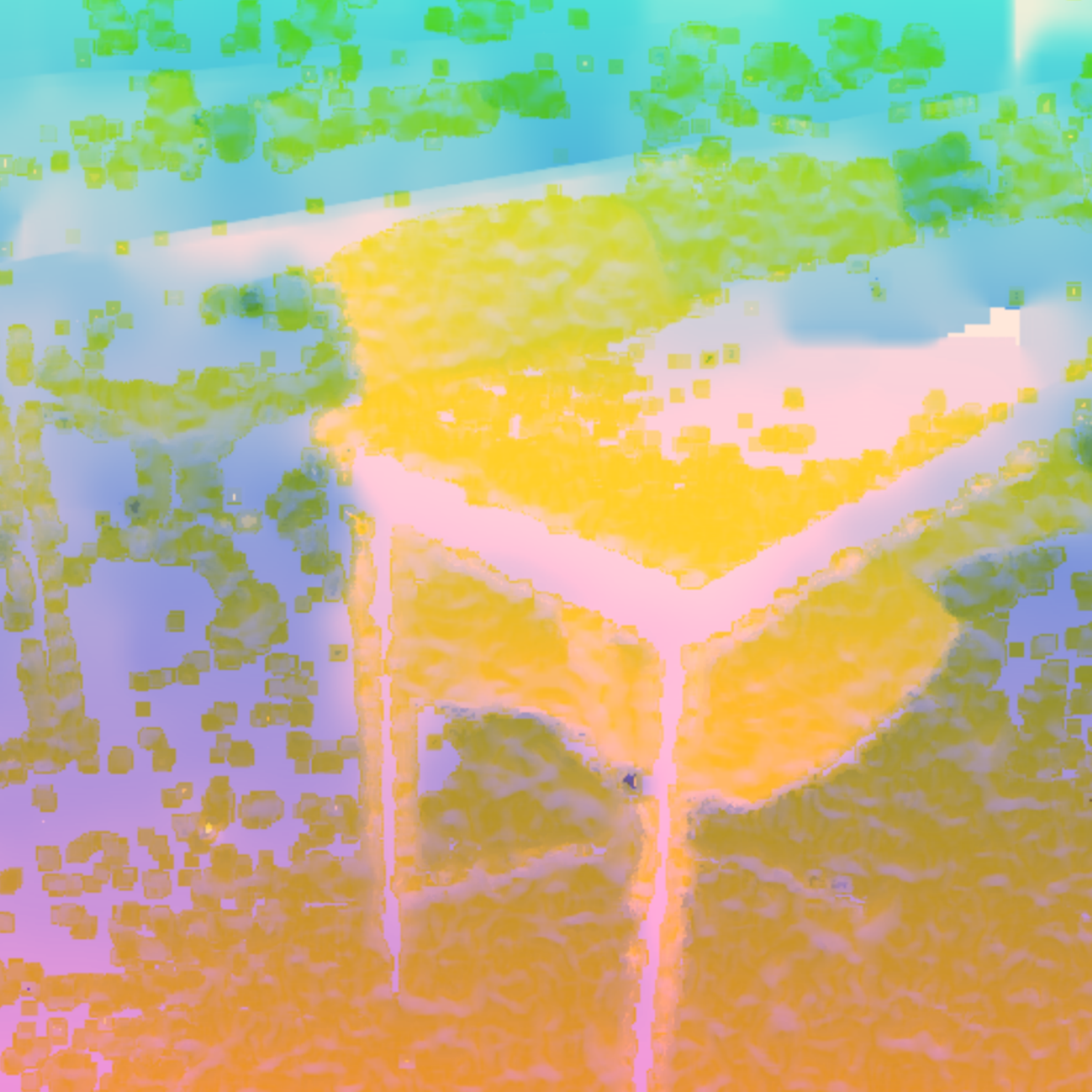} & \includegraphics[width=0.1\textwidth]{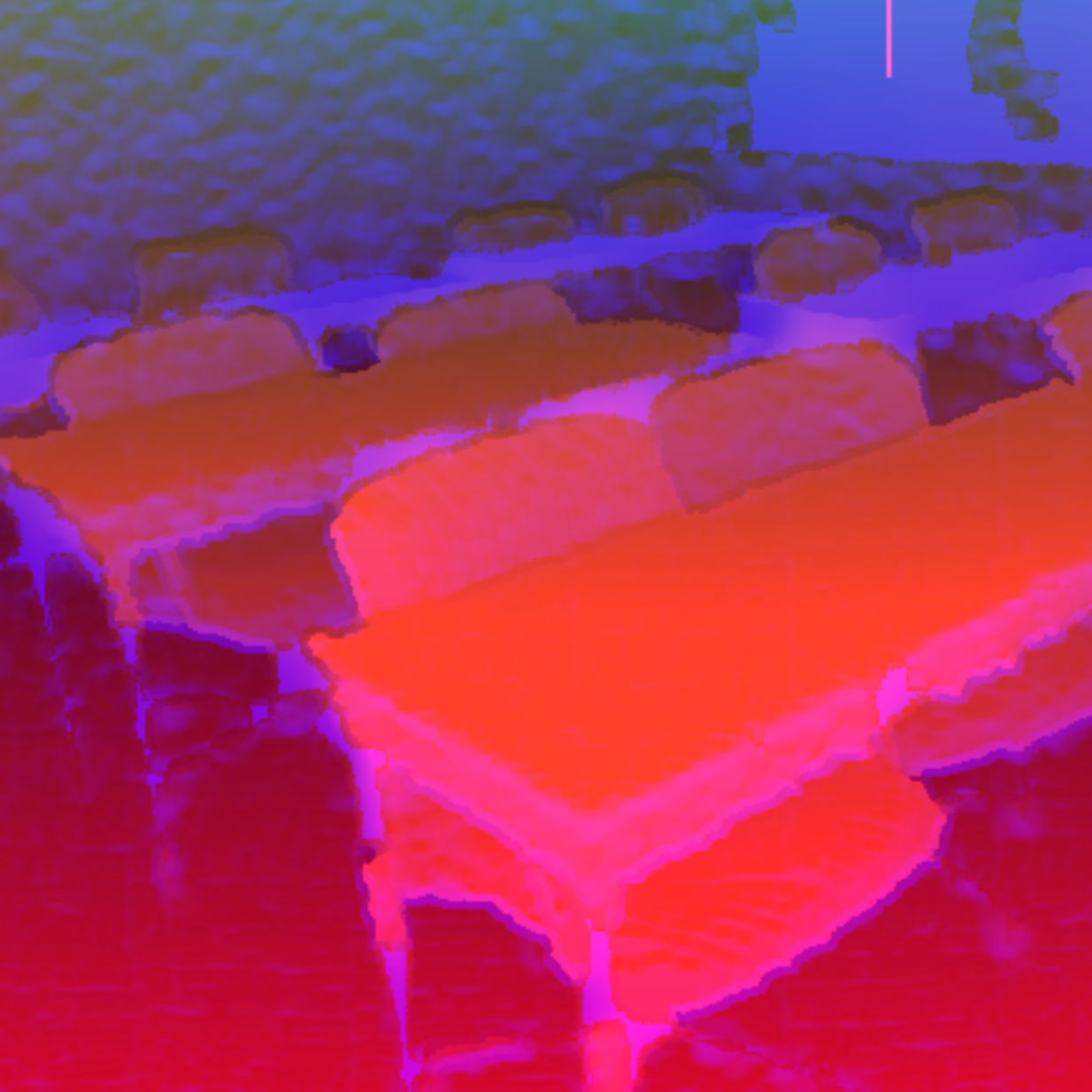}\\
 \hline
 & \multicolumn{4}{c}{discussion\_area} \\
 \rotatebox{90}{\quad ~\small{RGB}} &
 \includegraphics[width=0.1\textwidth]{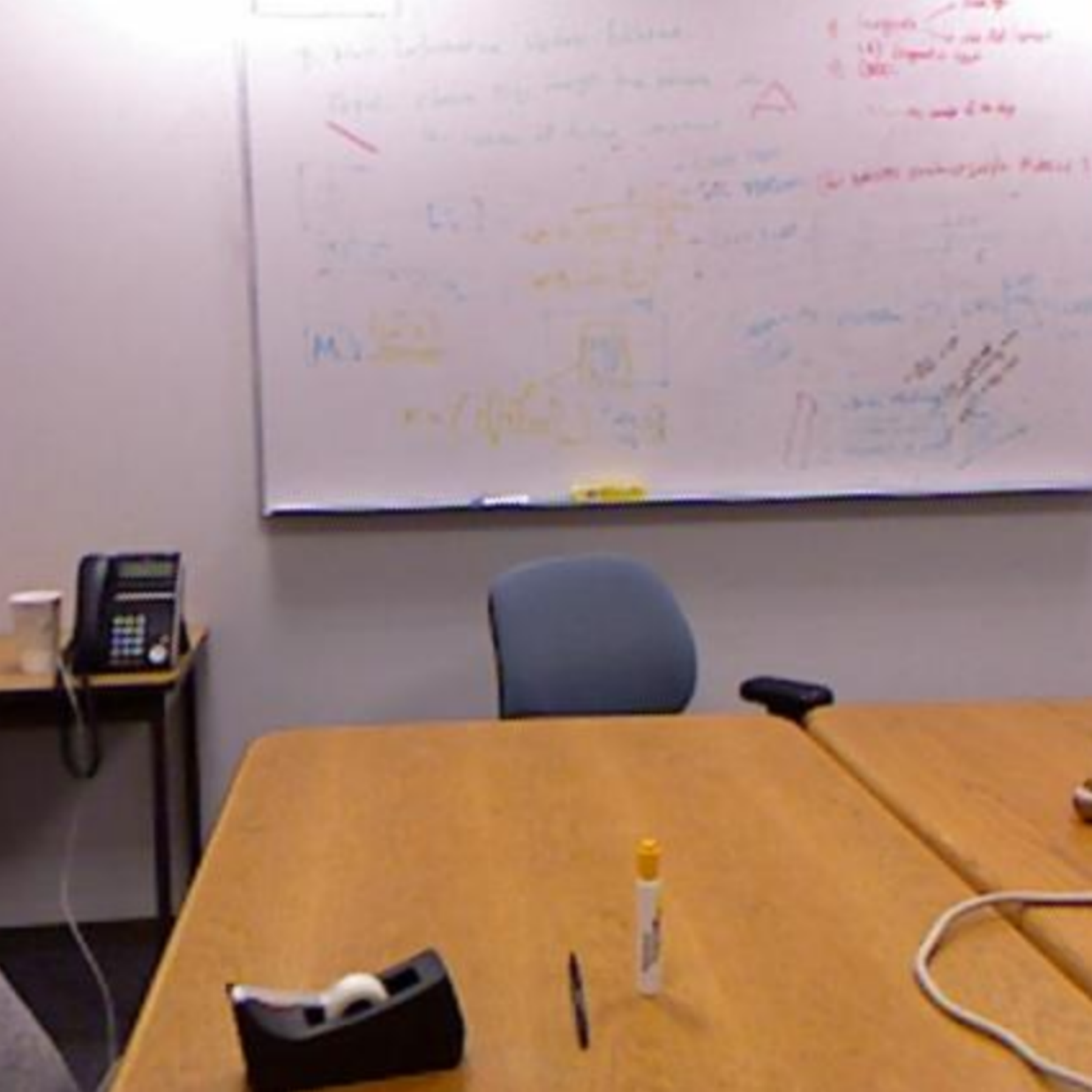} & \includegraphics[width=0.1\textwidth]{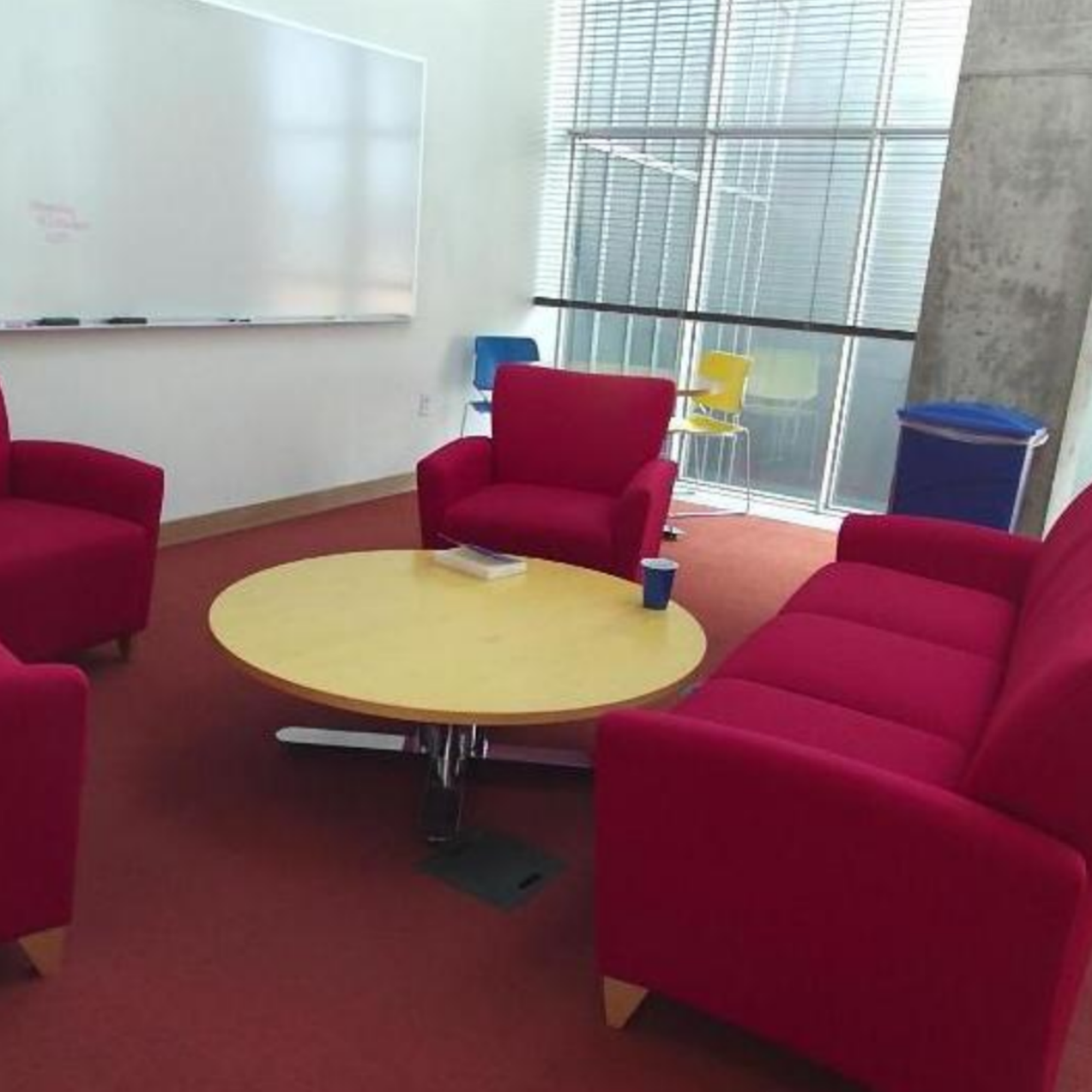} & \includegraphics[width=0.1\textwidth]{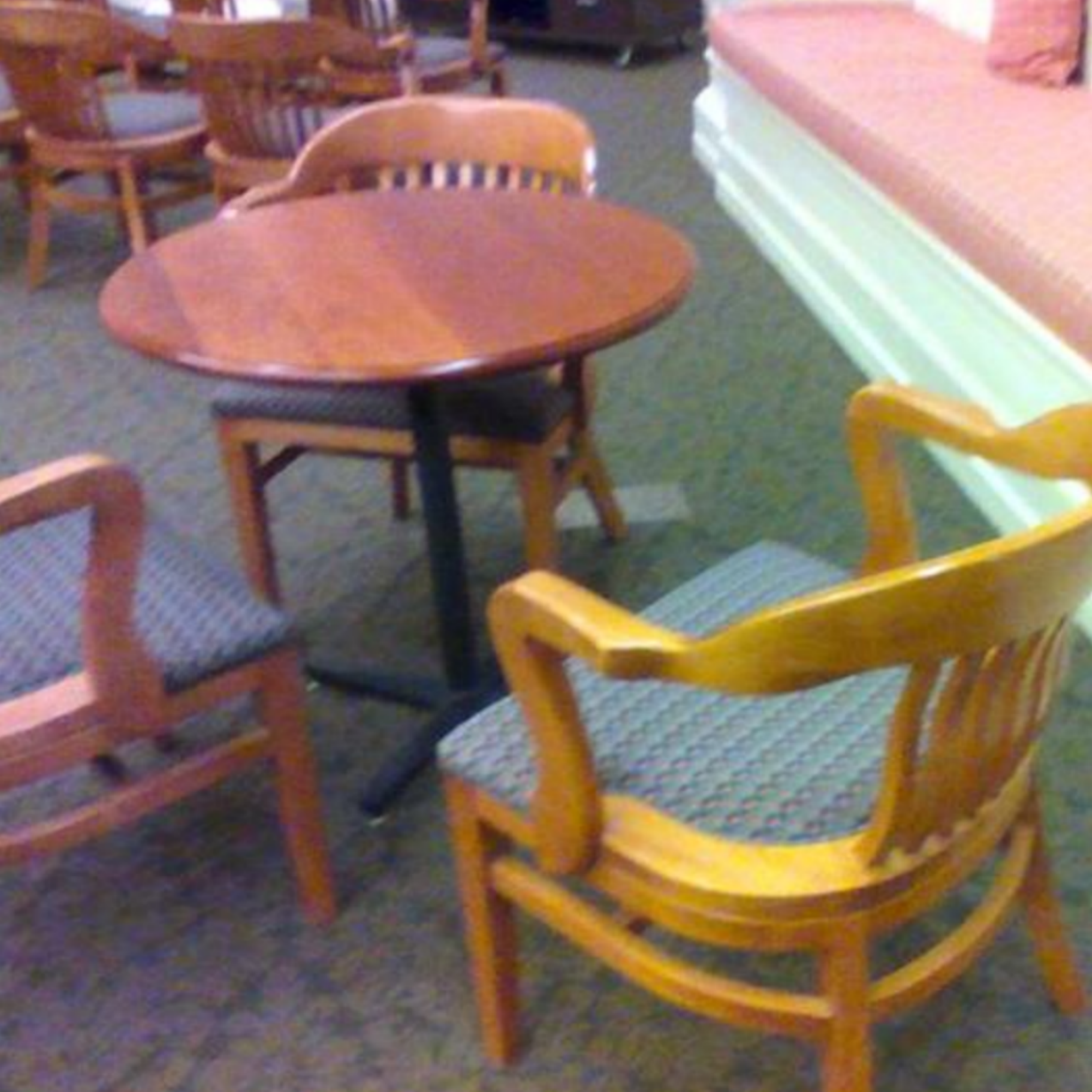} & \includegraphics[width=0.1\textwidth]{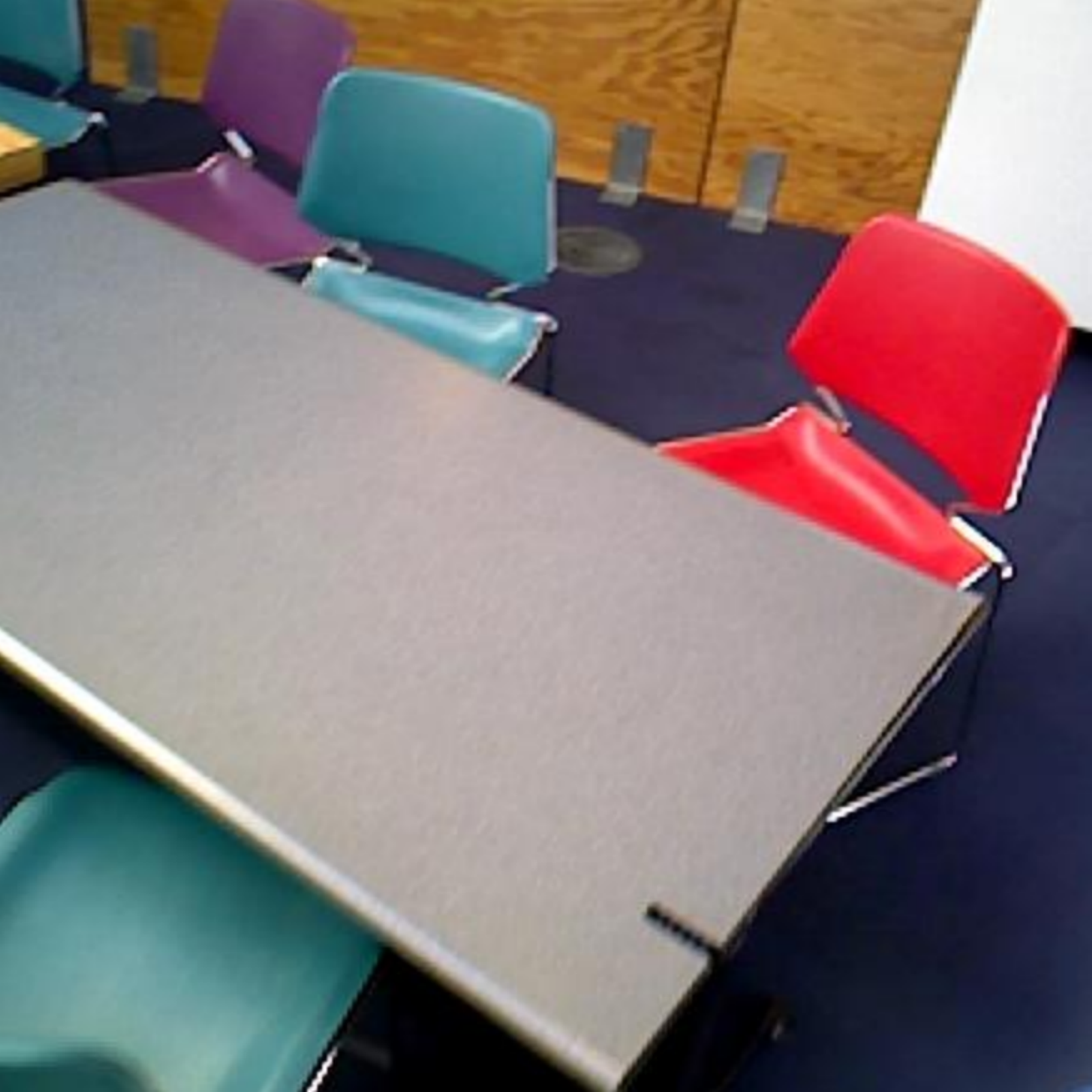} 
 \\
   \rotatebox{90}{\small{Depth (HHA)}} & 
   \includegraphics[width=0.1\textwidth]{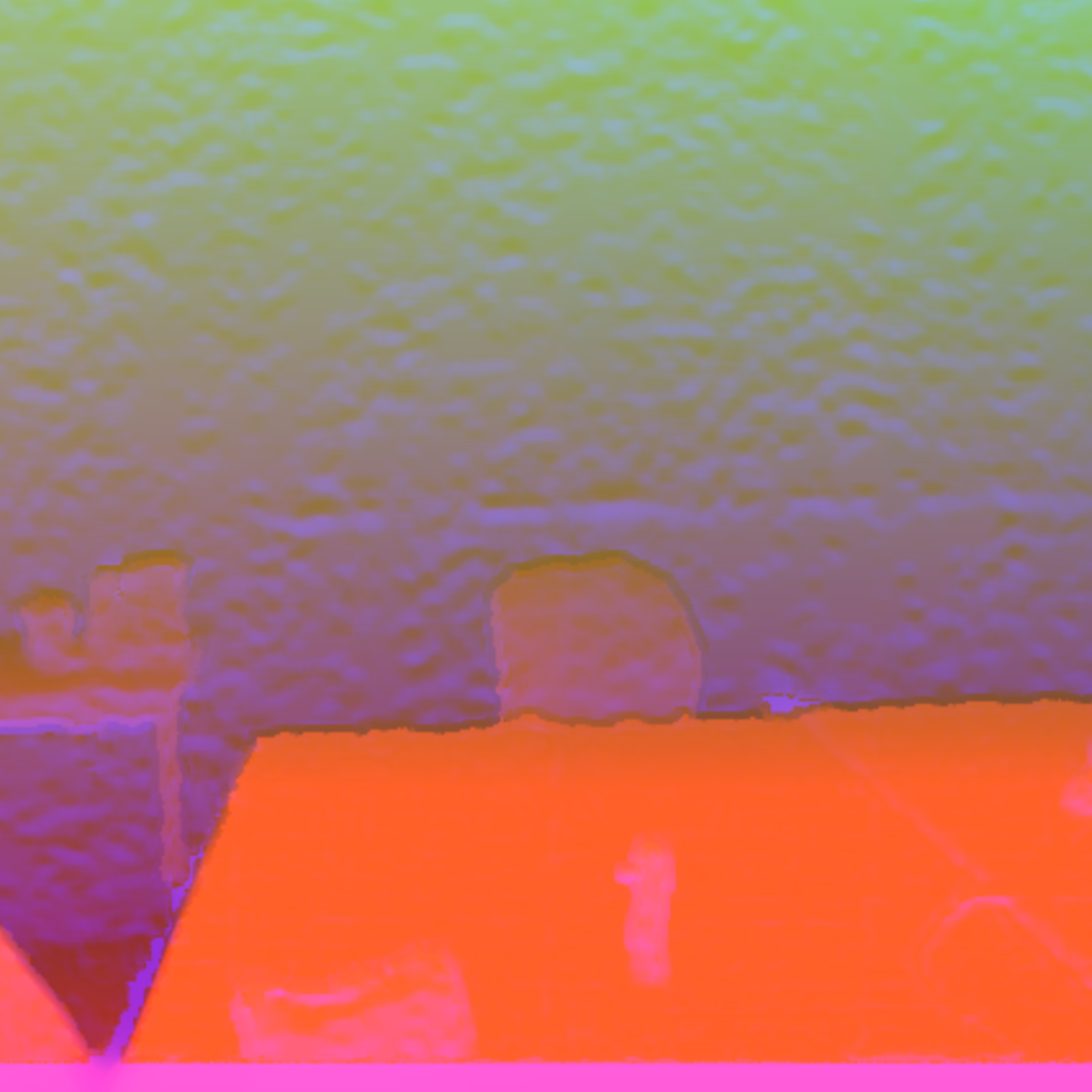} & \includegraphics[width=0.1\textwidth]{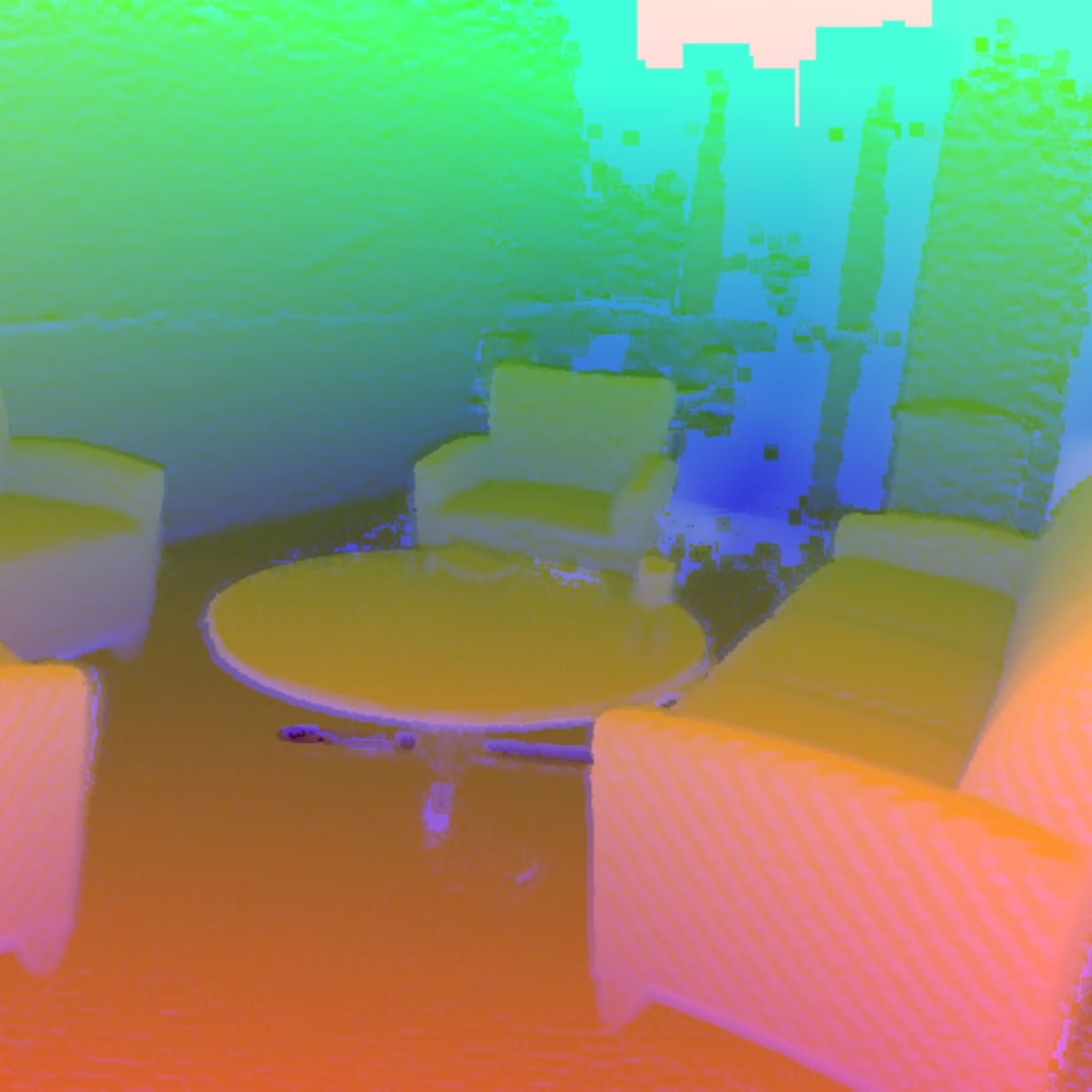} & \includegraphics[width=0.1\textwidth]{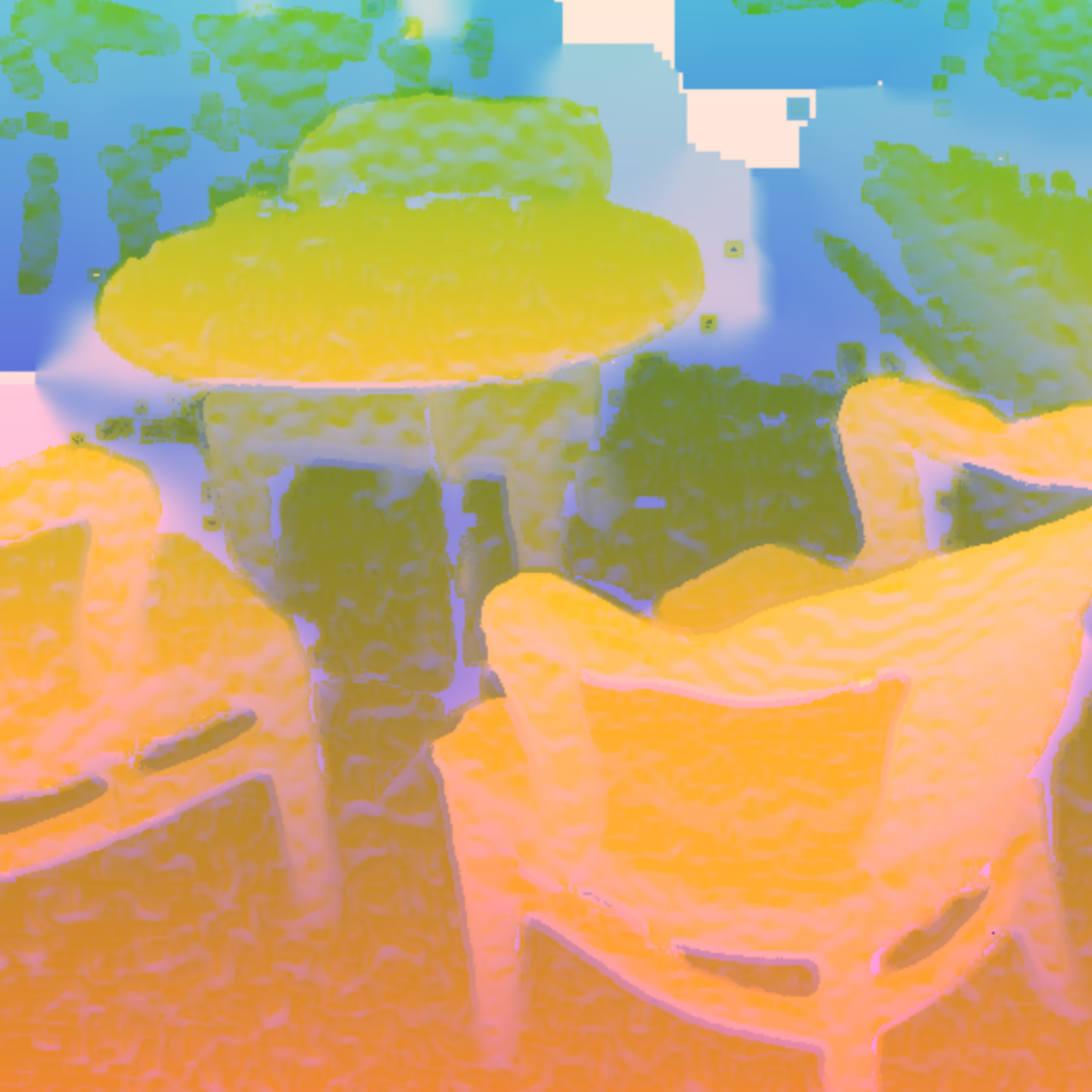} & \includegraphics[width=0.1\textwidth]{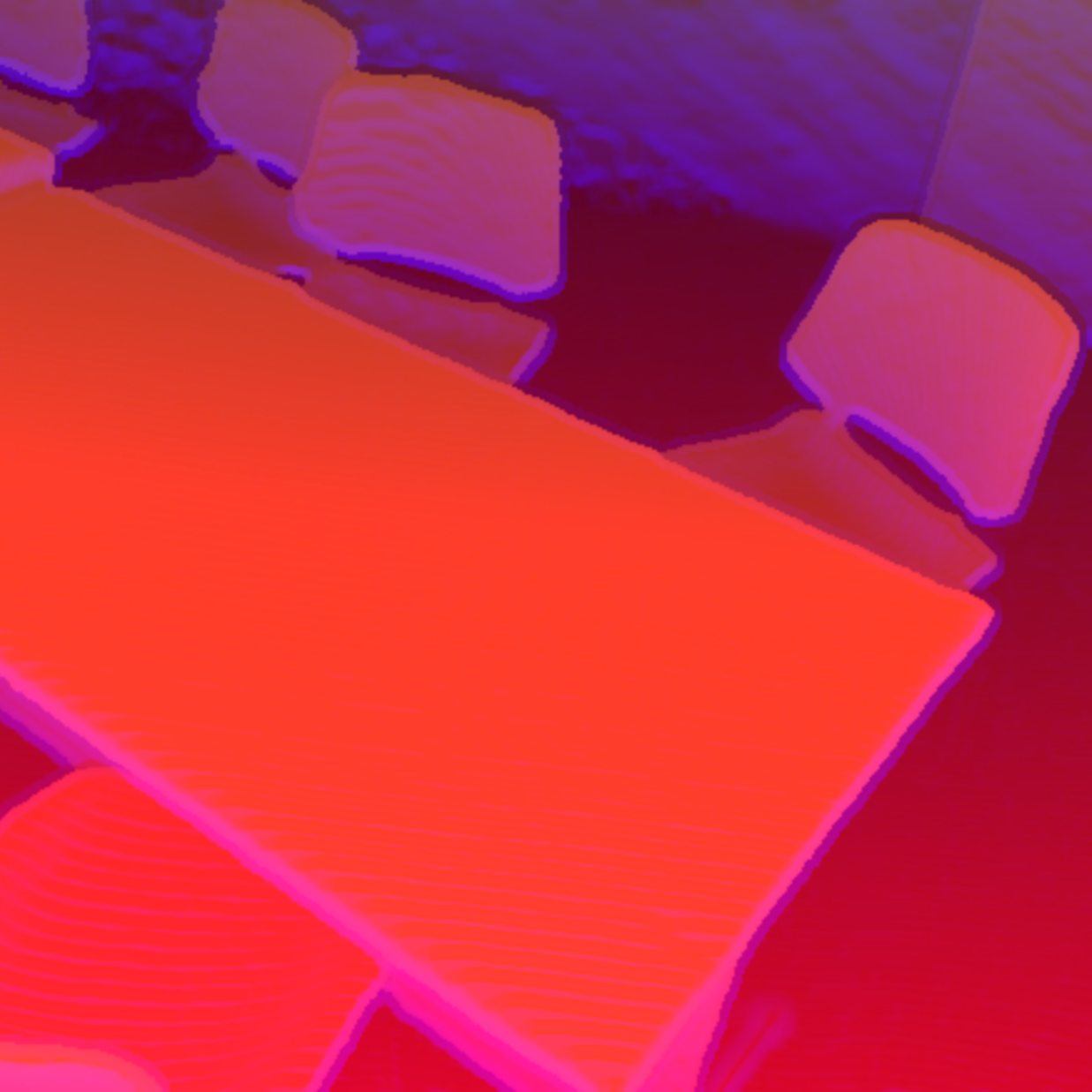} 
   \\
 \hline
\end{tabular}
}
\caption{Examples of RGB and Depth HHA \cite{gupta2014hha} images from all the cameras within the SUN RGB-D dataset \cite{sunrgbd}. The category \emph{classroom} contains images taken in the exact same place with Kinect v2, Realsense, and Xtion cameras, while the physical location captured with Kinect v1 is different although annotated with the same label.
For the category \emph{discussion\_area} there is no room overlap: despite the shared label, each camera captured images in a different physical location.
As can be noticed, the specific camera characteristics contribute to produce significant appearance differences. Best seen in color.
}
\label{fig:teaser}\vspace{-5mm}
\end{figure}

\emph{Scene recognition} consists in assigning a label as kitchen, office, bakery, or beach to an image, and it is a crucial vision problem for robot localization and decision-making \cite{survey_scene,survey_deep_scene}. An artificial learning agent needs to understand its surrounding environment by recognizing objects with their correlations and being robust to clutter which causes large intra-scene variations and inter-scene overlap. In this scenario, RGB images provide relevant appearance cues, while depth (D) information is essential to model object boundaries and capture the 3D space layout. 

Although gathering RGB scene images may be relatively easy (\eg by crawling the web), collecting a large RGB-D dataset is more difficult. This issue has initially moved research faster in the direction of RGB data-driven representation learning, as in the case of CNN models trained on the Places dataset \cite{zhou2014scenerecog}. In the last years, the diffusion of low-cost depth sensors has allowed to access sizable amounts of RGB-D images and several \emph{multi-modal learning} methods have been developed. Most of them feed depth and RGB samples through separate deep learning paths and the obtained representations are finally fused with different strategies \cite{Ayub_2020_BMVC,Yuan_Xiong_Wang_2019,song2018scenerecog,AAAI_df2}. 
Still, the existing literature leaves behind some important analysis on the nature of the used data which are considered as drawn from a single domain distribution. The generic name \say{RGB-D} hides a plethora of 3D cameras which may differ in many aspects, from the exact depth sensing technology (structured light, time-of-flight, active stereo), to the range and the field of view for the images. This sums over the existing variability within scenes annotated with the same class label but captured in different physical locations by heterogeneous cameras. Thus, several causes contribute to a significant domain shift among the data (see Figure \ref{fig:teaser}) and question the robustness of the developed approaches. 

\emph{Domain adaptation} addresses the problem of learning models on some source labeled data distribution that generalizes to a different unlabeled target distribution \cite{csurka_book}. Several techniques have been proposed to close visual domain gaps between synthetic and real images or photos and art pictures, but in all those cases both the source and the target domain are composed of single-modal instances (only RGB) or only one of the two domains has an extra modality (source RGB-D, target RGB). Moreover, those works usually tackle cross-domain object classification \cite{AFN,Ganin:DANN:JMLR16,CycleGAN2017}, or scene segmentation \cite{Tsai_adaptseg_2018,Liu_2021_CVPR,Saha_2021_CVPR}, overlooking the problem of scene recognition. 

With this work, we investigate for the first time a setting that combines three keywords: \emph{scene recognition}, \emph{multi-modal learning} and \emph{domain adaptation}. Our contributions can be summarized as follows:
\begin{itemize}[leftmargin=*]
\vspace{-2mm}\item We introduce a benchmark testbed\footnote{Dataset and code available at: \url{https://github.com/silvia1993/Multi-Modal_RGB-D_Scene_Recognition_Across_Domains}} for a novel unsupervised domain adaptation problem. We revisited the SUN RGB-D \cite{sunrgbd} dataset, identifying a subset of scene classes shared among four different 3D cameras. Each camera is considered as an RGB-D domain and we get an experimental framework with five multi-modal domain pairs (source RGB-D, target RGB-D).
\vspace{-2mm}\item We conduct a thorough study on state of the art methods originally developed to deal with only one or two of the considered keywords. Specifically we evaluate (a) the robustness of two \emph{multi-modal scene recognition} models on the proposed cross-domain scenario \cite{du2019translate,Ayub_2020_BMVC}; (b) the effect of several single-modal \emph{domain adaptation} approaches when extended on using multiple modalities for scene recognition \cite{AFN,Ganin:DANN:JMLR16,CycleGAN2017}; (c) the performance on scene recognition of a very recent \emph{multi-modal domain adaptation} approach originally developed for object classification \cite{LoghmaniRPPCV20}.  
\vspace{-2mm}\item Inspired by \cite{du2019translate}, we present a method able to exploit inter-modal translation to adapt across domains that we name \emph{Translate-to-Adapt}. Learning to generate the depth images from its RGB twin and vice-versa is a self-supervised task that can run both on the labeled source and on the unlabeled target data. We exploit both modality translation directions as auxiliary objectives in an end-to-end classification model, obtaining promising results across domains.
\end{itemize}

\section{Related Work}
\noindent\textbf{Multi-Modal Scene Recognition}
How to combine RGB and depth images for recognition task is an open question that has attracted a lot of attention in the machine learning and robotics community in the last years. In particular, multi-modal scene recognition research has rapidly evolved from models based on handcrafted features \cite{banica2015scenerecog, gupta2014scenerecog} to multi-layered networks able to learn the representation from a large amount of data \cite{wang2016scenerecog, gupta2016scenerecog, song2018scenerecog}. Fusing the modalities at \emph{input} level has been one of the first adopted solutions, with D considered as an extra image channel together with RGB \cite{couprie_ICLR13}. Some work also proposed \emph{output} score fusion techniques \cite{ChengCLZH17}. However, the largest part of the developed methods are based on multi-modal \emph{mid-level} feature combination strategies \cite{sunrgbd,song2018scenerecog,Wang_TM2015}. Recently the feature fusion approaches have been enriched with techniques that better capture the cross-modal relation, identifying both their correlative and distinct features with solutions ranging from CCA \cite{AAAI_df2} to the introduction of cross-modal graph convolution \cite{Yuan_Xiong_Wang_2019} and clustering \cite{Ayub_2020_BMVC}. Translate-to-Recognize \cite{du2019translate} belongs to this last group of methods and adopts an explicit translation from RGB to depth and vice-versa. The two-directional mappings are trained separately and combined only in a second stage with a scene classification model learned on pre-extracted features.

\noindent\textbf{Domain Adaptation}
The performance of a learning model naturally drops when training and testing data come from different distributions. Unsupervised Domain Adaptation is an extensively explored strategy to address this problem and it focuses on how to transfer knowledge learned from a labeled dataset (source domain) to another unlabeled one (target domain), whose data are available at training time \cite{csurka_book}.
In the most recent domain adaptation literature we can identify three main solutions.  \textit{Discrepancy-based} methods \cite{long2015learning,sun2016return,AFN} measure and minimize the distance between source and target distributions acting at feature level.
\textit{Adversarial learning} techniques \cite{Ganin:DANN:JMLR16,tzeng2017adversarial,russo2018source,CycleGAN2017} train a generator and a domain discriminator adversarially so that the optimal solution is the one in which the generator produces target features indistinguishable from those of the source. 
The last and more recent research line comprises the approaches that enhance the generalization ability of the network by introducing an auxiliary \textit{self-supervised} task \cite{PAMIselfsup,xu2019self}. The unlabeled target data can be used to optimize the self-supervised objective which helps to produce a robust representation for the main supervised task.

\noindent\textbf{Multi-Modal Domain Adaptation} Most of the existing domain adaptation works consider single modality images. The main focus is on RGB data, with only few efforts made to investigate the domain shift across depth images \cite{Patricia_2017_ICCV}, or considering both RGB and D modalities. In the last case, the proposed approaches either identify RGB as source and D as target \cite{SpinelloA12,hoffman2016cross-modal}, or deal with a multi-modal source (RGB-D) and a single-modal (RGB) target \cite{Li2017}, or simply use the depth information as an additional input channel for source and target, extending standard RGB domain adaptation methods to the RGB-D case \cite{wang2019unsupervised,bousmalis2017unsupervised}. Only recently Loghmani \etal \cite{LoghmaniRPPCV20} highlighted the importance of exploiting the inter-modal relation for adaptive learning. They proposed to predict the relative rotation between the RGB and its twin D image: since this task does not need sample annotation can run both on the labeled source and unlabeled target, helping to learn a domain agnostic representation. This approach was designed for object classification and does not seamlessly apply to scene recognition where the rotation task can be solved by exploiting shortcuts based on not semantically meaningful cues (\eg uniform pavement and ceiling), resulting in low accurate scene prediction.

As it is clear from the cited literature, no previous work focused on learning robust multi-modal scene recognition models across domains. Here we propose the task, we define its experimental testbed and a first learning approach that exploits self-supervised modality translation.

\section{Dataset}
\begin{table}[t]
\centering
\resizebox{\columnwidth}{!}{
\begin{tabular}{lcccc}
\hline
 Class name & Kinect v1 & Kinect v2 & Realsense & Xtion\\
 \hline
0. bathroom & 147 & 150 & 67 & 260\\
1. bedrooom & 442 & 121 & 0 & 521\\
2. classroom & 49 & 535 & 73 & 366 \\
3. computer\_room & 6 & 65 & 40 & 68\\
4. conference\_room & 5 & 69 & 53 & 163\\
5. dining\_area & 0 & 192 & 125 & 80\\
6. discussion\_area & 6 & 62 & 30 & 103\\
7. kitchen & 291 & 86 & 20 & 183\\
8. office & 295 & 418 & 46 & 287\\
9. rest\_space & 6 & 407 & 285 & 226\\
\hline
Total & 1247 & 2105 & 739 & 2257 \\
\hline
\end{tabular}
}
\caption{Number of images in the considered classes.}
\label{dataset}
\vspace{-2mm}
\end{table}
\begin{figure}[t!]
    \centering
\includegraphics[width=0.48\textwidth]{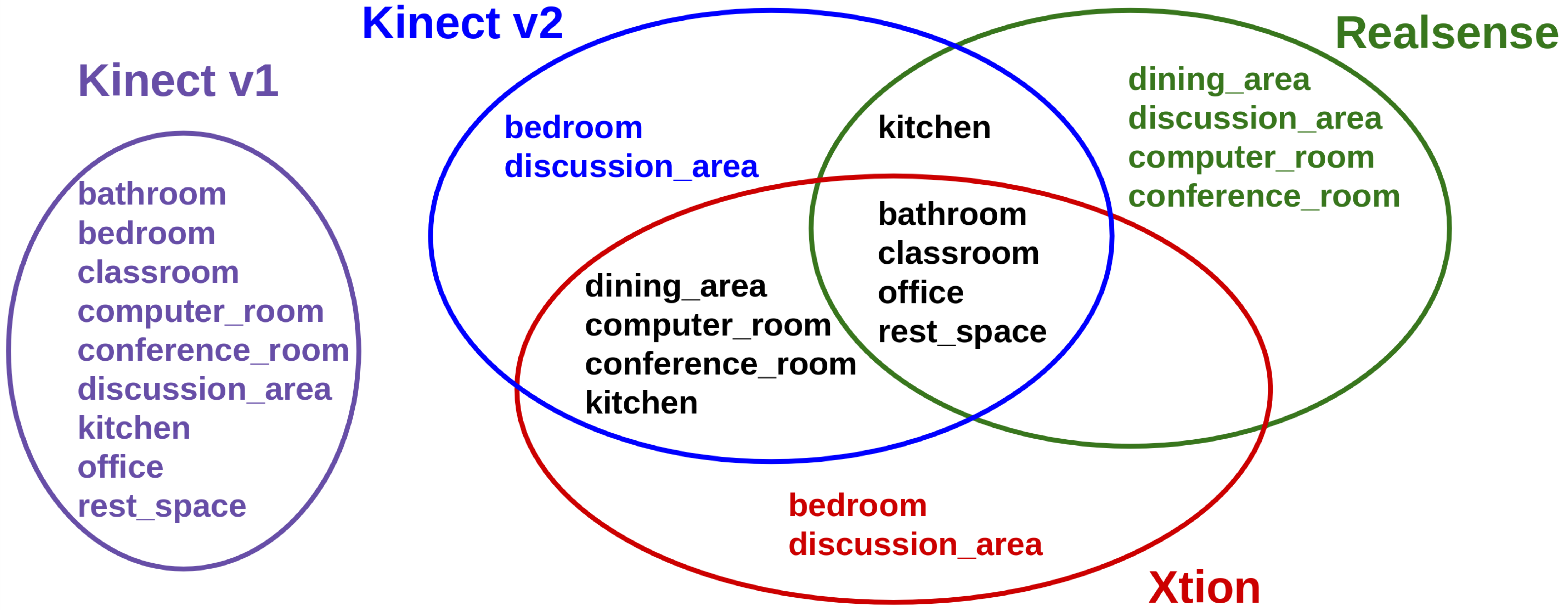}
    \caption{Physical place overlapping. Some of the scene classes contain images of the exact same places taken with multiple cameras, while others are collected from different locations. We use the following color code: black indicates a class that contains images taken in the exact same place by multiple cameras, while blue/green/red/violet indicate classes with specific room images captured only with Kinect v2 / Realsense / Xtion / Kinect v1.}
    \label{fig:overlapping}\vspace{-3mm}
\end{figure}

\begin{figure*}[!t]
\hspace{-4mm}
\begin{tabular}{c@{~}c@{~}c@{~}c@{~}c@{~}c@{~}c@{~}c@{~}c@{~}c@{~}c@{~}c}
 & \multirow{2}{*}[-0.04in]{\includegraphics[width=0.10\textwidth]{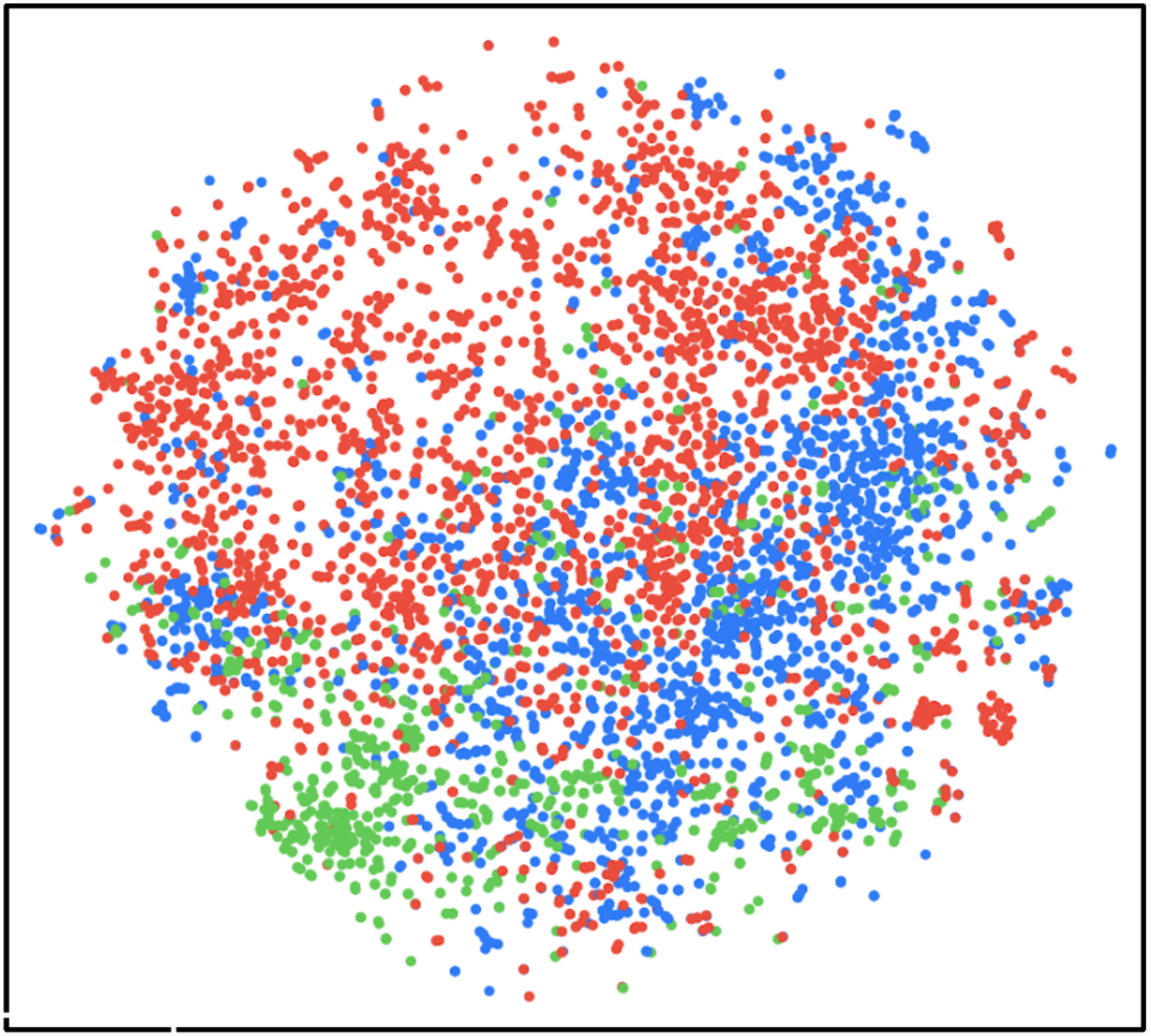}} &
\scriptsize{bathroom} & \scriptsize{bedroom} & \scriptsize{classroom} & \scriptsize{computer\_room} & \scriptsize{conference\_room} 
& \scriptsize{dining\_area} & \scriptsize{discussion\_area} & \scriptsize{kitchen} & \scriptsize{office} & \scriptsize{rest\_space} \\
{\rotatebox[origin=c]{90}{\qquad\qquad\small{RGB}}} & & \includegraphics[width=0.08\textwidth]{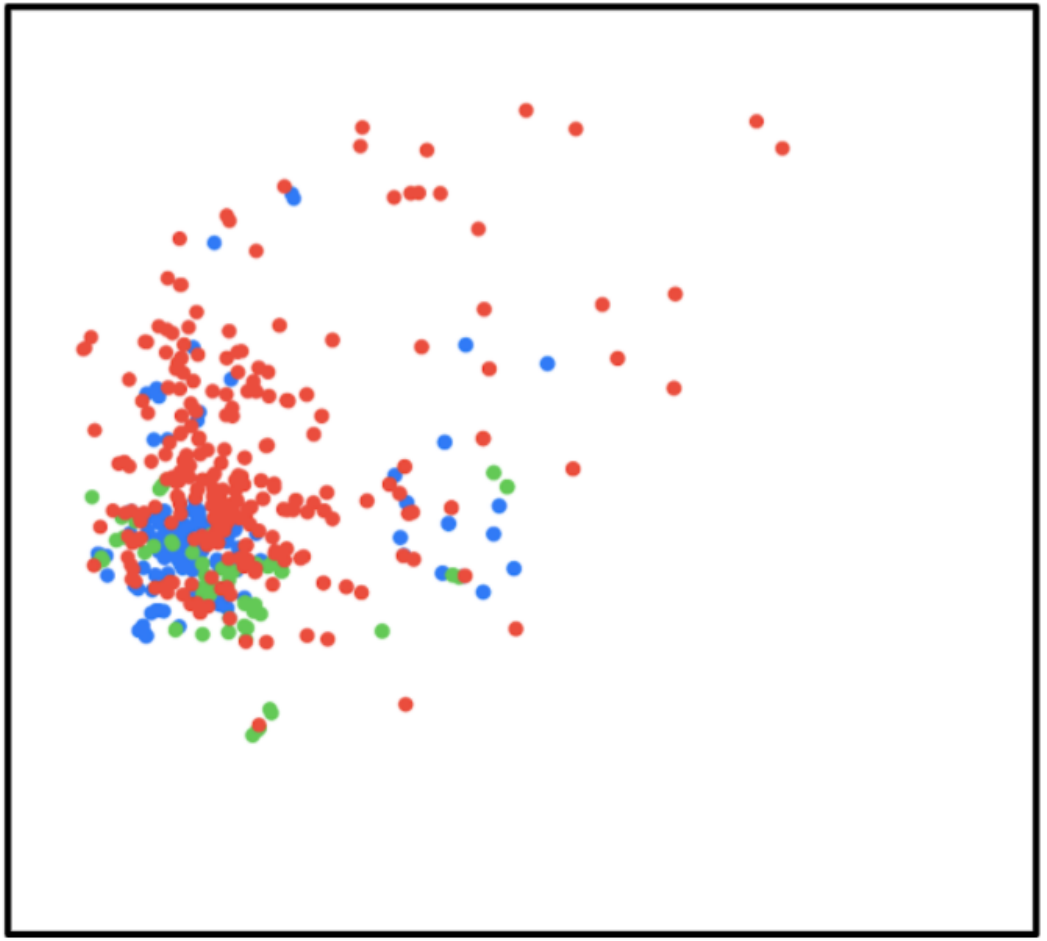} & \includegraphics[width=0.08\textwidth]{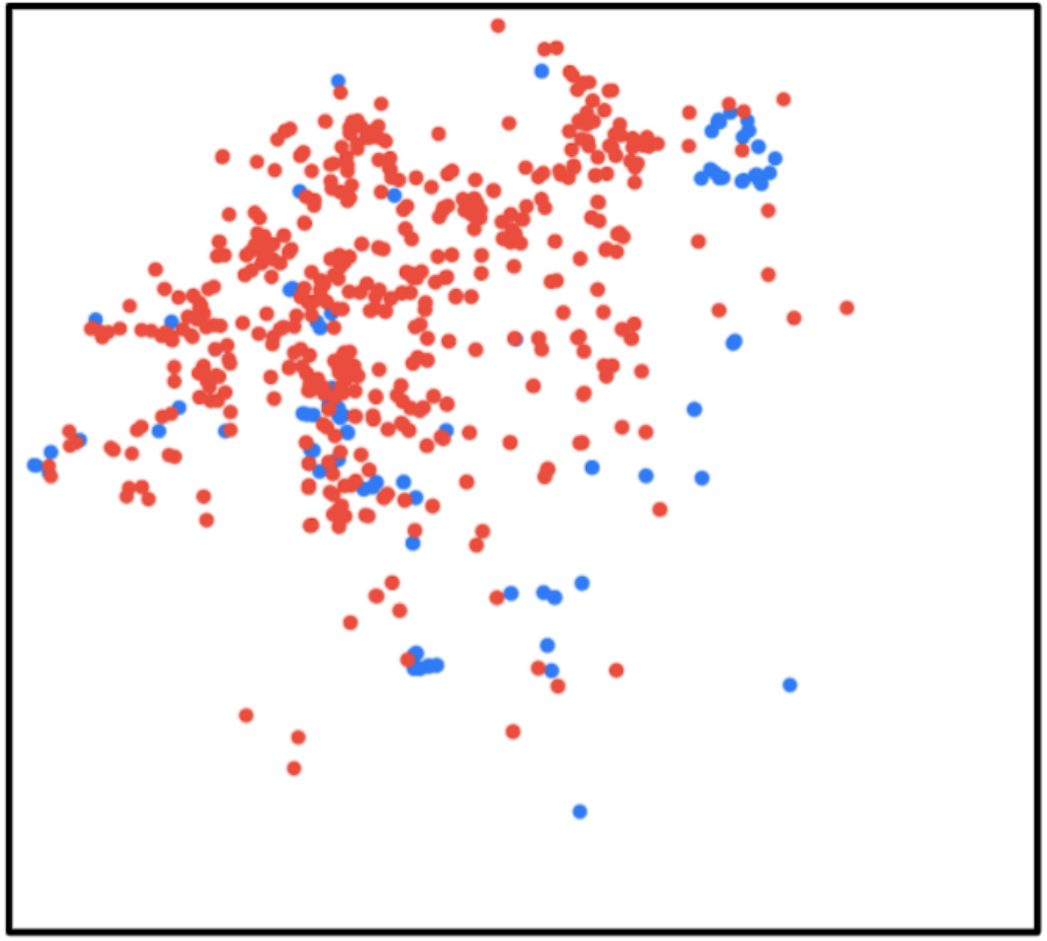} & \includegraphics[width=0.08\textwidth]{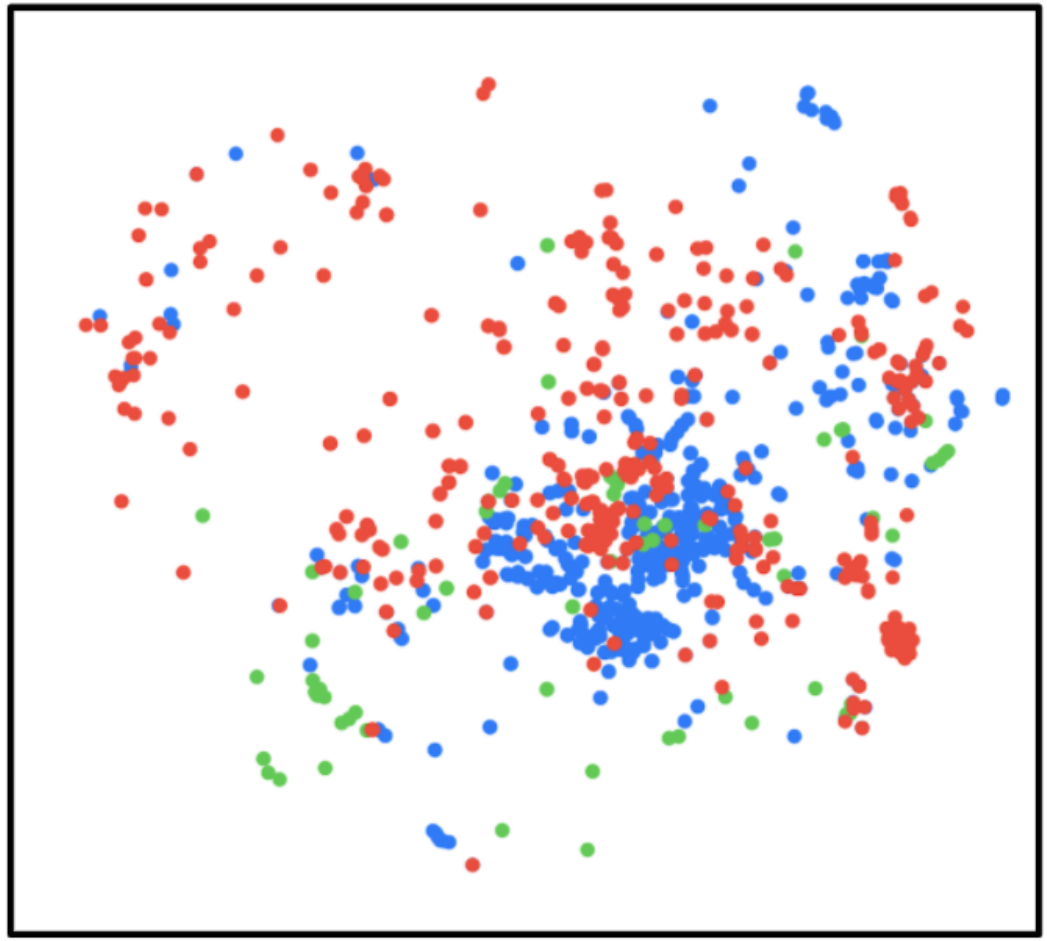}& \includegraphics[width=0.08\textwidth]{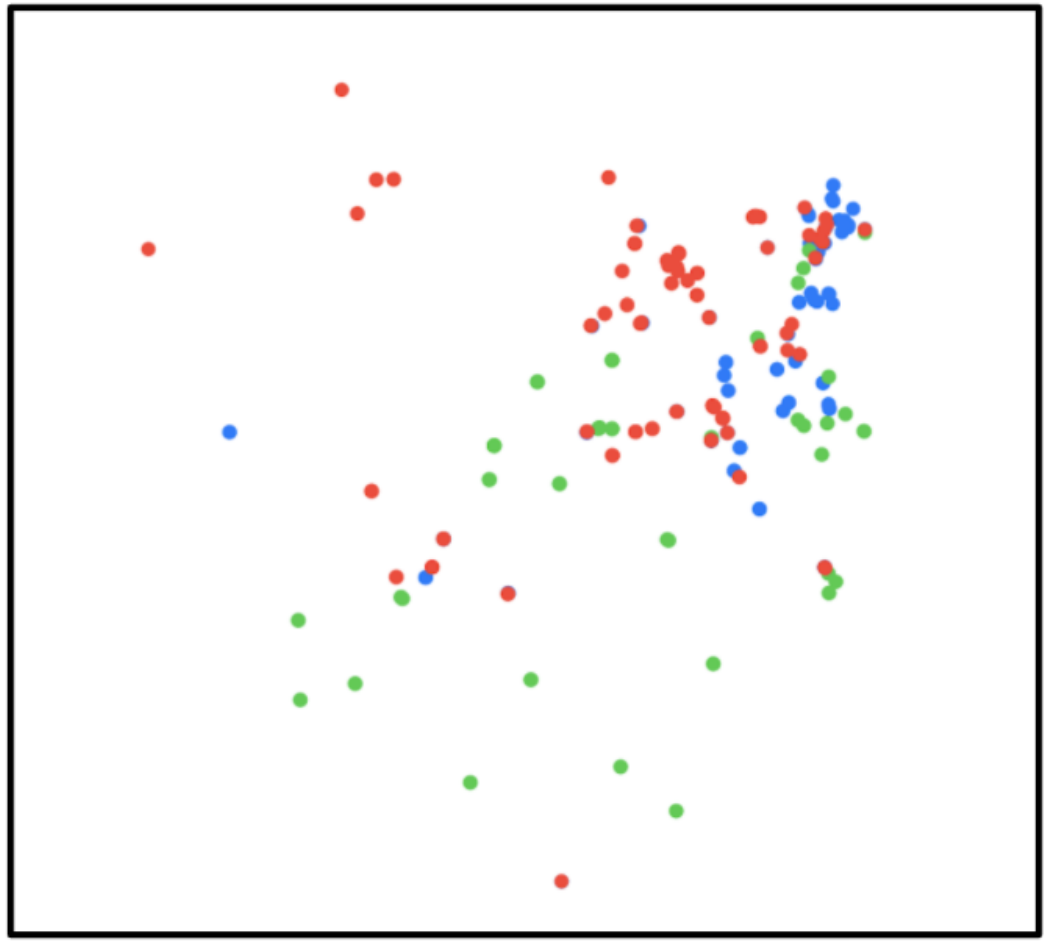} & \includegraphics[width=0.08\textwidth]{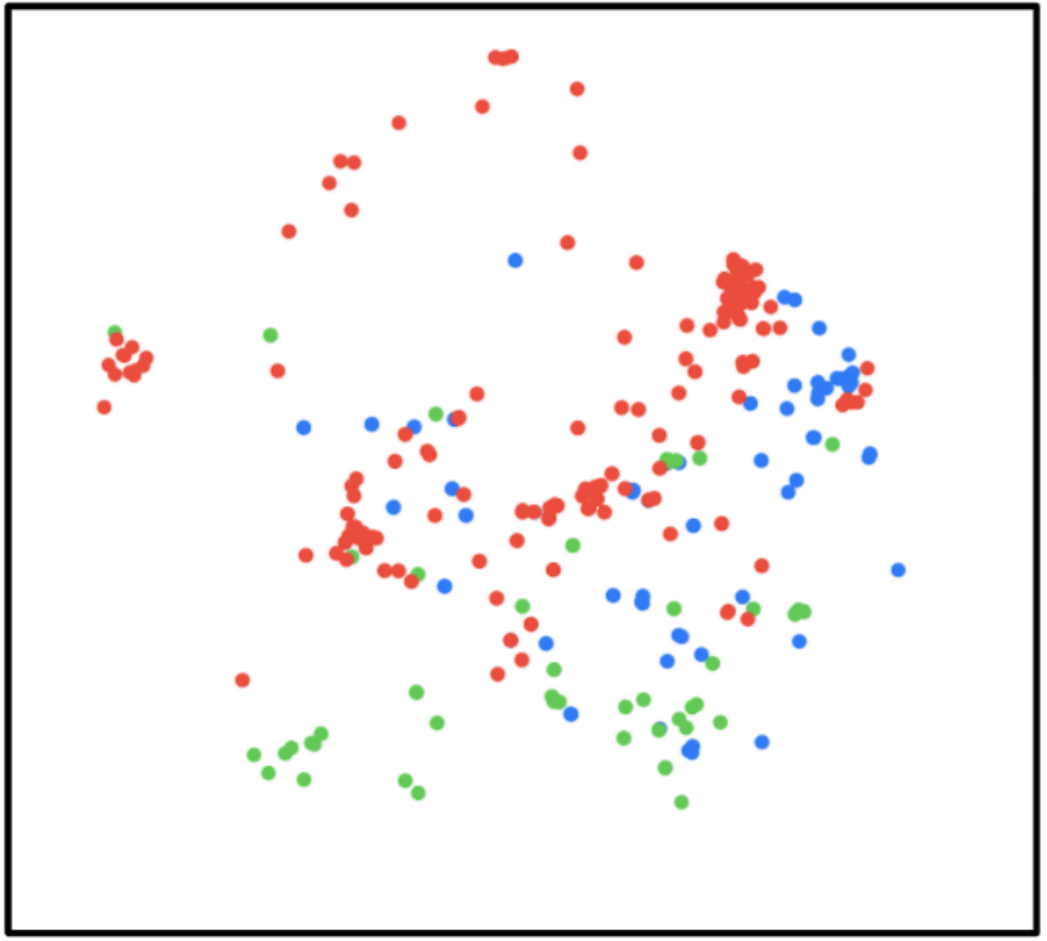} & \includegraphics[width=0.08\textwidth]{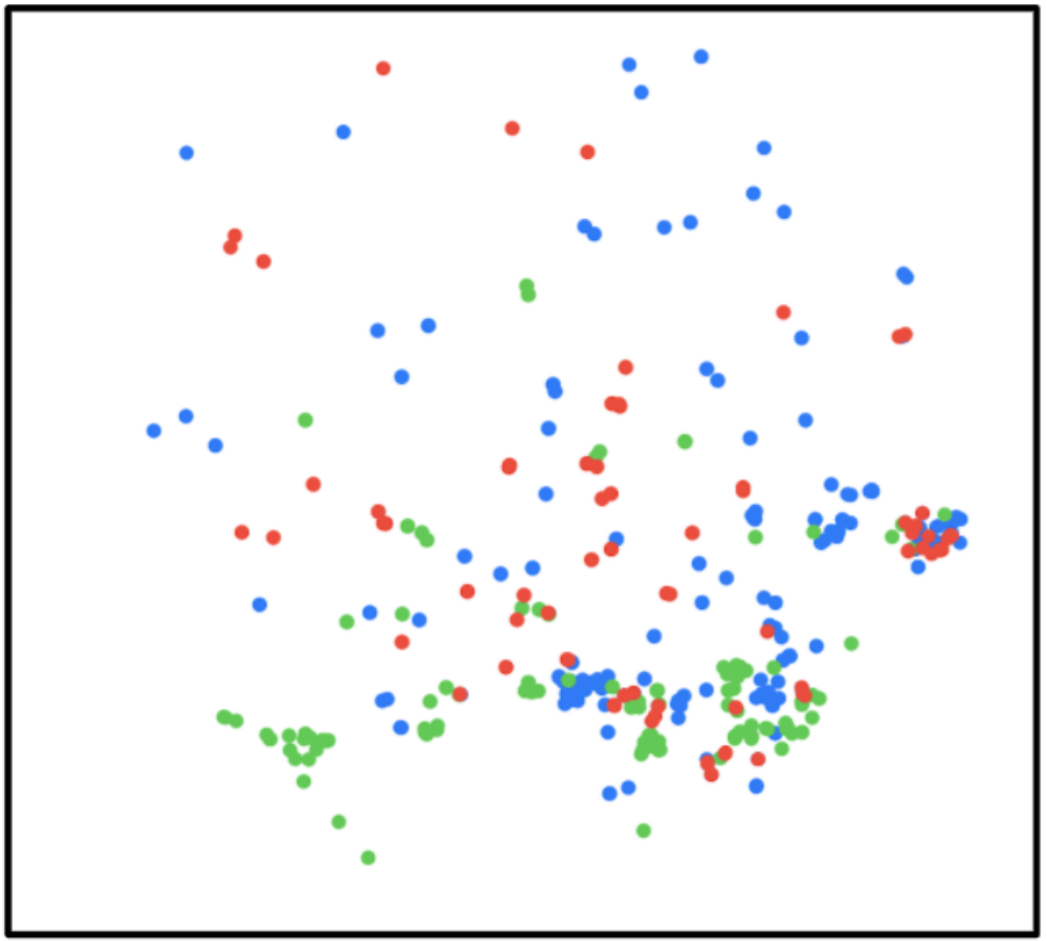} & \includegraphics[width=0.08\textwidth]{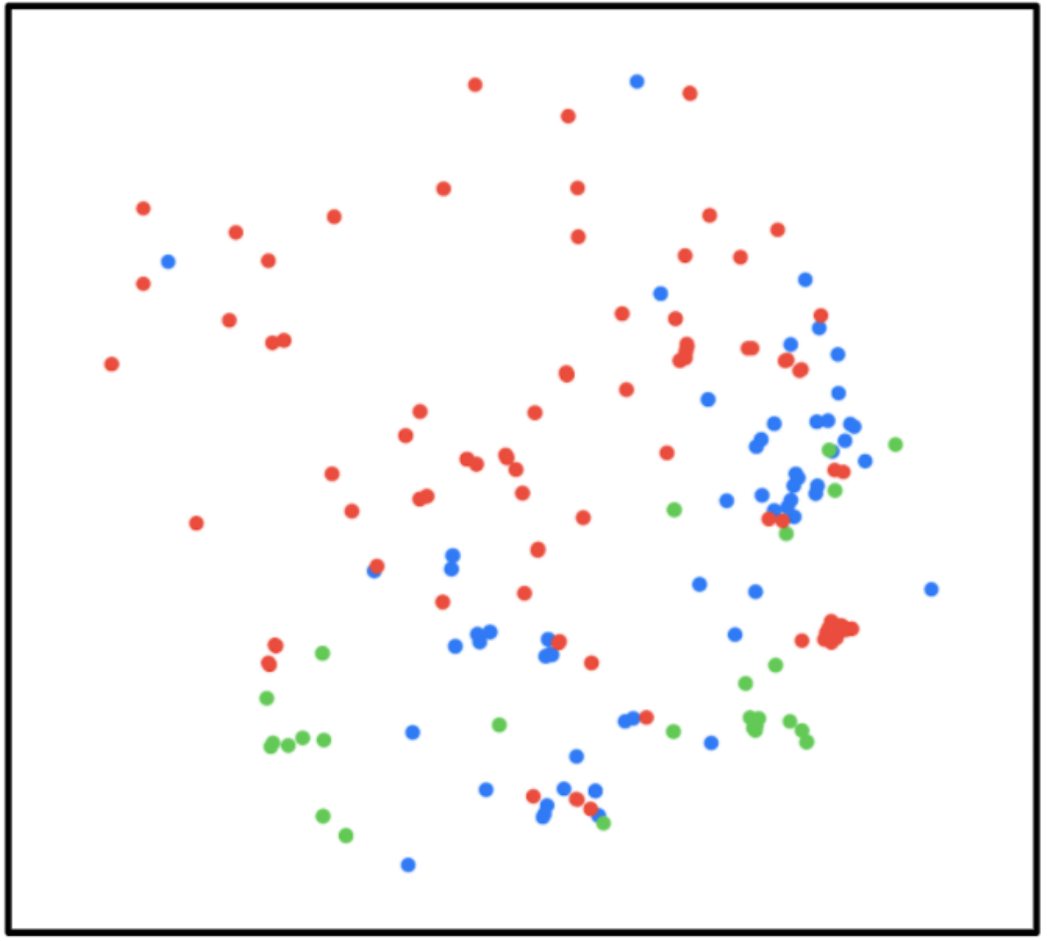} & \includegraphics[width=0.08\textwidth]{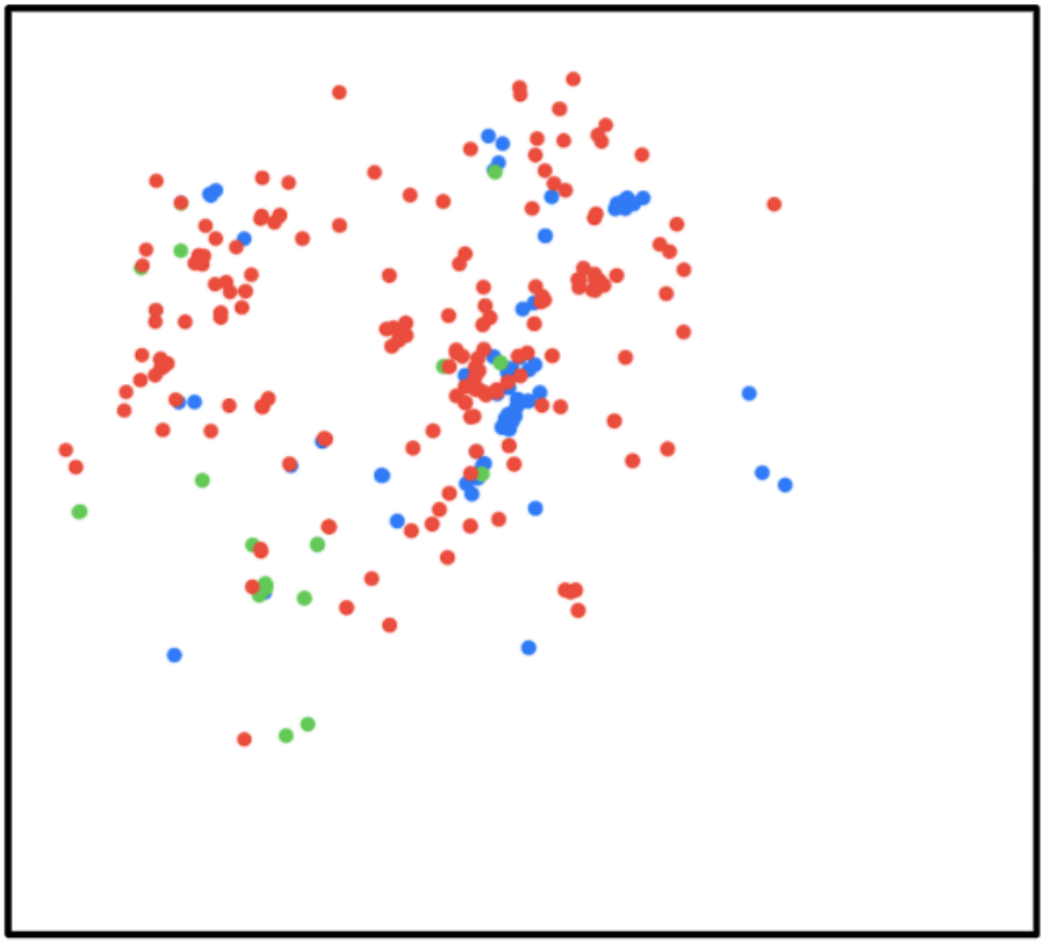} & \includegraphics[width=0.08\textwidth]{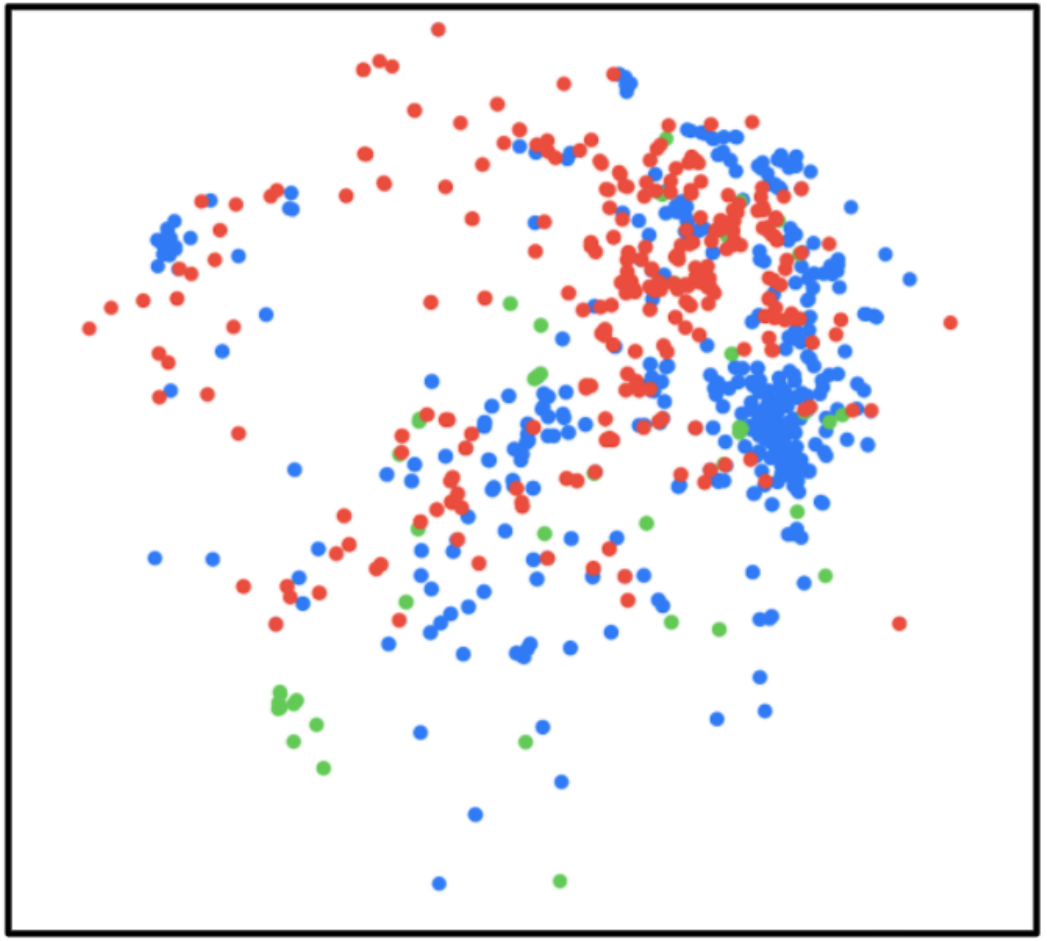}& \includegraphics[width=0.08\textwidth]{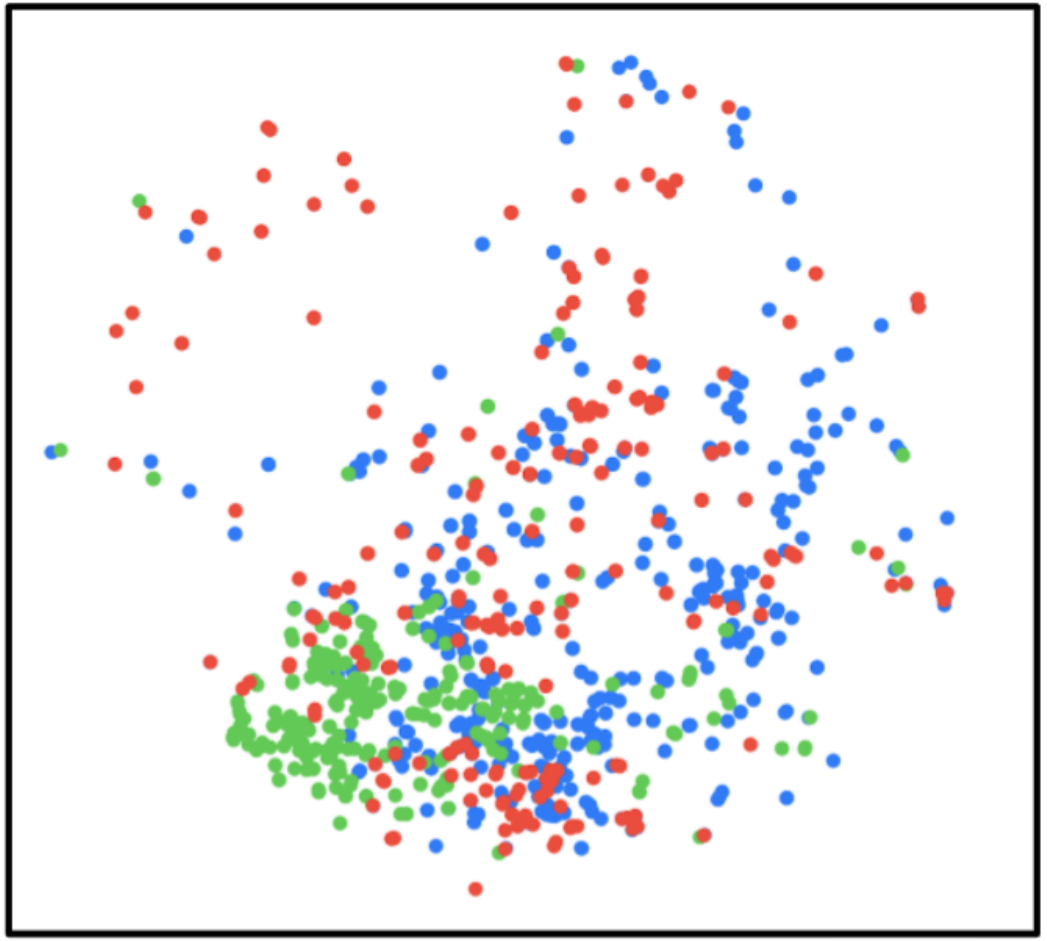} \vspace{-8mm}\\
 & \multirow{2}{*}[-0.04in]{\includegraphics[width=0.10\textwidth]{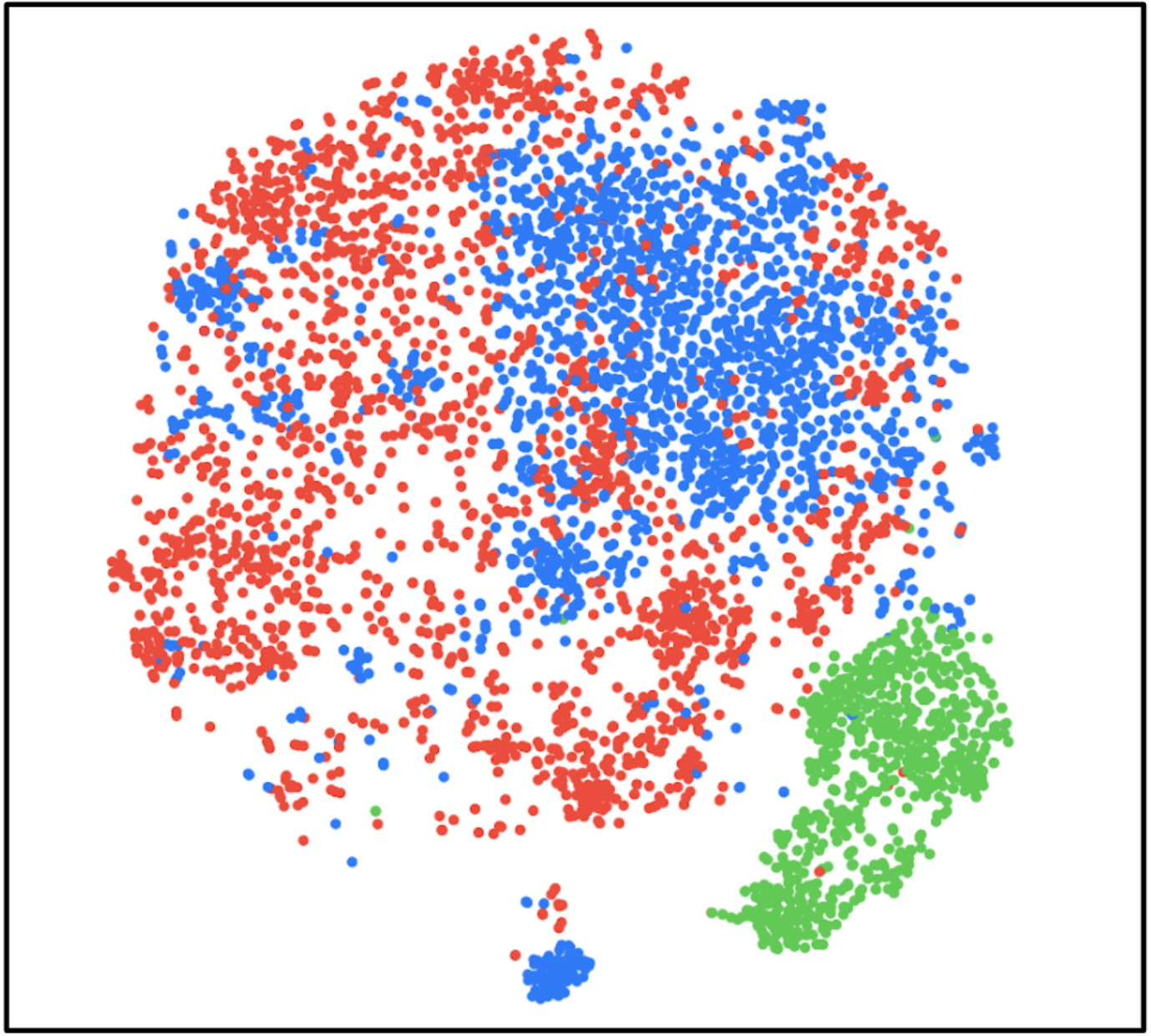}} &
\scriptsize{bathroom} & \scriptsize{bedroom} & \scriptsize{classroom} & \scriptsize{computer\_room} & \scriptsize{conference\_room} 
& \scriptsize{dining\_area} & \scriptsize{discussion\_area} & \scriptsize{kitchen} & \scriptsize{office} & \scriptsize{rest\_space} \\
{\rotatebox[origin=c]{90}{\qquad\qquad\small{Depth}}} & & \includegraphics[width=0.08\textwidth]{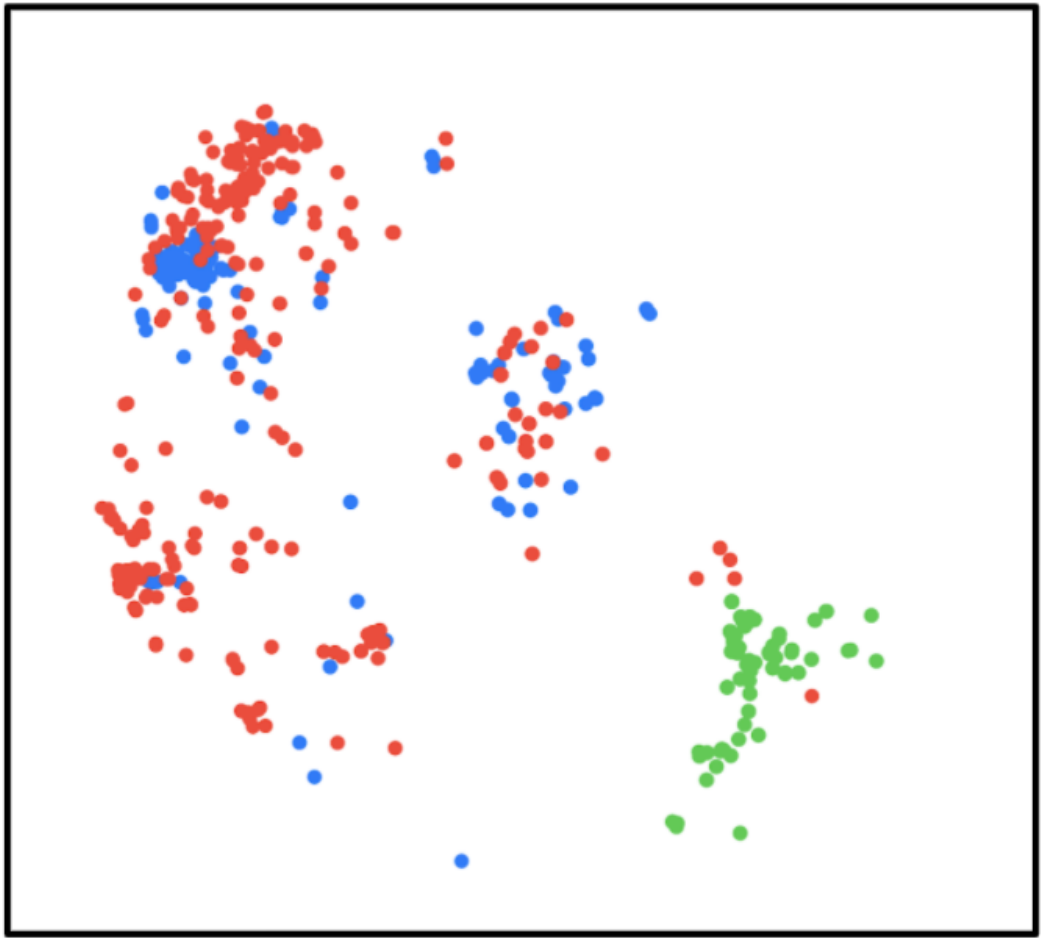} & \includegraphics[width=0.08\textwidth]{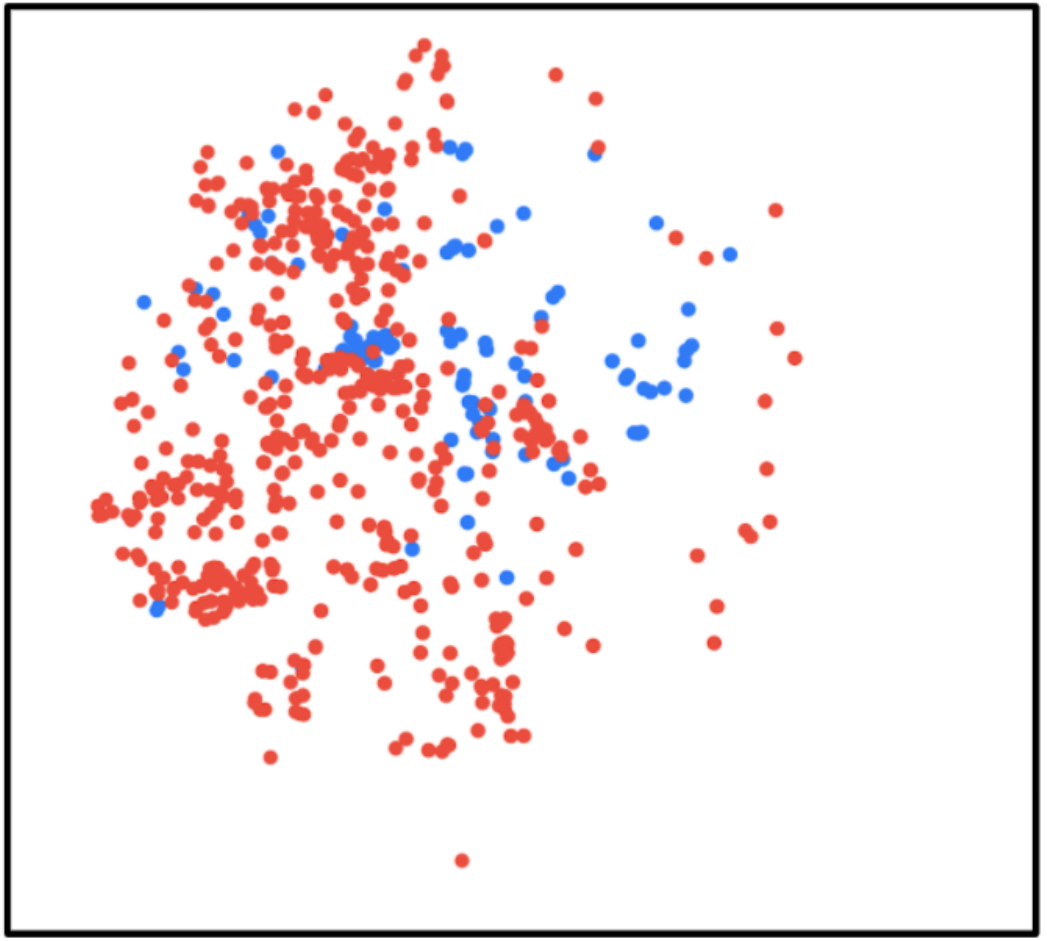} & \includegraphics[width=0.08\textwidth]{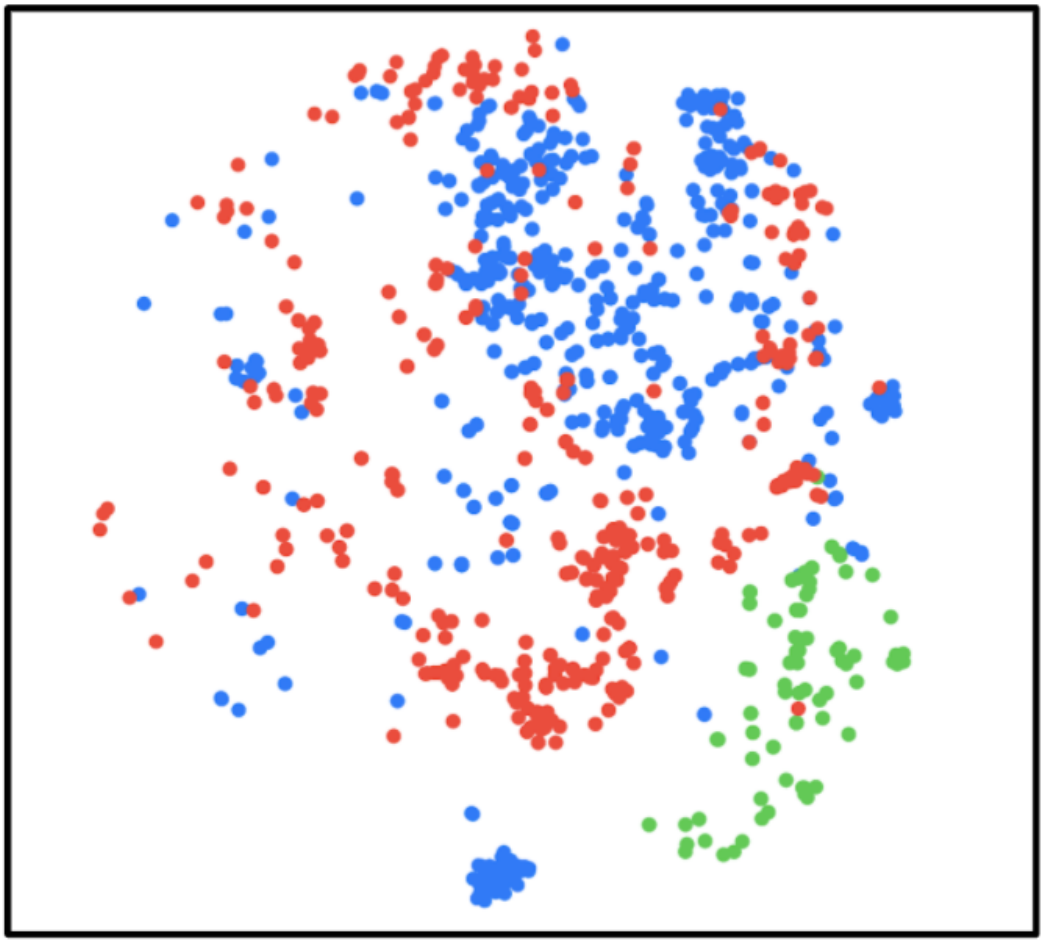}& \includegraphics[width=0.08\textwidth]{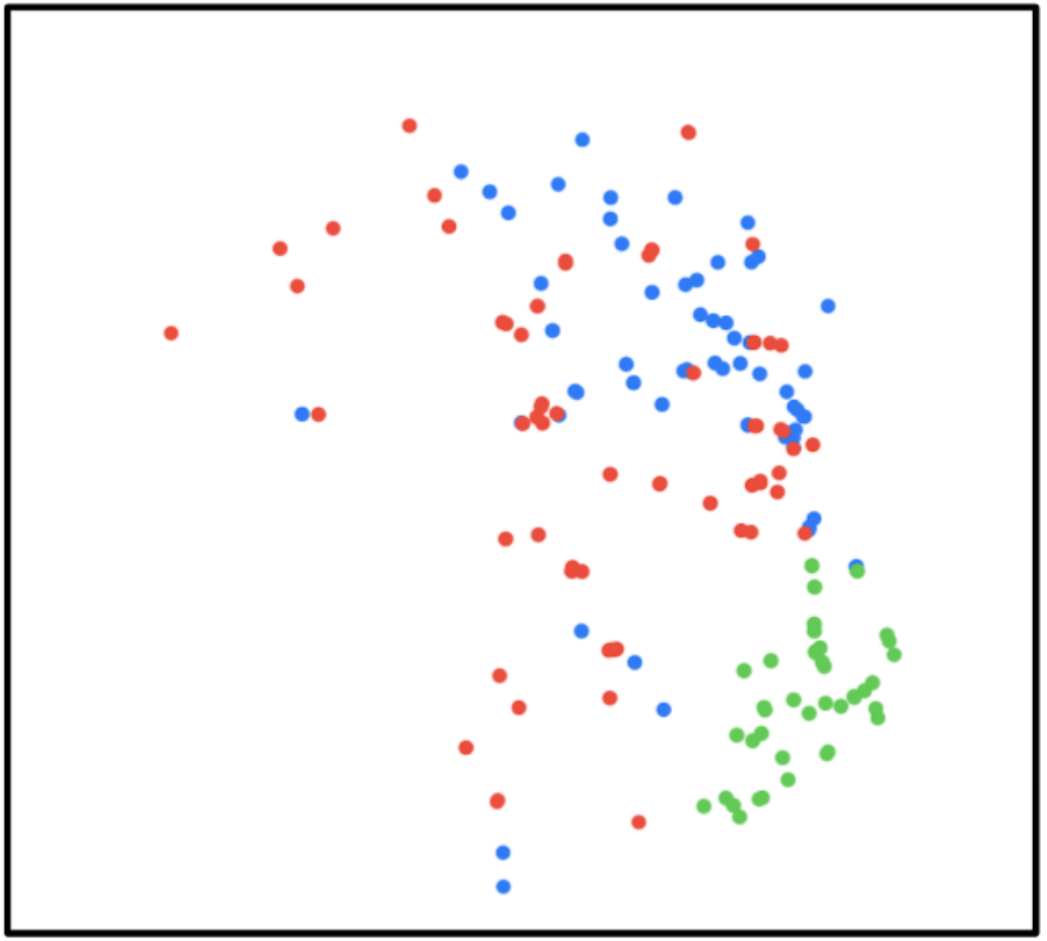} & \includegraphics[width=0.08\textwidth]{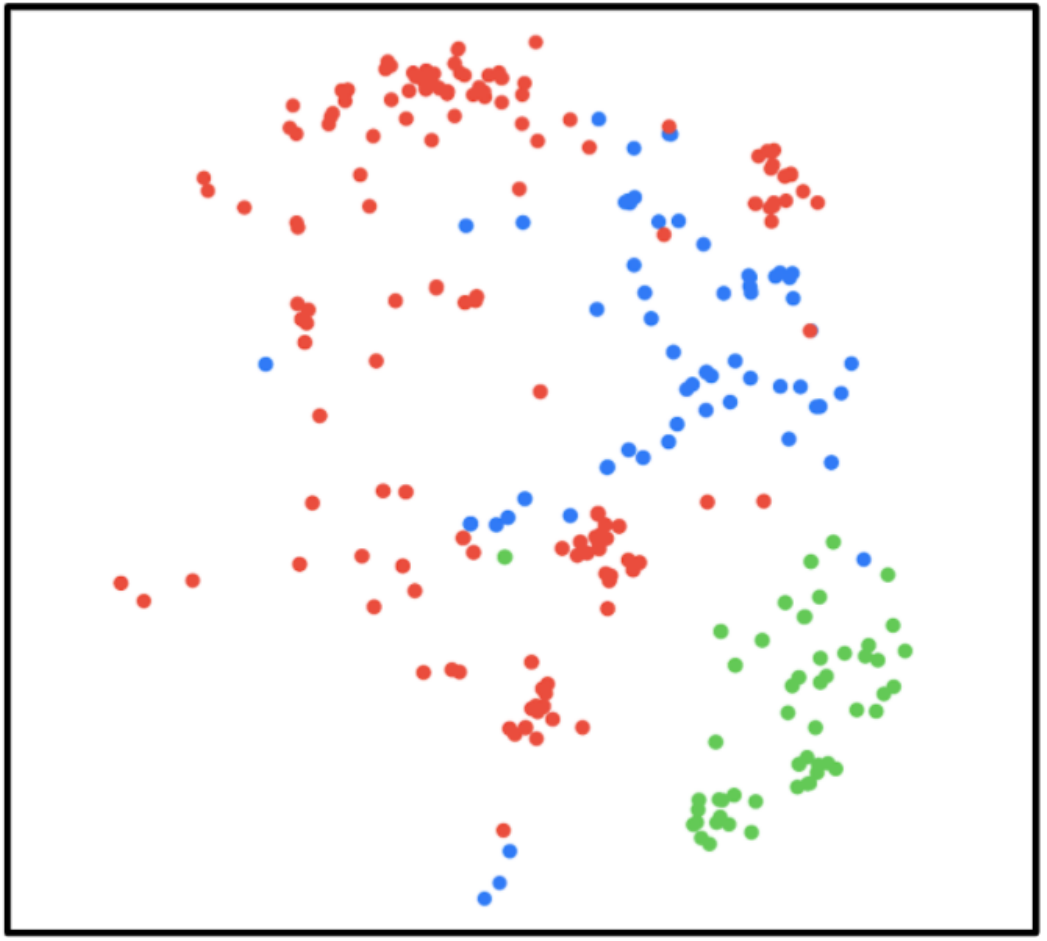} & \includegraphics[width=0.08\textwidth]{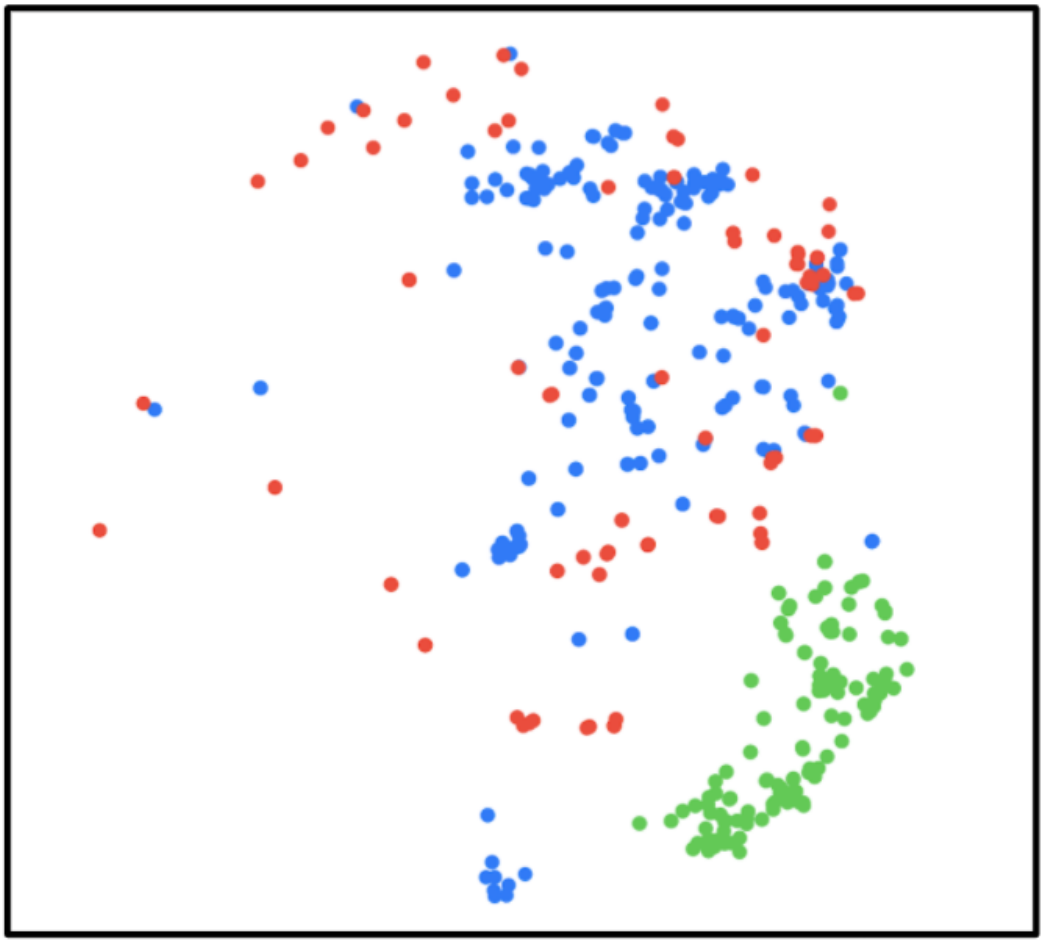} & \includegraphics[width=0.08\textwidth]{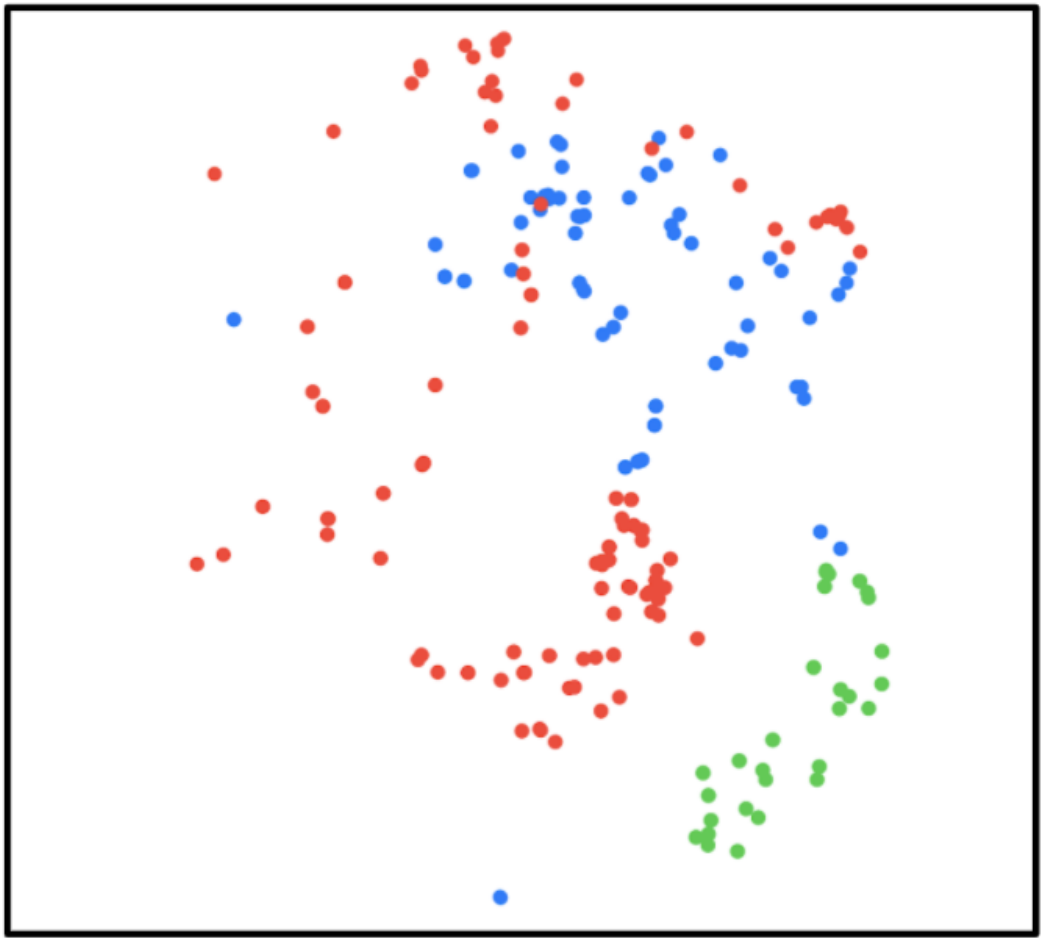} & \includegraphics[width=0.08\textwidth]{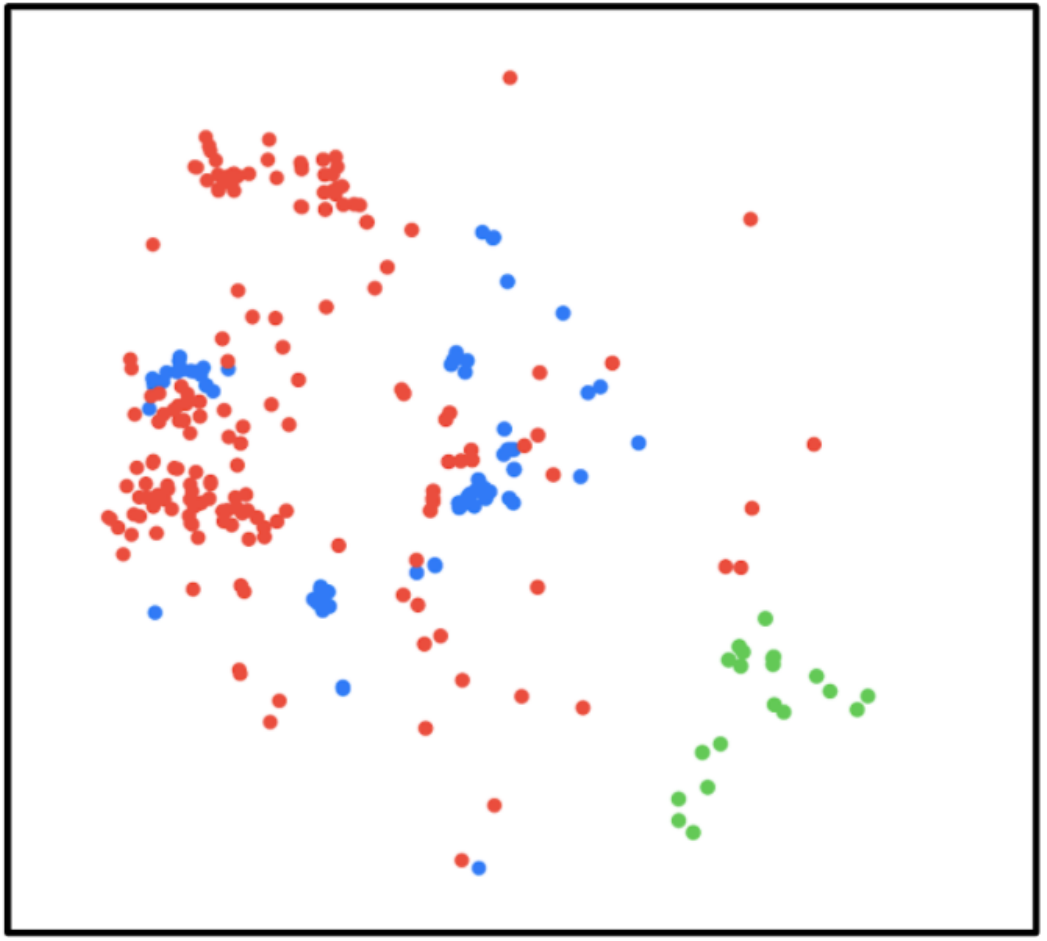} & \includegraphics[width=0.08\textwidth]{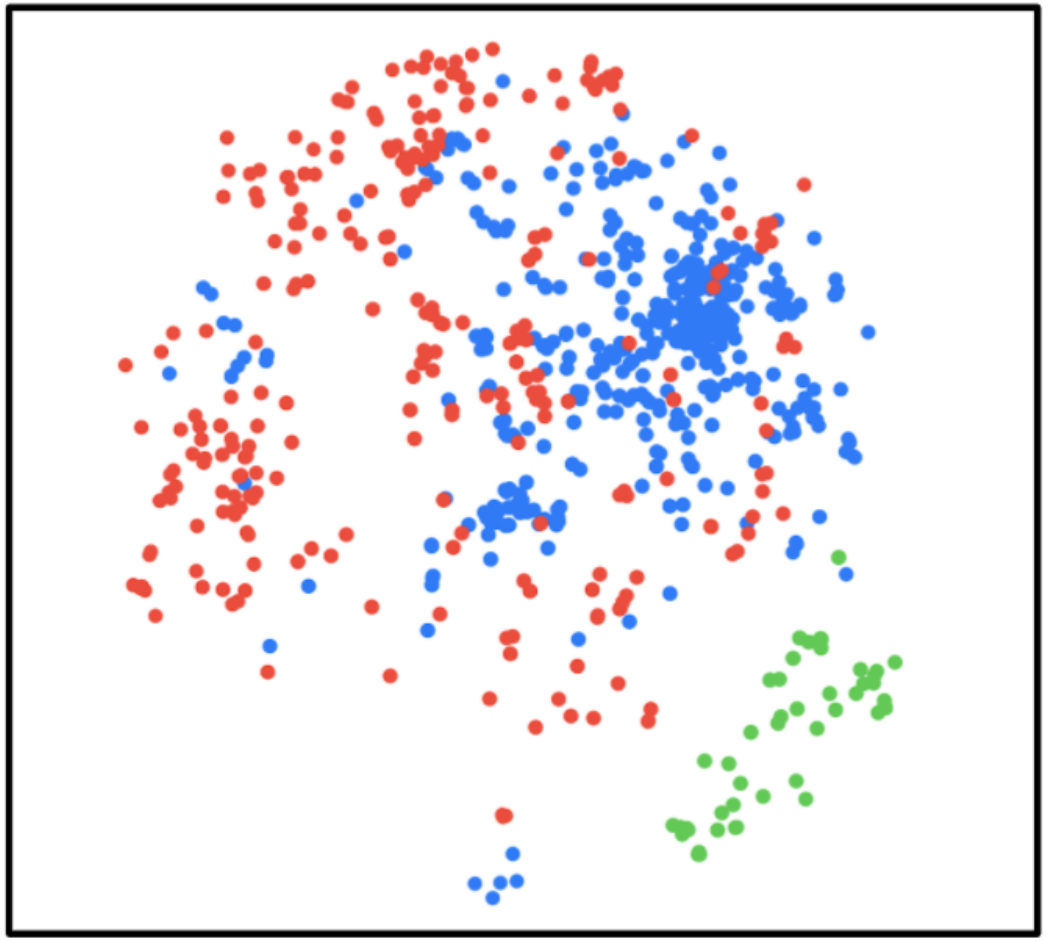}& \includegraphics[width=0.08\textwidth]{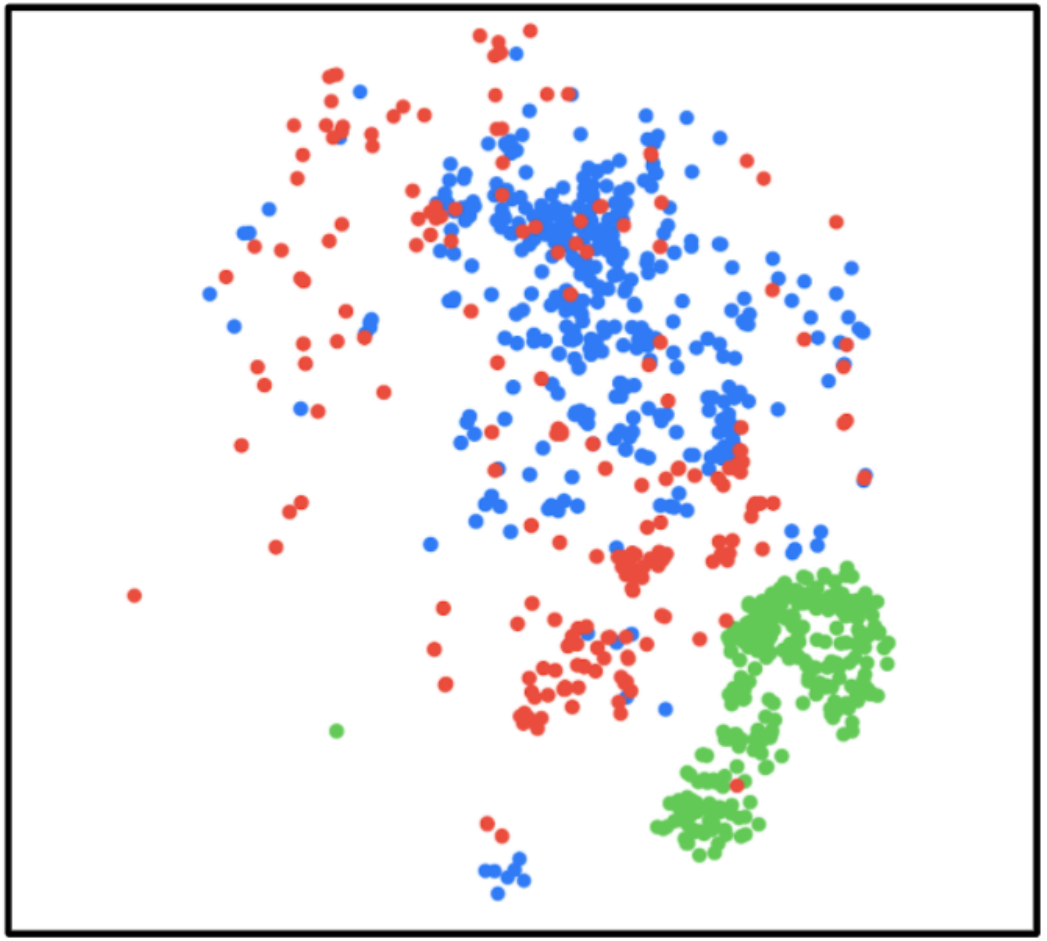}\vspace{-8mm}\\
\end{tabular}
\vspace{-3mm}\caption{Tsne \cite{tsne} visualization of the three domains of our multi-modal cross-domain scene classification testbed. Each domain is composed by images of a different camera: Kinect v2 (blue),  Realsense (green), Xtion (red).}
\label{fig:tsne}\vspace{-3mm}
\end{figure*}

The largest existing multi-modal cross-domain scene data collection, SUN RGB-D \cite{sunrgbd}, contains 3784 Microsoft Kinect v2 images, 3389 Asus Xtion images, 2003 Microsoft Kinect v1 images,  and 1159 Intel RealSense images. 

The Asus Xtion, as well as the Kinect v1, belong to the family of near-IR light pattern cameras. The raw depth maps from both sensors have low noise but an observable quantization effect \cite{sunrgbd}.

The time-of-flight based Kinect v2 has the largest field of view among the considered cameras. The raw depth map is less smooth than the structured light sensors and it may fail more frequently for black objects and slightly reflective  surfaces, but for ranges greater than 2 meters is more precise.
The Intel Realsense is  a  lightweight  and low-power consuming IR active stereo camera. Along with the Kinect v2, the RGB camera has the highest resolution of all tested cameras. However, its raw
depth is worse than that of other RGB-D sensors: failures of the stereo matching may lead to several artifacts and the 
effective range for reliable depth is shorter (depth gets very
noisy around 3.5 meters) \cite{sunrgbd,grenzdorffer2020ycb}. 

As shown in Figure \ref{fig:teaser} there is an ample variation in the appearance of the obtained images. This implies that a user who wants to leverage existing scene recognition models should pay particular attention in choosing one trained on images of the correct camera to avoid incurring in a significant drop in performance. To study in detail this domain shift, we searched for the scene classes shared among the four SUN RGB-D cameras and containing the largest amount of samples per class. To get a higher cardinality we merged the \emph{office\_kitchen} with the \emph{kitchen} class. The final collection subset is summarized in Table \ref{dataset}. 
Overall we have 10 classes, however the \emph{dining\_area} and the \emph{bedroom} are missing respectively for Kinect v1 and Realsense. 

Some of the scene classes contain the same physical location recorded with multiple cameras. We visualize the overlap in Figure \ref{fig:overlapping}. For instance, the same group of  office rooms has been visited with Kinect v2, Realsense, and Xtion cameras and the captured image constitutes the class \emph{office}. The class \emph{kitchen} has some instances recorded in the same places with Kinect v2 and Realsense, while others come from rooms shared between Xtion and Kinect v2. For the class \emph{discussion\_area}, each camera recorded images in different physical locations. Finally, the data captured with Kinect v1 do not share any location with the other cameras.

We decided to focus on the Kinect v2 (K) and  Xtion (X) to define a 10 class domain adaptation problem with both the camera used as source and target in turn. Moreover, due to its limited number of samples, we considered the Realsense (R) images only as target, with K, X, and their combination KX as source. 
Finally, we kept the Kinect v1 (Kv1) out of our current classification setting due to its severe class unbalance, but we will use it as a testbed for modality hallucination (see Section \ref{sec:exper}). In all the cases we employ geocentric HHA (Horizontal disparity, Height above ground and Angle with gravity) \cite{gupta2014hha} to encode depth images which has been shown to help in capturing the geometrical properties of depth data. 

\begin{table}[t]
\begin{center}
\begin{tabular}{c|cc||c|cc}
\hline
 & RGB & Depth & & RGB & Depth\\
 \hline
K $\rightarrow$ K & 77.09 & 72.09 & X $\rightarrow$ X & 79.98 & 72.79\\
K $\rightarrow$ X & 51.90 & 42.07 & X $\rightarrow$ K & 57.50 & 54.43\\
\hline
drop & 25.19 & 30.02 & drop &  22.48 & 18.36 \\
\hline 
\end{tabular}

\vspace{2mm}
\begin{tabular}{c@{}c}
\hspace{-5mm}\includegraphics[width=0.25\textwidth]{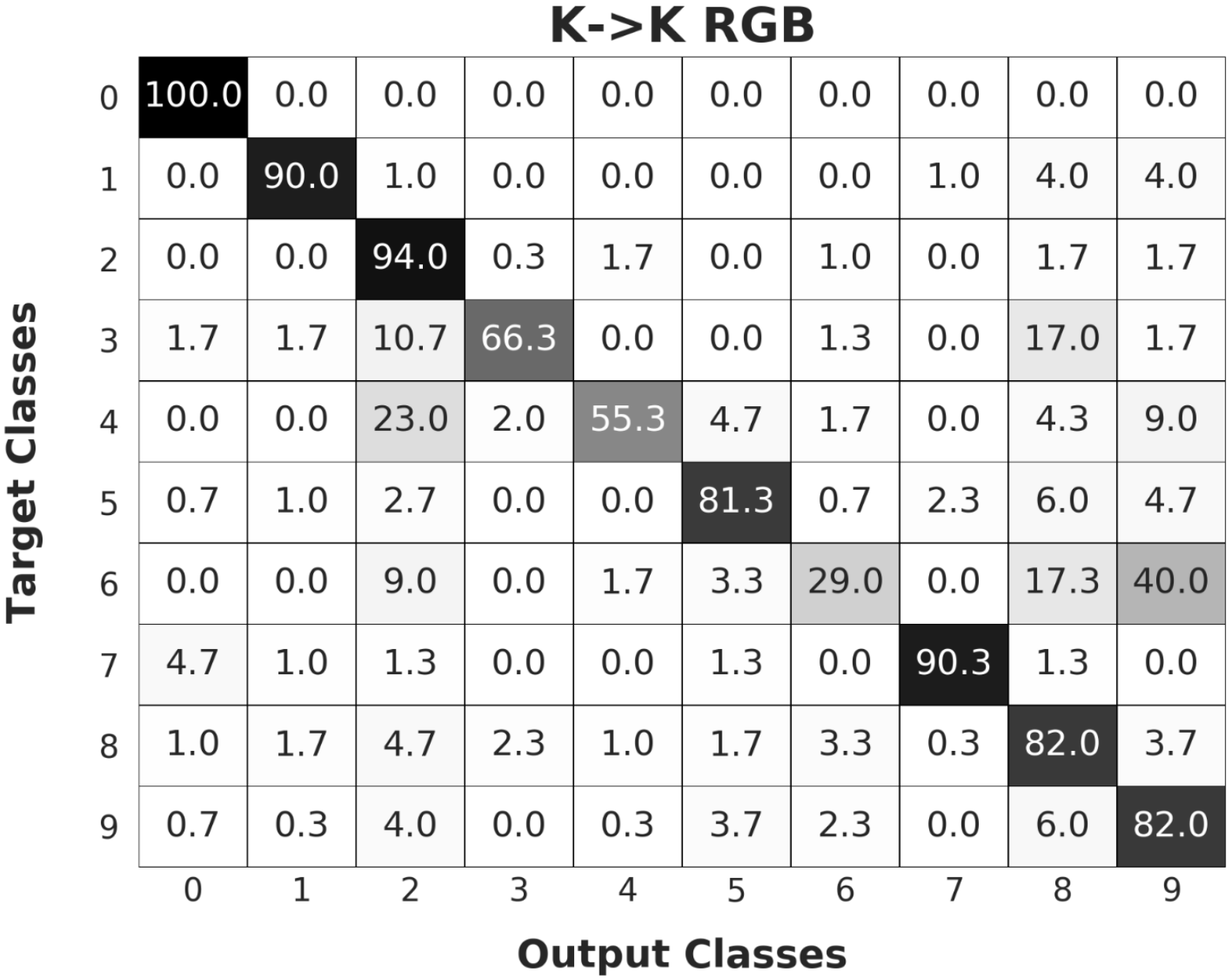} & \includegraphics[width=0.236\textwidth]{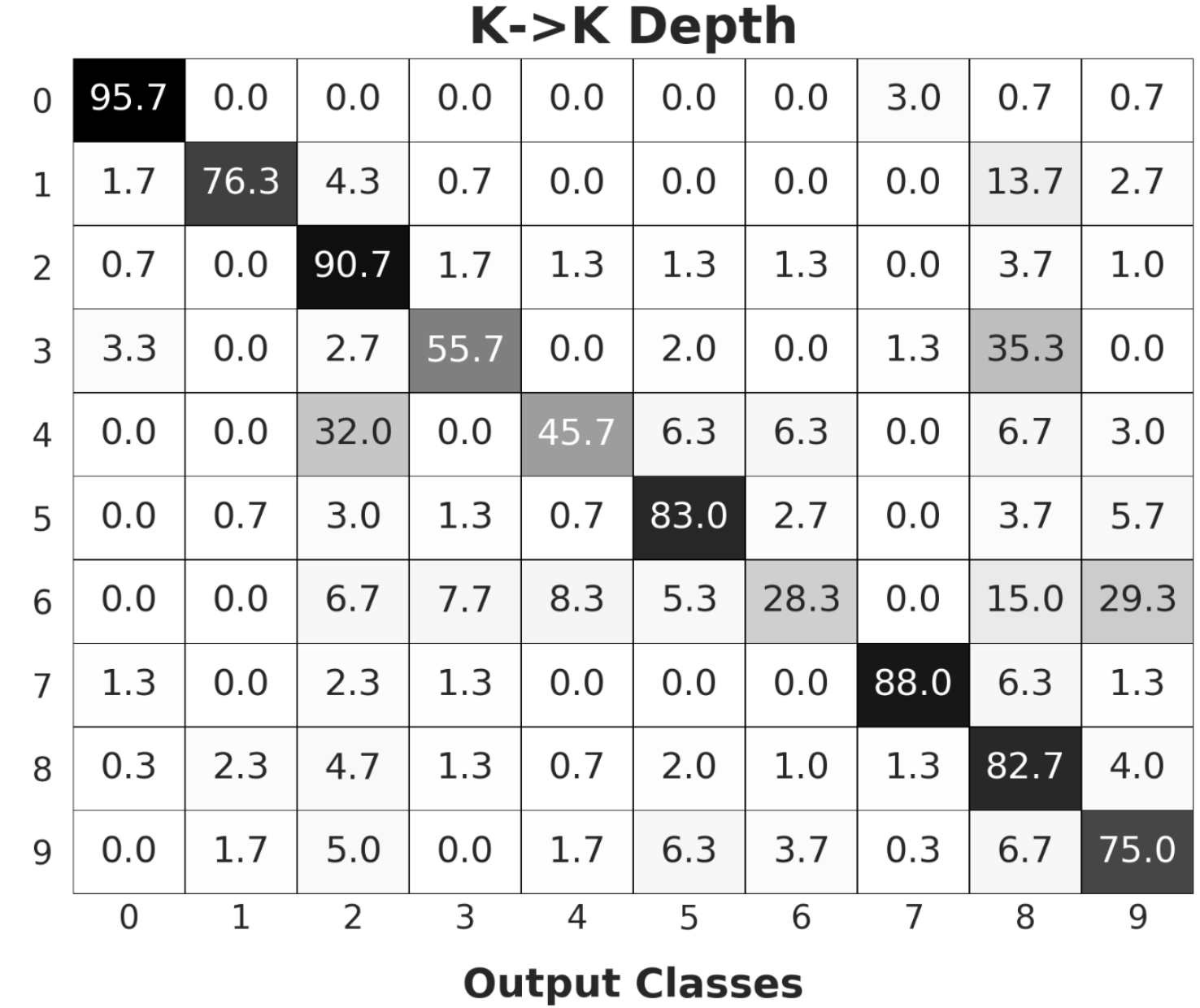} \\
\hspace{-5mm}\includegraphics[width=0.25\textwidth]{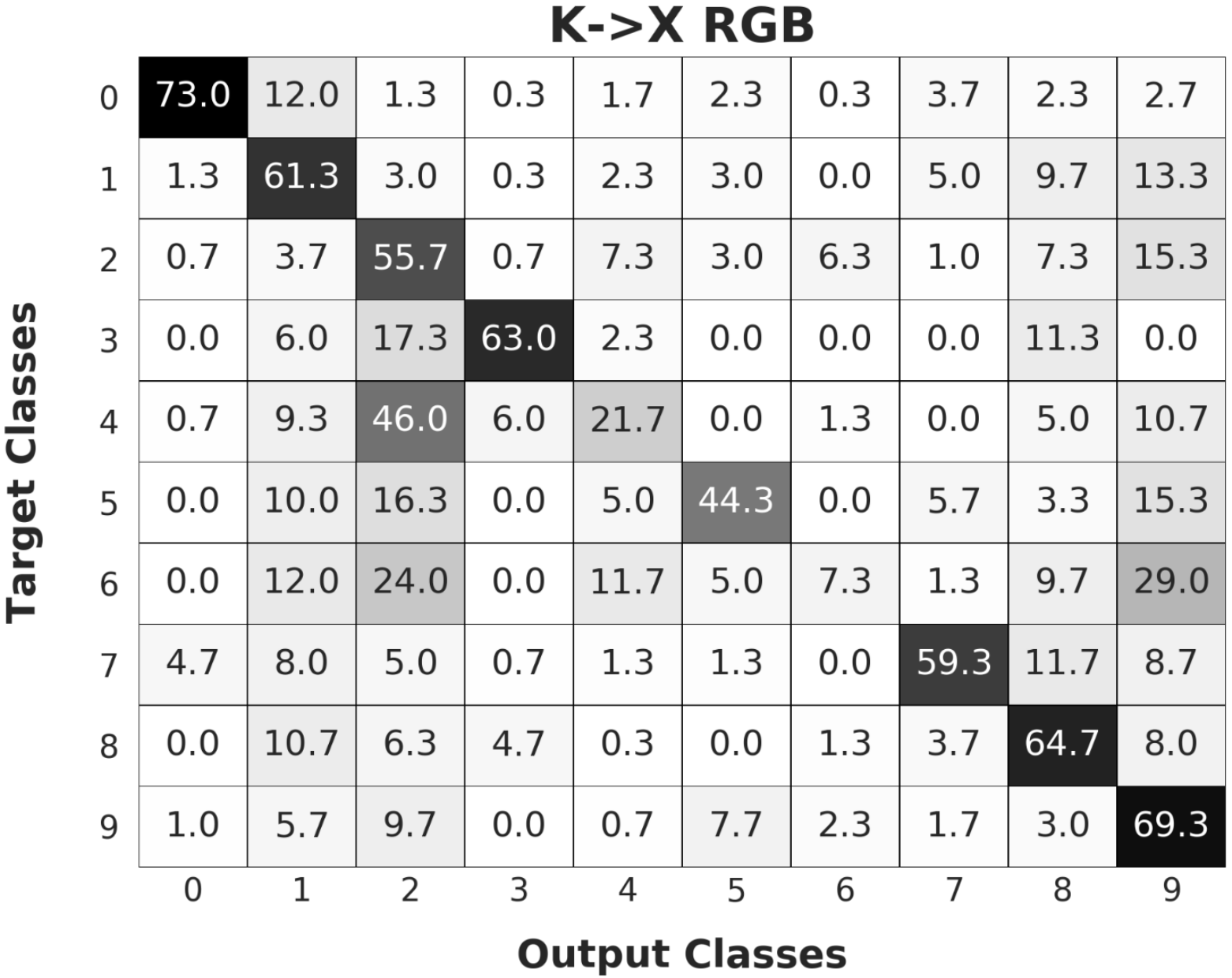} &\includegraphics[width=0.236\textwidth]{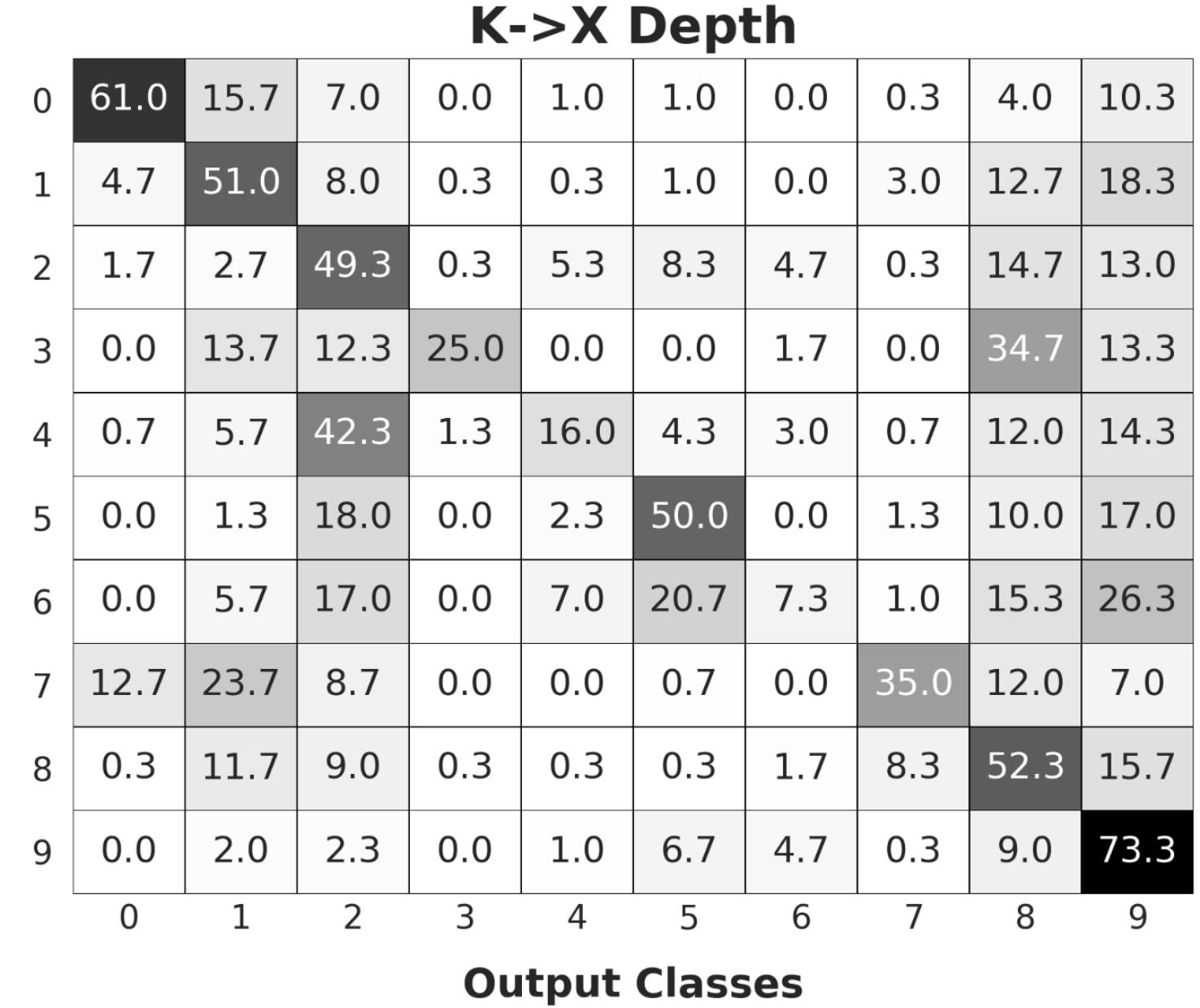} \\
\end{tabular}
\vspace{-2mm}
\caption{Accuracy (\%) across domains for single modality. The performance drop shows the effect of the domain shift.
The confusion matrices for the case K$\rightarrow$X also indicate that the behavior across domains is different for the two modalities: classes 2 (classroom) and 7 (kitchen) are the ones mainly affected by the domain shift, respectively for the RGB and depth modalities.}
\label{domain_shift}
\end{center}
\vspace{-5mm}
\end{table}

The qualitative tsne \cite{tsne} data analysis in Figure \ref{fig:tsne} shows that the samples from each camera belong to different distributions and tend to occupy a different region of the space. This is more evident for the depth modality where the samples from Realsense are well separated from the other domains. We also verified quantitatively that the observed appearance variation among the images of the different cameras causes a domain shift problem. We defined a simple experiment focusing on the K and X cameras and organizing their images into three 70\%/30\% train/test splits. We trained a simple ResNet-18 classification model and we evaluated it both within each camera and across cameras: the average results over the splits are respectively reported in the first and second row of Table \ref{domain_shift} for each of the two modalities. The drop in performance (summarized in the last row) clearly demonstrates the existence of a significant domain shift. Moreover, the confusion matrices of the K$\rightarrow$X case show how the domain shift affects the per-class recognition accuracy.

\section{Method}
\label{sec:method}
\noindent\textbf{Intuition} Several recent works have discussed how self-supervised learning supports visual domain generalization \cite{PAMIselfsup,xu2019self}. When dealing with multi-modal source and target domains, one basic self-supervised task is that of transforming one modality into the other and vice-versa. In our setting, this means predicting the depth information from an RGB instance and generating the RGB information from a depth image. This second direction is of course more difficult than the first, however by optimizing for both these objectives, we train a model that captures the core relation between the two modalities. When this is done at the same time over source and target, the model focuses on what makes the relation between RGB and depth domain invariant. Thus, we expect the obtained multi-modal representation to help in cross-domain scene classification. 

\begin{figure*}[!t]
    \centering
\includegraphics[width=1\textwidth]{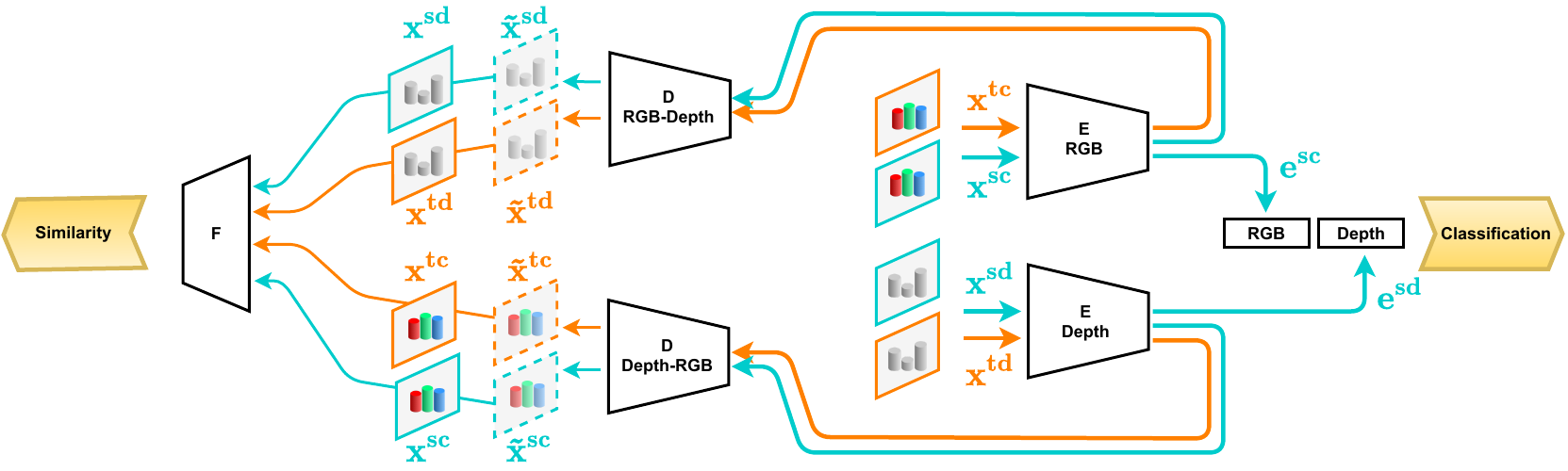}
    \caption{Overview of our Translate-to-Adapt method for RGB-D scene recognition across domains. The main components are the encoders (E), the inter-modality decoders (D), the semantic feature extractor (F), and finally the Classification and Similarity evaluation heads. The two encoders are identical and each deals with one of the image modality: RGB or depth. The obtained features are concatenated and enter into the classifier. The two decoders have the same structure but each focuses on one modality translation direction: from RGB to depth or vice-versa. Every image generated by the decoders is paired with its corresponding original version: the features are extracted via F and compared by the similarity head. Note that only the supervised source data enter the classification task, while both source and target data  go through the inter-modality generation self-supervised task. We use the notation presented in Section \ref{sec:method}.}
    \label{fig:architecture}\vspace{-3mm}
\end{figure*}

\noindent\textbf{In more Technical Terms} Starting from the source labeled and the target unlabeled multi-modal images, our goal is to predict the scene class of the target data. 
In the following we will indicate with
$S=\{(\bx_i^{sc},\bx_i^{sd}),\by_i^{s}\}_{i=1}^{N^s}$ the source samples. The superscripts $c,d$ refer respectively to the color (RGB) and depth modality, while $\by_i^{s} \in \mathbb{R}^{|\mathcal{Y}|}$ denotes the one-hot encoded scene class label and $|\mathcal{Y}|$ indicates the number of classes. The target samples $T=\{(\bx_i^{tc},\bx_i^{td})\}_{i=1}^{N^t}$ are unlabeled and are drawn from a different distribution with respect to the source, but shares with it the same class set.
The relation between the two data modalities may contain helpful cues for scene recognition. One way to extract and exploit those cues is to add to the main classification task the auxiliary objective of inter-modal translation: both $\bx^{*c}\rightarrow\bx^{*d}$ and $\bx^{*d}\rightarrow\bx^{*c}$.
We used the star $*$ to indicate a generic domain: since this mapping is self-supervised it can be applied both on source and target.
Thus, it bridges the two domains adapting the learned representation.

\noindent\textbf{Network architecture and Optimization} The architecture of our Translate-to-Adapt method is presented in Figure \ref{fig:architecture}. It consists of six main components: two modality-specific encoders (E), two decoders (D), one for each modality translation direction, a feature extractor (F) and the final classifier. Both source and target data enter the two encoders that map the original images into a feature embedding of equal dimensionality for the two modalities $\be_i^{*c}=E_{rgb}(\bx_i^{*c})$, $\be_i^{*d}=E_{depth}(\bx_i^{*d})$. 
The main classification task runs on the concatenated  features of the source data $\{\be_i^{sc},\be_i^{sd}\}_{i=1}^{N^s}$. 
The obtained representations for both source and target are fed as input to the corresponding decoders that translate them in the twin modality: $\tilde{\bx}_i^{*d}=D_{rgb-depth}(E_{rgb}(x_i^{*c}))$ and  $\tilde{\bx}_i^{*c}=D_{depth-rgb}(E_{depth}(x_i^{*d}))$. The generated images are paired with their original version and the difference among the features extracted by F is minimized for each case: $\{\tilde{\bx}_i^{sc}, \bx_i^{sc}\}_{i=1}^{N^s}$, $\{\tilde{\bx}_i^{sd}, \bx_i^{sd}\}_{i=1}^{N^s}$, $\{\tilde{\bx}_i^{tc}, \bx_i^{tc}\}_{i=1}^{N^t}$, $\{\tilde{\bx}_i^{td}, \bx_i^{td}\}_{i=1}^{N^t}$.
Overall, the two objectives of classification and instance similarity are jointly optimized respectively via a cross-entropy loss function $\mathcal{L}_{cls}$ and the content similarity loss $\mathcal{L}_{sim}$ among the generated-original sample pairs. The latter is an L1 loss
\begin{equation}
    \sum_{l=1}^L||F^l(\tilde{\bx}_i^{*c}) - F^l({\bx}_i^{*c})||_1 + ||F^l(\tilde{\bx}_i^{*d}) - F^l({\bx}_i^{*d})||_1\
\end{equation}
measured over multiple internal layers of the F module (l= layer1-layer4 in ResNet-18). 
Finally, the total loss is 
\begin{equation}
    \mathcal{L}_{cls} + \alpha^s\mathcal{L}_{sim}^s +  \alpha^t\mathcal{L}_{sim}^t~.
\end{equation}

\noindent\textbf{Implementation Details}
The defined optimization problem guides the training of encoders and decoders, while for F we used a frozen model. All the components have a ResNet-18 structure pre-trained on Imagenet. 
The loss hyperparameteres $\alpha^s$ and $\alpha^t$ are set respectively to 10 and 3 (see the ablation analysis in Section \ref{sec:results}). 

We designed the network modules by following \cite{du2019translate}, but the learning procedure differs. Besides including the target data, in our Translate-to-Adapt the multi-modal fusion strategy for classification is learned end-to-end with all the other network components, rather than with a two-step process.
We trained the model with ADAM stochastic optimization, 
setting the batch size to 40 and a total of 70 epochs. The initial learning rate is $2\times10^{-4}$ and decreases linearly for the last 50 epochs.
Depth images are encoded offline to HHA \cite{gupta2014hha} and together with the RGB images are resized and randomly cropped.  At test time, we used the central crops.

\section{Experiments}
\label{sec:exper}
\noindent\textbf{Reference Methods}
To understand the challenges of learning a multi-modal cross-domain scene recognition model, we perform a benchmark analysis with existing approaches originally developed either for multi-modal scene recognition or single-modal cross-domain object classification. 

From the first family we consider the approach named {Translate-to-Recognize} (Tran-Rec) \cite{du2019translate}, and the recent {Centroid Based Concept Learning} (CBCL) \cite{Ayub_2020_BMVC} which outputs class assignments on the basis of the linear combination of multi-modal sample distances. Both those methods were developed to work on training and test data drawn from a single domain. We also consider as baseline the basic {ResNet-18}. In general, those methods are \emph{Source Only}, meaning that during training the target test data is not available.
We use \emph{Fusion} to indicate a simple multi-modal strategy where a separate network is trained for each modality until convergence. The feature extractors are then frozen, while the produced representations are concatenated and fed as input to a fully connected layer that is trained on them for scene classification.  We indicate instead with \emph{Fusion++} a network that deals at once with the two modalities, by training  end-to-end both the feature representation and the multi-modal classifier. 

For the second family of methods, the unlabeled target data are provided together with the labeled training.  {GRL} \cite{Ganin:DANN:JMLR16} relies on a domain classifier exploited in an adversarial fashion to reduce the feature distribution difference between source and target.
{AFN} \cite{AFN} starts from the observation that target samples are often characterized by feature norm values much lower than those of the source data and proposes to progressively increase them.
{CycleGAN} \cite{CycleGAN2017} is an unsupervised generative approach that can be used to change the style of the source data and make them resemble the target. We use it to produce target-like RGB and depth images from the annotated source samples. Finally, the models trained on them are combined with the Fusion strategy.

Up to our knowledge, there is only one previous work that focused on multi-modal cross-domain object classification. We indicate the proposed method as {Relative Rotation} (Rel. Rot) \cite{LoghmaniRPPCV20}: it exploits the homonym auxiliary self-supervised task to infer the correlation between RGB and depth in order to produce robust domain-invariant features for the main recognition task. 

All those reference methods are compared against our Translate-to-Adapt (Tran-Adapt) which is designed as a Fusion++ approach with the encoders, decoders, and classification model trained at once. 

\noindent\textbf{Results and Ablation}
\label{sec:results} Table \ref{sota_comparisons} shows the classification accuracy values obtained by the considered reference approaches and by our Tran-Adapt method. Specifically, the top part contains the Source Only baselines whose results indicate that combining the two modalities of the source data improves the recognition performance across domains. For completeness, we also developed the Fusion++ version of the Tran-Rec method, although the end-to-end training procedure was not included in the original paper \cite{du2019translate}. 
The CBCL Fusion approach outperforms the others.

\begin{table}[t]
\begin{center}
\resizebox{\columnwidth}{!}{
\begin{tabular}{c@{~~}l@{~~}c@{~~}c|c@{~~}c@{~~}c|c}
\hline
Method & &  K $\rightarrow$ X & X $\rightarrow$ K & K $\rightarrow$ R & X $\rightarrow$ R & KX  $\rightarrow$ R & AVG\\
\hline
\multirow{4}{*}{ResNet-18} & RGB & 47.56 &57.55&	38.34&	44.88&	41.82&	46.03 \\
& Depth & 38.76&	54.42&	26.56&	26.87&	30.98&	35.52\\
& Fusion & 50.66&	62.91&	44.54&	46.54&	42.56&	49.44\\
& Fusion++ & 47.54&	60.27&	39.56&	36.32&	43.71&	45.48\\
\hline
\multirow{4}{*}{Tran-Rec \cite{du2019translate}}& RGB-D & 52.54&	61.68&	38.63&	46.24&	44.59&	48.74\\
& D-RGB & 37.13&	53.49&	29.77&	29.06&	32.25&	36.34 \\
& Fusion & 53.92&	63.40&	39.35&	43.40&	48.29&	49.67 \\
& Fusion++ & 51.17&	62.62&	39.53&	41.38&	50.87&	49.11 \\
\hline
CBCL \cite{Ayub_2020_BMVC} & {Fusion} & {55.35} &	{60.57} &	{50.51} & {42.45} & {49.94} & {51.76}  \\
\hline\hline
\multirow{4}{*}{GRL \cite{Ganin:DANN:JMLR16}}&	RGB&	50.11&	59.88&	53.30&	51.18&	46.82&	52.26\\
&Depth&	45.25&	54.29&	37.30&	32.41&	37.80&	41.41\\
&Fusion&	48.28&	64.73&	\textbf{53.53}&	51.91&	47.51&	53.19\\
&Fusion++&	50.94&	61.91&	53.45&	48.90&	48.85&	52.81\\
\hline
\multirow{4}{*}{AFN \cite{AFN}}&	RGB&	51.59&	56.73&	52.11&	47.63&	46.86&	50.98\\
&Depth&	40.22&	51.88&	34.20&	32.33&	35.20&	38.77\\
&Fusion&	51.29&	61.88&	47.84&	50.25&	50.07&	52.27\\
&Fusion++&	56.74&	57.89&	52.13&	49.05&	45.66&	52.30\\
\hline
CycleGAN \cite{CycleGAN2017} & Fusion & 54.25 & 63.19 & 53.02 & 48.02 & 54.65 & 54.63 \\
\hline
Rel. Rot. \cite{LoghmaniRPPCV20}& Fusion++ & 50.98&	\textbf{65.99}&	48.33&	52.24&	53.53&	54.21\\
\hline\hline
\multirow{4}{*}{Tran-Adapt} & RGB-D & 52.11 & 61.91 & 46.93 & 51.27 & 54.88 & 53.42 \\
& D-RGB & 48.09 & 55.69 & 38.95 & 38.78 & 40.79 & 44.46 \\
& Fusion & 55.61 & 65.23 & 41.90 & 43.59 & 48.03 & 50.87 \\
& Fusion++ & \textbf{56.79}&	64.41&	48.13&	51.02&	55.31&	55.13 \\ 
\hline
 Tran-Adapt Aug & Fusion++ &  55.65 & 65.92	& 53.01	& \textbf{52.56}	& \textbf{55.59}	&	\textbf{56.55} \\ 
\hline\vspace{1mm}
\end{tabular}
}
\begin{tabular}{c@{~~}c}
\hspace{-5mm}\includegraphics[height=0.2\textwidth]{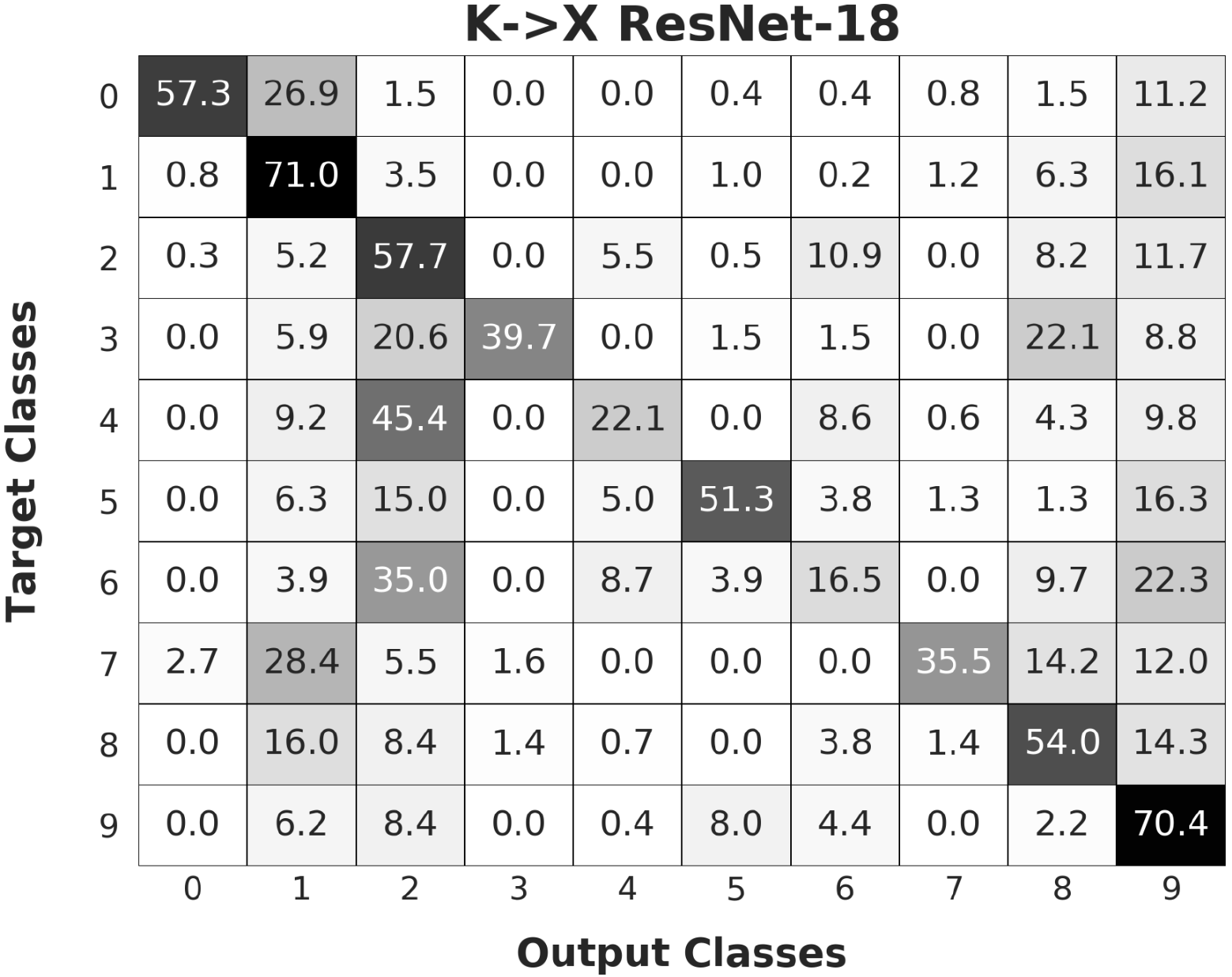} &\includegraphics[height=0.2\textwidth]{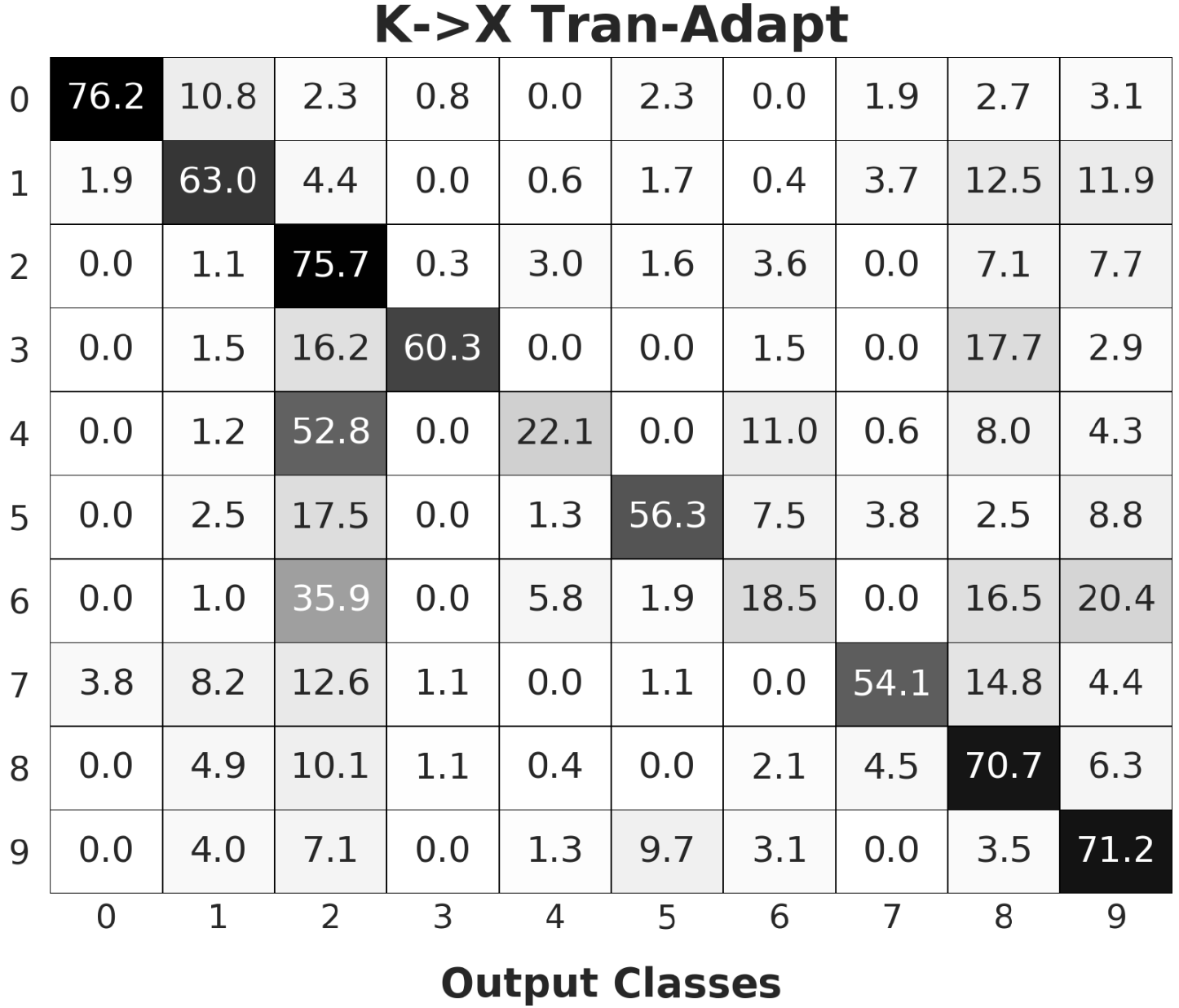} \\
\end{tabular}\vspace{-2mm}
\caption{Accuracy (\%) of several methods for RGB-D domain adaptation. Top results in bold. The confusion matrices show the K$\rightarrow$X per-class results for the ResNet-18 baseline and Tran-Adapt (Fusion++).} \vspace{-6mm}
\label{sota_comparisons}
\end{center}
\end{table}

\begin{figure*}[!t]
\centering
\begin{tabular}{c|c|c}
\hline
{Original} & {Rel. Rot.} & {Tran-Adapt}\\
\hline
\includegraphics[height=0.14\textwidth]{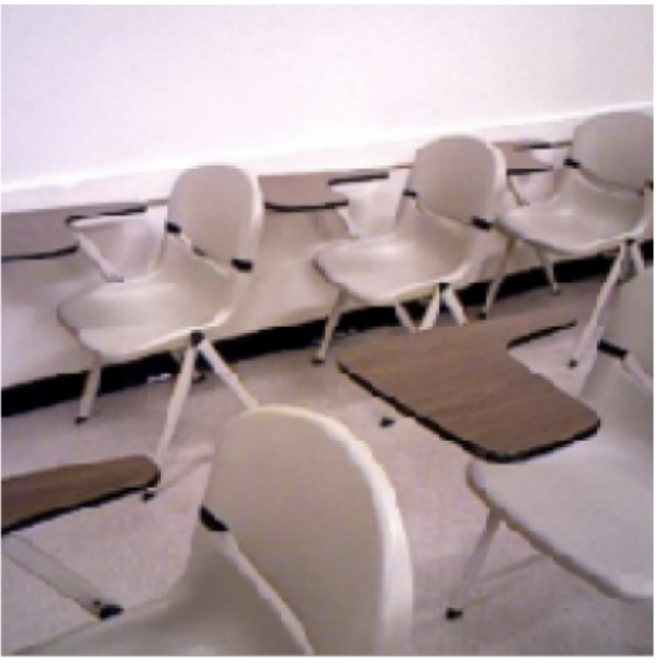}& \includegraphics[height=0.14\textwidth]{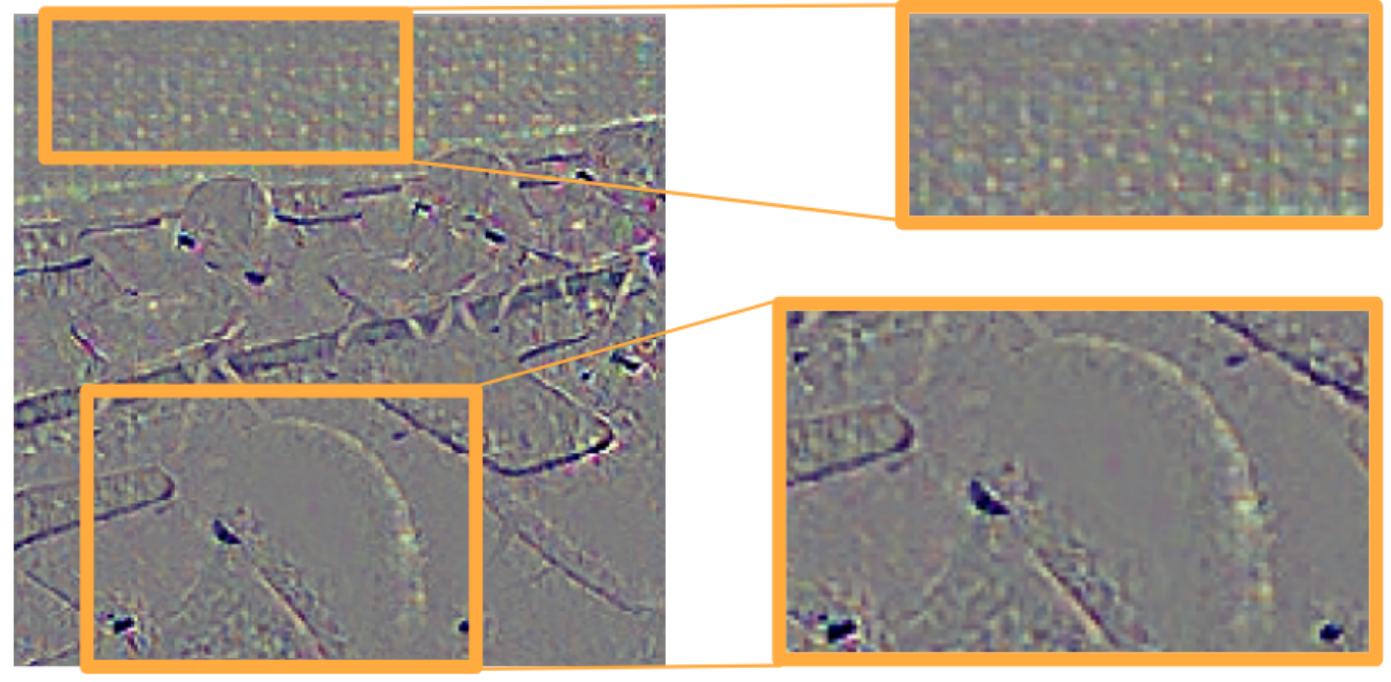} & 
\includegraphics[height=0.14\textwidth]{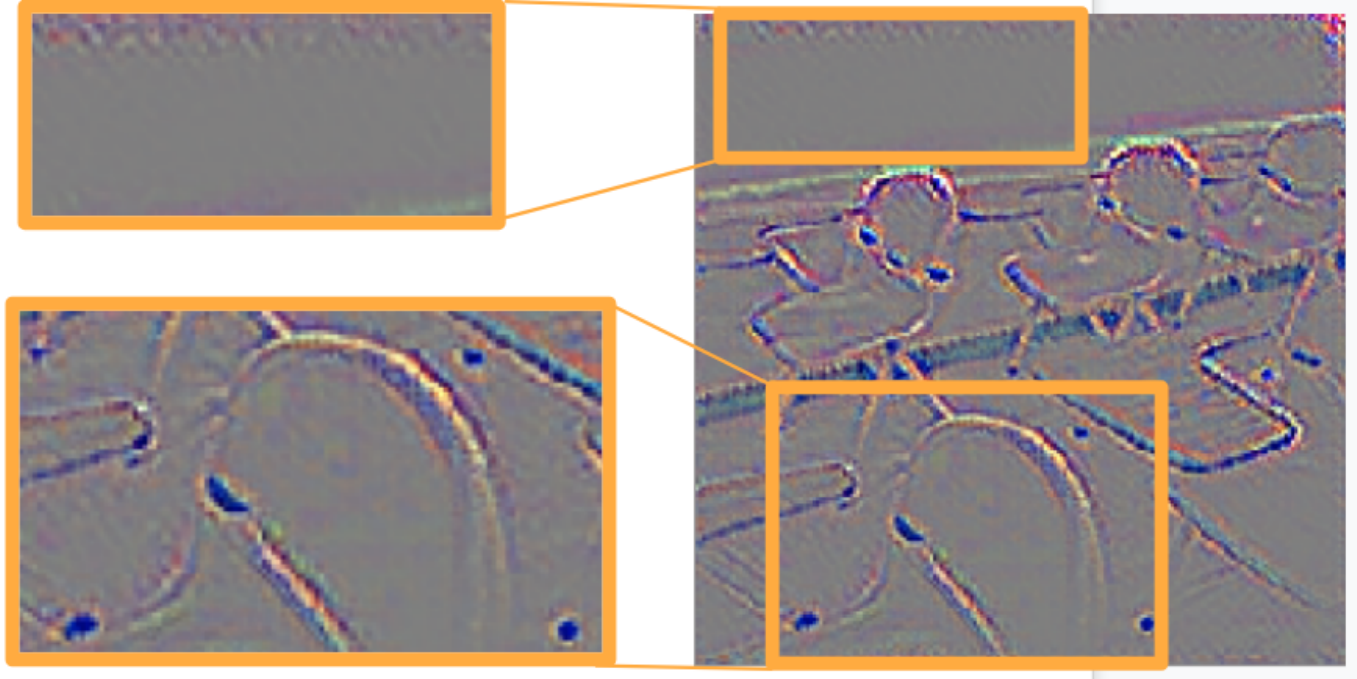} \\
\includegraphics[height=0.14\textwidth]{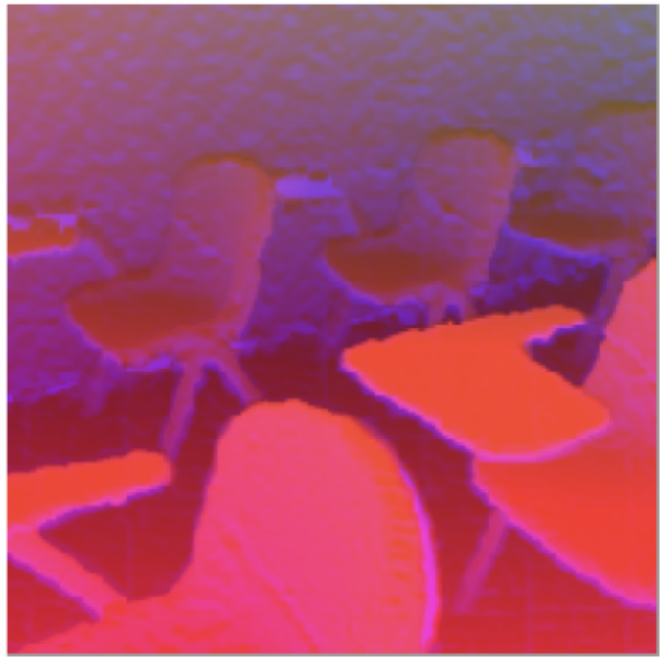}& \includegraphics[height=0.14\textwidth]{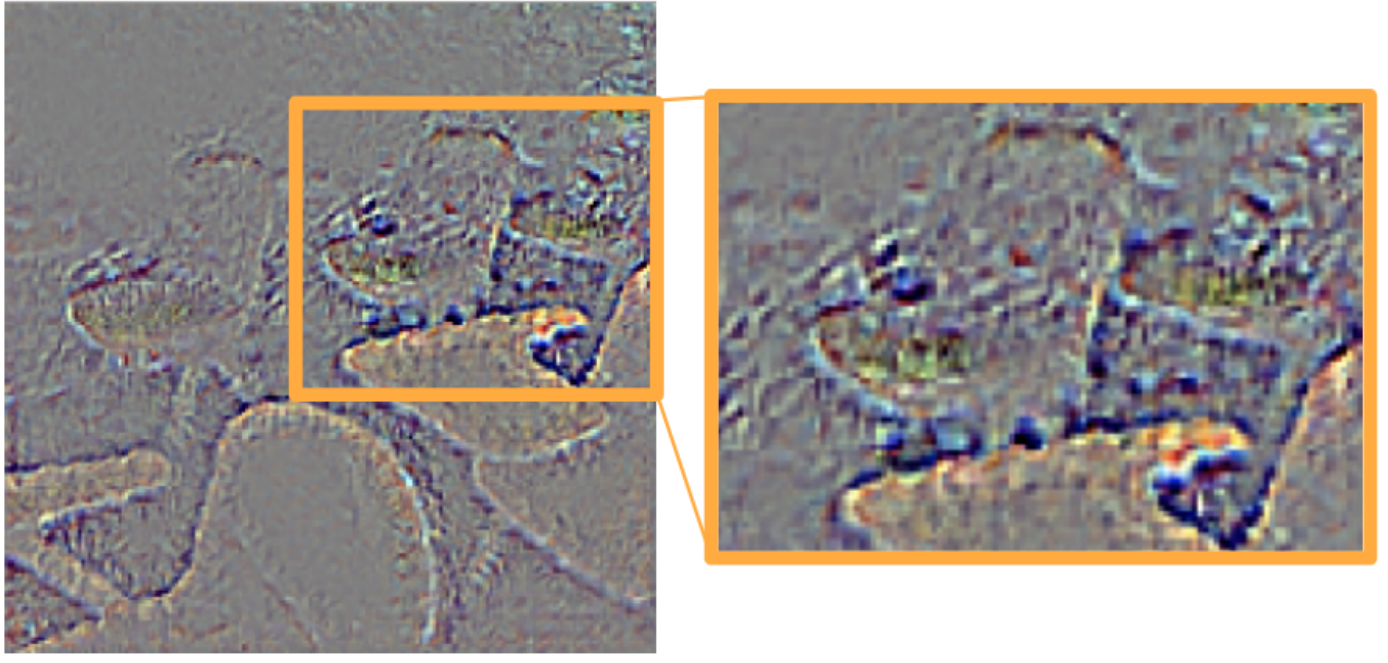} & 
\includegraphics[height=0.14\textwidth]{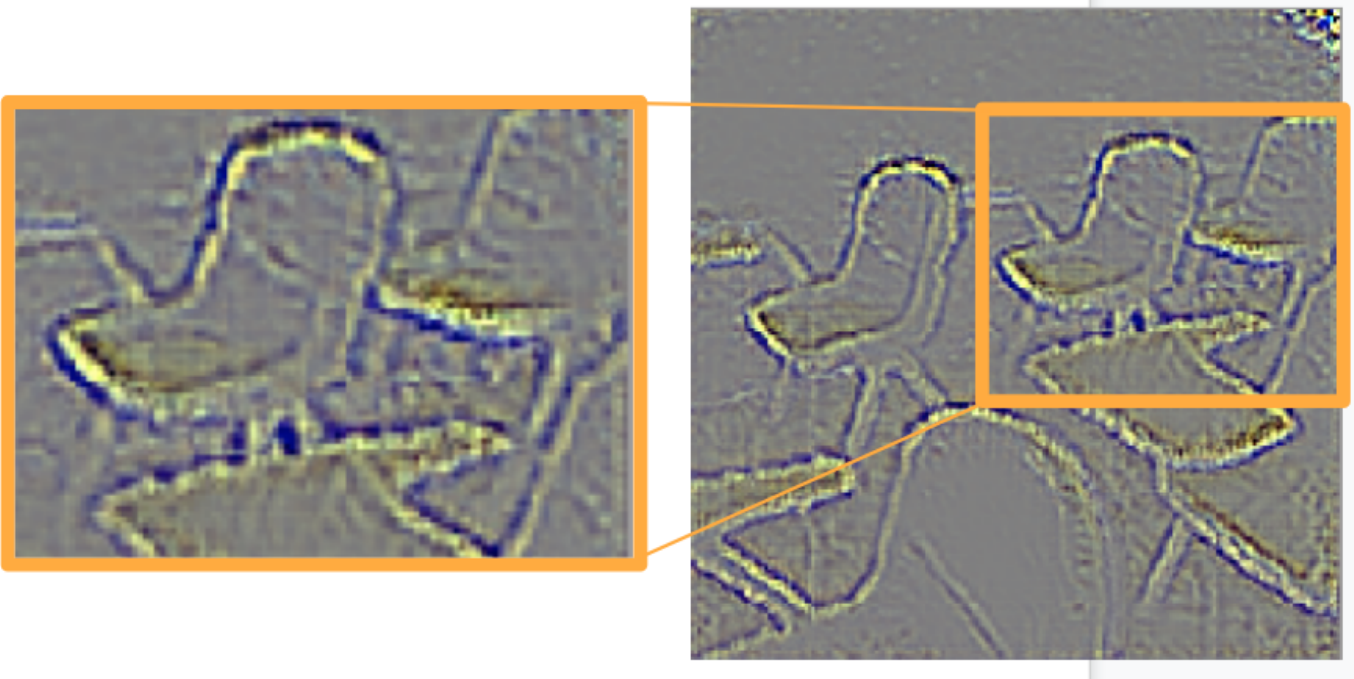} \\
\hline
\includegraphics[height=0.14\textwidth]{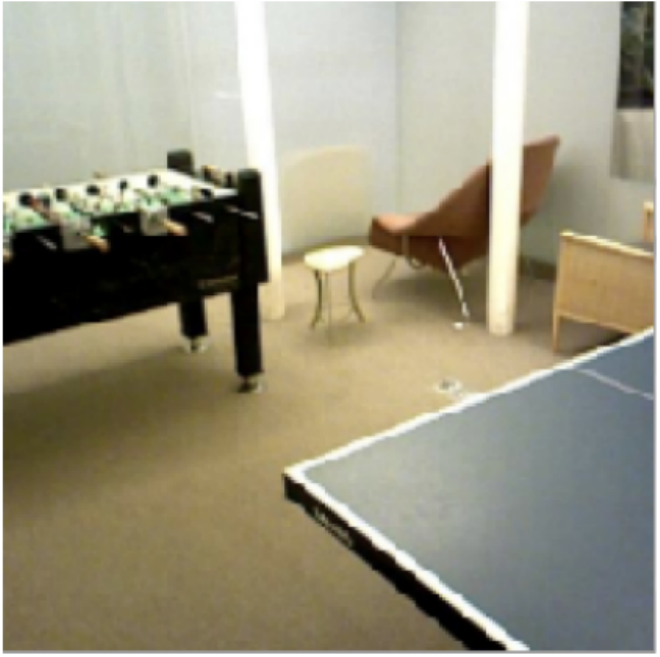}& \includegraphics[height=0.14\textwidth]{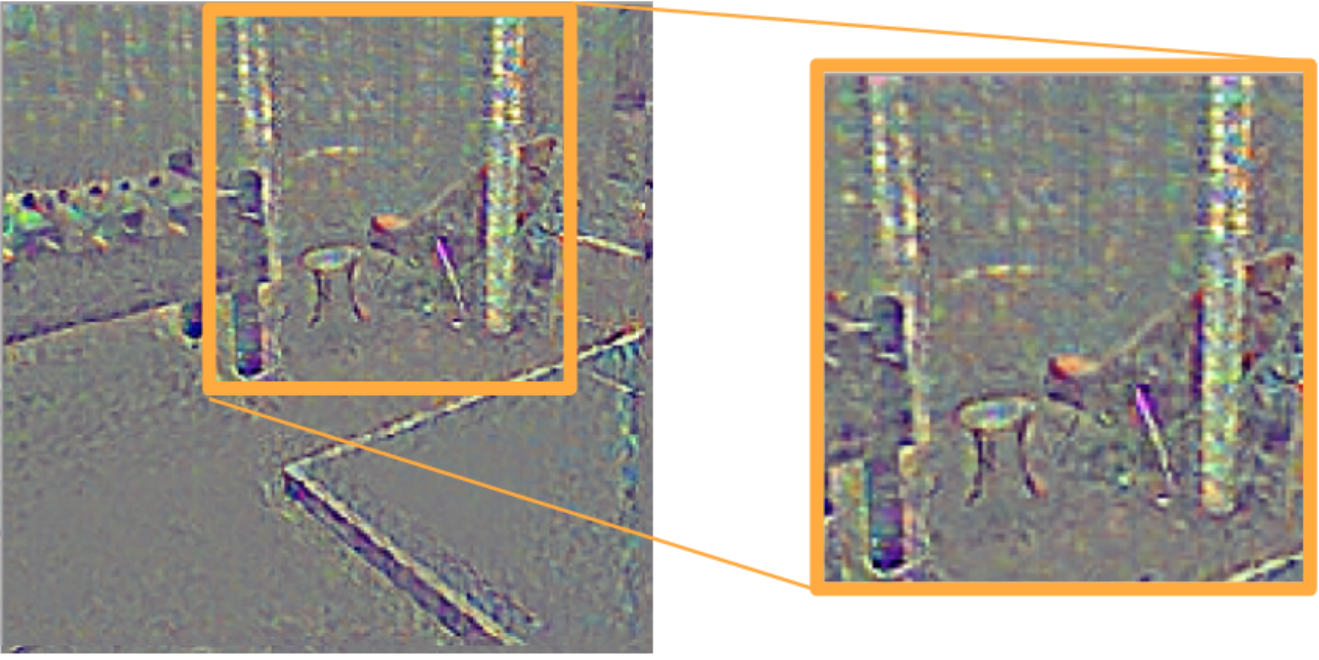} & 
\includegraphics[height=0.14\textwidth]{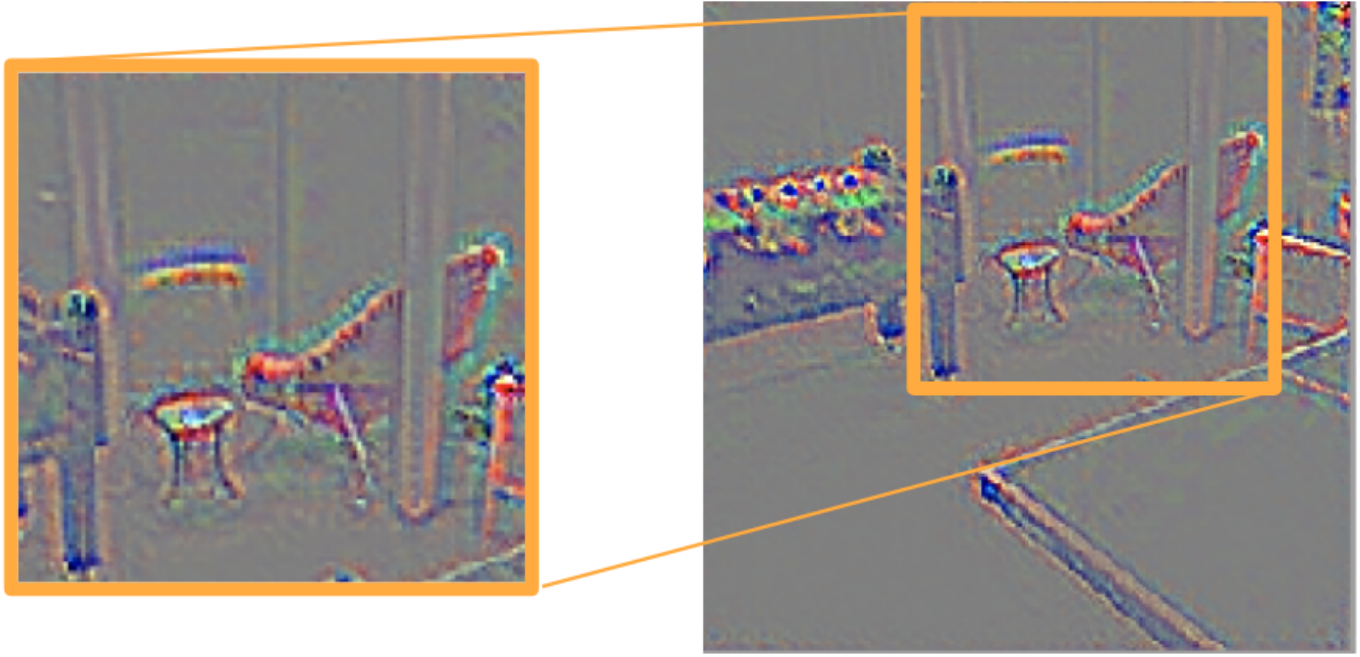} \\
\includegraphics[height=0.14\textwidth]{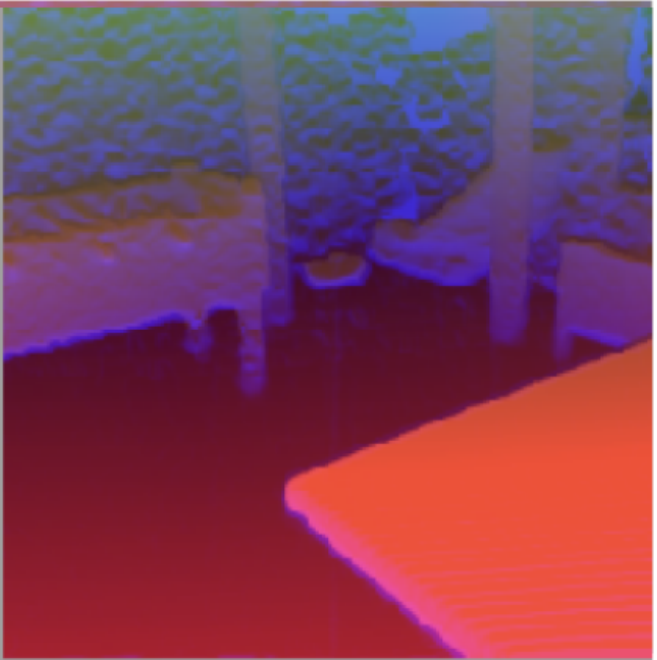}& \includegraphics[height=0.14\textwidth]{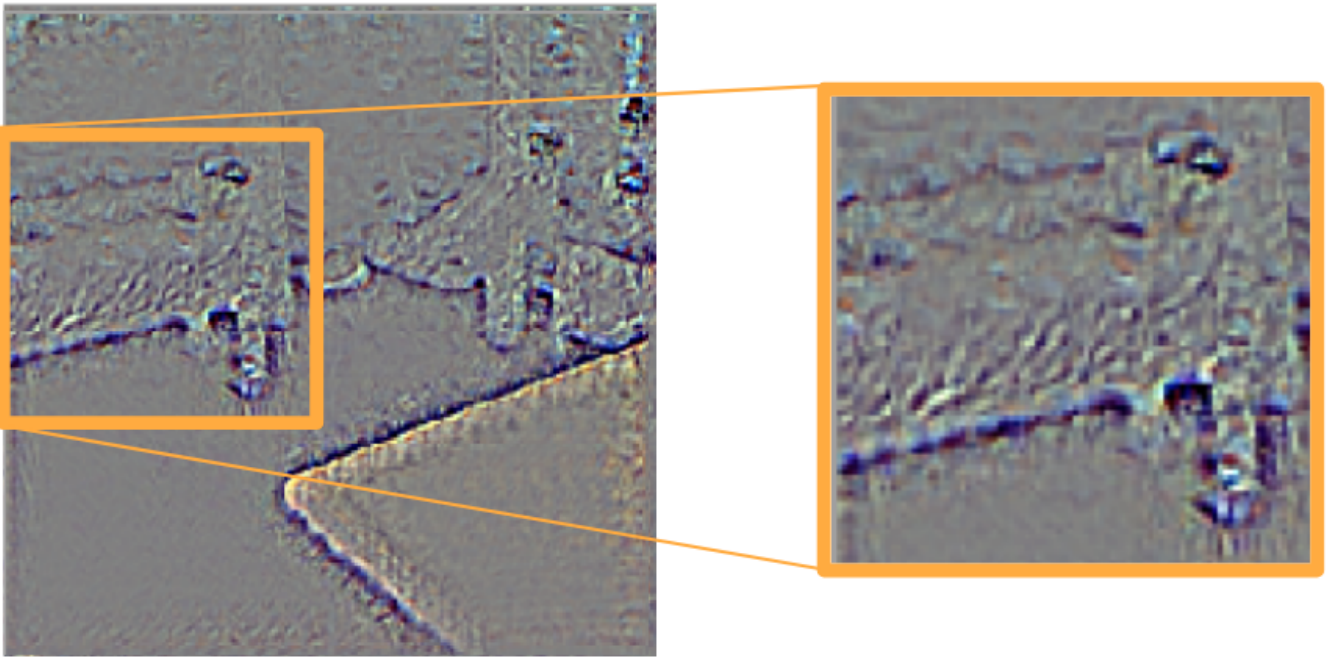} & 
\includegraphics[height=0.14\textwidth]{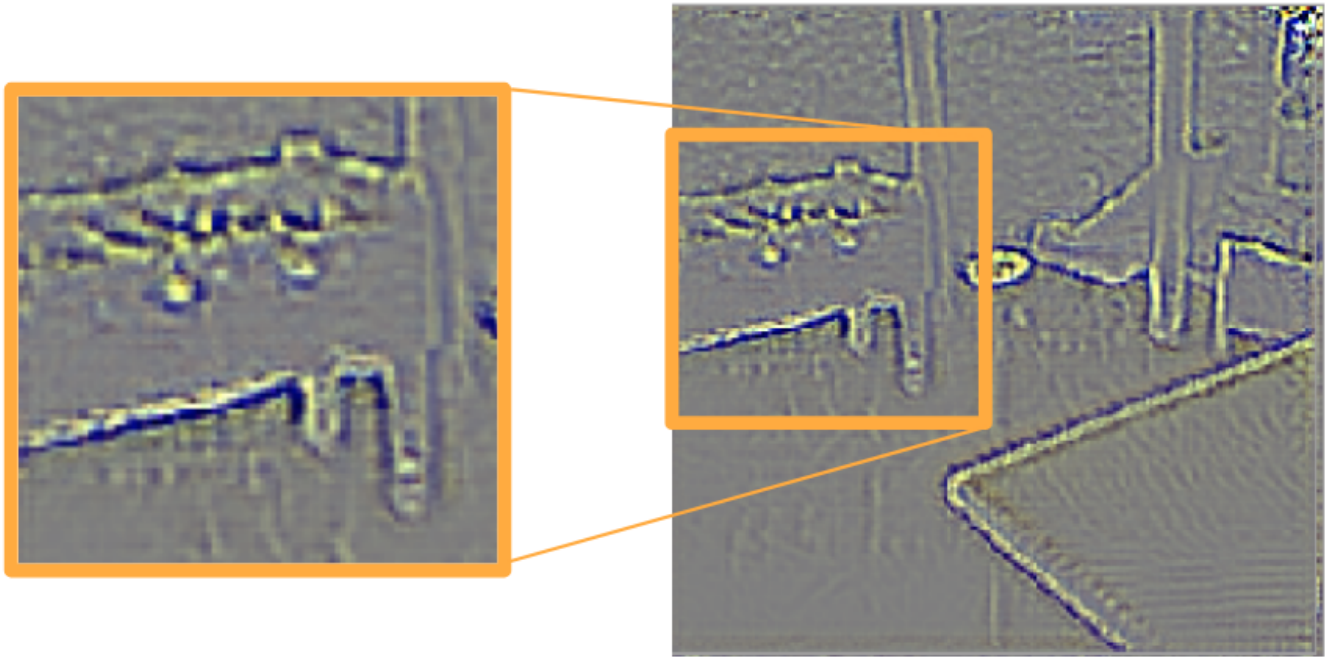} \\
\hline
\end{tabular}
\caption{ Visualizations obtained by guided backpropagation \cite{springenberg2014striving} that show the most important pixels used by Rel. Rot. \cite{LoghmaniRPPCV20} and our Tran-Adapt.}\vspace{-2mm}
        \label{fig:visualizations}
\end{figure*}

The central part of the table presents the results of the domain adaptive methods. Even in this case, the multi-modal versions improve over the corresponding single-modal ones. The advantage is more evident for the style-transfer-based CycleGAN method than for the feature alignment approaches GRL and AFN. 
Finally, the performance of Rel. Rot., the only existing method that exploits the inter-modality relation in both the domains, is slightly lower than that of CycleGAN.

The bottom part of the table shows the results of our Tran-Adapt. Specifically, the Fusion++ version outperforms all the considered competitors. By looking at the confusion matrices of the K$\rightarrow$X experiment we observe that for most of the classes there is a clear performance gain when using Tran-Adapt. A reduction in the wrong assignments is evident between class 7. kitchen and 1. bedroom, as well as between 2. classroom and 3. computer\_room.

We can also take advantage of the generative nature of Tran-Adapt by exploiting the produced images as data augmentation. More precisely, the depth images generated by the RGB-D model and the RGB images produced by the D-RGB model can enter the Fusion++ network as both source and target input data. By following \cite{du2019translate} we used a random subset of the generated data with the number controlled as 30\% of the batch size. The obtained Tran-Adapt Aug version shows a further gain in performance, producing the top average accuracy 56.55\%. 

As specified in the previous section, for Tran-Adapt we set  $\alpha^s=10$ and $\alpha^t=3$.  
The first value is the same used by Tran-Rec in \cite{du2019translate} and we kept it fixed. The second tunes the importance of the self-supervised task running on the target data: to get discriminative features the main focus should remain on the annotated source, with  $\alpha^t<\alpha^s$. We chose $\alpha^t=3$  from a preliminary validation analysis on the separate RGB-D and D-RGB directions and we maintain that value also for the Fusion and Fusion++ versions of our approach. An ablation analysis on the role of the source/target self-supervised translation task can be done considering that we get back to 
Tran-Rec ($\alpha^s=10$, $\alpha^t=0$, Fusion++  49.11\%) when turning off the target contribution.
 Instead, by turning off the source contribution while maintaining the target one ($\alpha^s=0$, $\alpha^t=3$, Fusion++  54.22\%), we observe an adaptation effect, which improves when leveraging on both the source and target components ($\alpha^s=10$ and $\alpha^t=3$, Tran-Adapt, Fusion++  55.13\%). In particular keeping $\alpha^s=10$, but changing $\alpha^t=\{1,2,3,4\}$ causes a minimal average result variation for Fusion++ \{54.71, 54.44, 55.13, 54.81\} (\%). 

\noindent\textbf{Self-supervision for Cross-Domain Scene Recognition} Both Rel. Rot. and Tran-Adapt exploit self-supervised tasks (rotation recognition and RGB-depth image mapping) to learn inter-modality cues that support cross-domain adaptation. Still, considering the observed performance difference, we decided to investigate more in depth their behavior. Specifically, we searched for possible shortcuts followed by the rotation auxiliary task that might have misled the scene recognition process. Indeed, Rel. Rot. was originally designed for object recognition on datasets where the objects are typically well centered in the images and the background information are marginal. When dealing with scenes, the risk of focusing on low semantically meaningful cues to predict the image orientation increases, affecting also the final scene class assignment.  
In Figure \ref{fig:visualizations} we show the results of the \emph{guided backpropagation} \cite{springenberg2014striving} approach. By visualizing the most relevant pixels used by Rel. Rot. and Tran-Adapt we can claim that both the methods focus on object boundaries, but Rel. Rot. includes spurious information on uniform regions, and relies on neat lines (see the third image row and the columns in the image) in the background.

\begin{figure}[t]
\hspace{-2mm}
    \begin{center}
    \begin{tabular}{@{~}c|@{~}c@{~}|@{~}c@{~}|c@{~}}

    \hline
    \multirow{2}{*}{\small{Original}} & \multicolumn{3}{c}{\small{Generated}}\\
    \cline{2-4}
    & \small{Tran-Rec \cite{du2019translate}} & \small{Tran-Adapt} & \hspace{-2mm}   \small{Tran-Adapt Aug} \\

    \hline
    \includegraphics[width=0.1\textwidth]{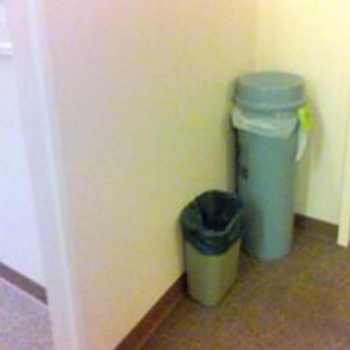} &     \includegraphics[width=0.1\textwidth]{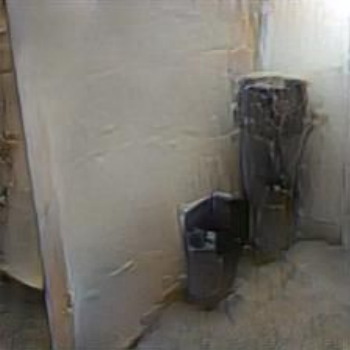} &     \includegraphics[width=0.1\textwidth]{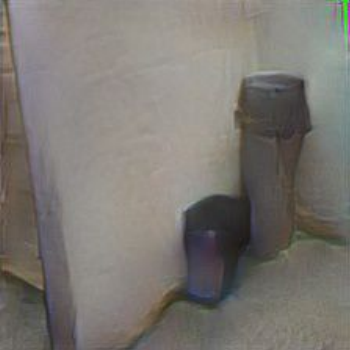} & \hspace{-2mm}     \includegraphics[width=0.1\textwidth]{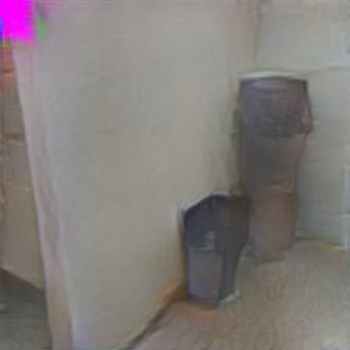}\\
     \includegraphics[width=0.1\textwidth]{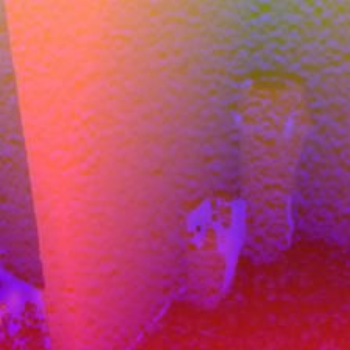} &     \includegraphics[width=0.1\textwidth]{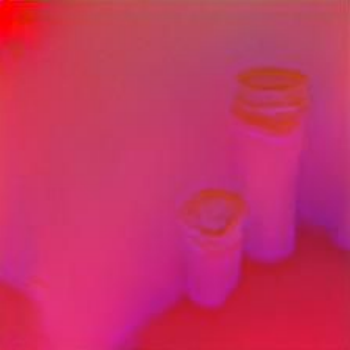} &   \includegraphics[width=0.1\textwidth]{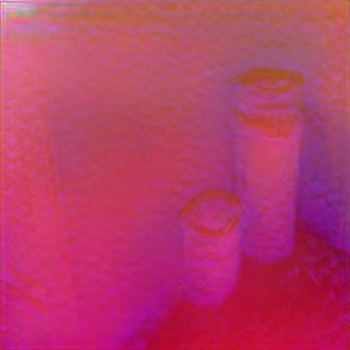} & \hspace{-2mm}     \includegraphics[width=0.1\textwidth]{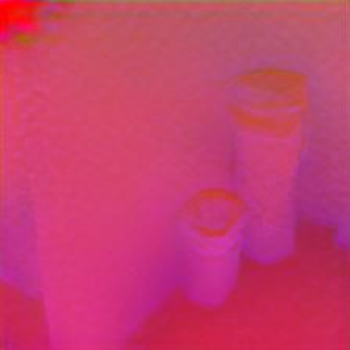}\\   
    \hline
    \includegraphics[width=0.1\textwidth]{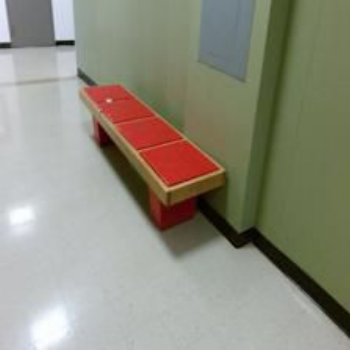} & \includegraphics[width=0.1\textwidth]{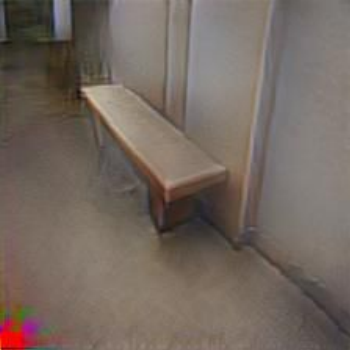} &     \includegraphics[width=0.1\textwidth]{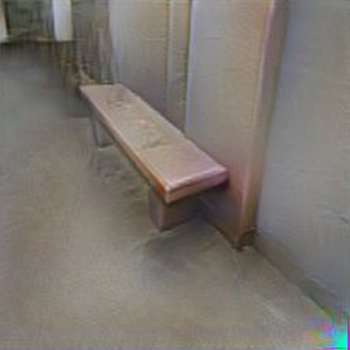} &  \hspace{-2mm}     \includegraphics[width=0.1\textwidth]{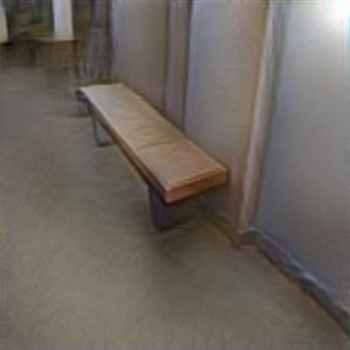}\\
    \includegraphics[width=0.1\textwidth]{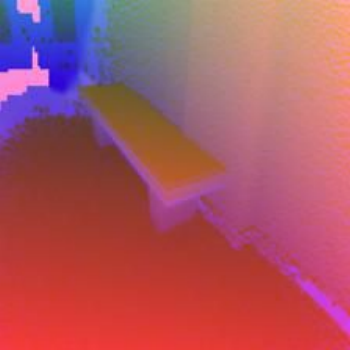} &     \includegraphics[width=0.1\textwidth]{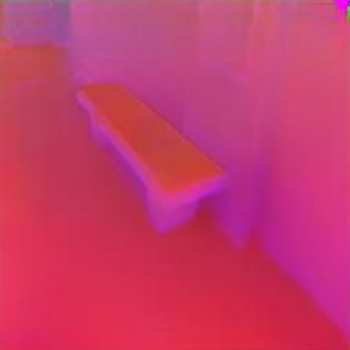} &     \includegraphics[width=0.1\textwidth]{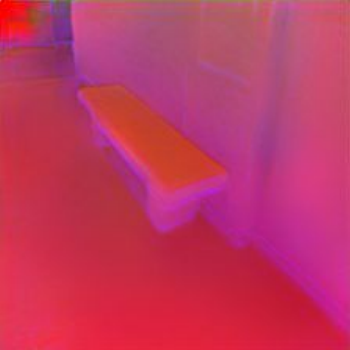} &  \hspace{-2mm}     \includegraphics[width=0.1\textwidth]{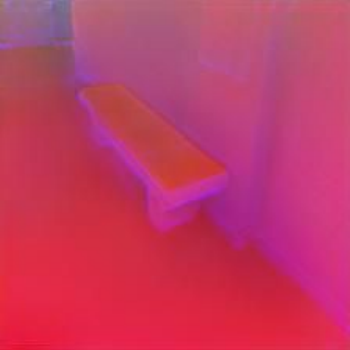}\\
    \hline
    
 \end{tabular}

    \end{center} 
    \vspace{-5mm}
    \caption{Qualitative comparison of real and generated images on the unseen class \emph{corridor}. In these examples, the improvement of Tran-Adapt and its Aug version is particularly evident on the RGB images where the uniform regions (walls and floor) appear smoother than in Tran-Rec. } \vspace{-3mm}
    \label{fig:unseen_classes}
\end{figure}

\noindent\textbf{Missing Modality prediction on Novel Target Scenes} Since the final purpose of the proposed model is scene recognition we mainly focused on the classification performance output. Still, the similarity objective and the decoders included in Tran-Adapt provide a generative tool that can be exploited for side tasks. One possibility is that of producing the RGB or depth modality for single-modal input images. Indeed in case of problems with the sensing devices, it might happen that one of the modalities is missing and needs to be hallucinated. When this lost modality issue affects images belonging to scene categories never seen during training the task becomes particularly challenging. To evaluate Tran-Adapt in this setting, we selected from SUN RGB-D three classes not originally included in our collection and we created a new small dataset over all the four available cameras (see Table \ref{dataset2}). We tested the generation performance on both the image modalities of the pre-trained Tran-Rec and Tran-Adapt models. Specifically, we measured the pixel-to-pixel L2 difference between the generated and original image: the results in Table \ref{unseen_classes} show that Tran-Adapt is better able to approximate the ground truth image than Tran-Rec, further demonstrating its generalization abilities. Some examples of the generated images are shown in Figure \ref{fig:unseen_classes}.

\begin{table}[t]
\resizebox{\columnwidth}{!}{
\begin{tabular}{lcccc}
\hline
 Class name & Kinect v1 & Kinect v2 & Realsense & Xtion\\
 \hline
 corridor & 15 & 153 & 23 & 182\\
 printer\_room & 4 & 43 & 9 & 21\\
 study\_space & 7 & 121 & 26 & 38 \\
\hline
Total & 26 & 317 & 58 & 241 \\
\hline
\end{tabular}
}
\caption{Number of samples in extra classes considered for the missing modality prediction.} \vspace{-3mm}
\label{dataset2}
\end{table}
\begin{table}[t]
\resizebox{\columnwidth}{!}{
\begin{tabular}{c|cc|cc|cc}
\hline
&  \multicolumn{2}{c|}{Tran-Rec \cite{du2019translate}}  &  \multicolumn{2}{c|}{Tran-Adapt} &
\multicolumn{2}{c}{Tran-Adapt Aug} \\
 & RGB & Depth &   RGB & Depth  &   RGB & Depth\\
 \hline
K $\rightarrow$ X & 0.33 & 0.13  & 0.37 & 0.14 & 0.28 & 0.12 \\
X $\rightarrow$ K & 0.25 & 0.12  & 0.19 & 0.12 & 0.21 & 0.12 \\
K $\rightarrow$ R & 0.26 & 0.22  & 0.22 & 0.17 & 0.25 & 0.18 \\
X $\rightarrow$ R & 0.26 & 0.20  & 0.24 & 0.22 & 0.23 & 0.17 \\
KX $\rightarrow$ R & 0.24 & 0.22  & 0.26 & 0.18 & 0.22 & 0.19 \\
\hline
AVG & 0.27 & 0.18 &  0.26 & 0.17 & \textbf{0.24} & \textbf{0.15}\\
\hline
\end{tabular}
}
\caption{Pixel-to-pixel L2 distance between real and generated images from unseen classes of the target domain. Top results in bold (the lower the better).}\vspace{-6mm}
\label{unseen_classes}
\end{table}

\section{Conclusion}
In this work, we focused on cross-domain learning for multi-modal scene recognition. 
We started by observing the large variability introduced by the plethora of 3D cameras used to collect images in existing scene databases and highlighted that this can cause a significant domain shift that needs a tailored solution. We defined a testbed for studying this problem and performed an evaluation benchmark on several existing methods to evaluate how approaches originally developed for single-domain multi-modal scene recognition and multi-modal cross-domain object classification work on the considered task. 
Moreover, we presented a classification model that exploits self-supervised inter-modality translation as an auxiliary task to reduce domain shift.
Our Translate-to-Adapt successfully outperforms the competitors, showing the effectiveness of its self-supervised task in scene recognition. 

We believe that the novel setting can be of interest to the computer vision and robotics community: the testbed and the experimental analysis are proposed as baselines to pave the way for future research.

\vspace{2mm}\textbf{Acknowledgements}. Computational resources were provided by HPC@PoliTo.

{\small
\bibliographystyle{ieee_fullname}
\bibliography{egbib}
}

\end{document}